# Bayesian learning of forest and tree graphical models

Edmund Jones

A dissertation submitted to the University of Bristol
in accordance with the requirements for award of the
degree of PhD in the Faculty of Science

School of Mathematics

Statistics Group

March 2013

This is a compact version of my PhD thesis, not the official version that was submitted to the University of Bristol. The front matter has been shortened and the page design is different, but the main content is the same. Minor corrections have been made, and these are listed at the end.

Research from this thesis has been published in two journal articles:

## Abstract

Frequentist methods for learning Gaussian graphical model structure are unsuccessful at identifying hubs when $n < p$. An alternative is Bayesian structure-learning, in which it is common to restrict attention to certain classes of graphs and to explore and approximate the posterior distribution by repeatedly moving from one graph to another, using MCMC or other methods such as stochastic shotgun search (SSS). I give two corrected versions of an algorithm for non-decomposable graphs and discuss random graph distributions in depth, in particular as priors in Bayesian structure-learning.

The main topic of the thesis is Bayesian structure-learning with forests or trees. Forest and tree graphical models are widely used, and I explain how restricting attention to these graphs can be justified using theorems on random graphs. I describe how to use methods based on the Chow–Liu algorithm and the Matrix Tree Theorem to find the MAP forest and certain quantities in the full posterior distribution on trees.

I give adapted versions of MCMC and SSS for approximating the posterior distribution for forests and trees, and systems for storing these graphs so that it is easy and efficient to choose legal moves to neighbouring forests or trees and update the stored information. Experiments with the adapted algorithms and simulated datasets show that the system for storing trees so that moves are chosen uniformly at random does not bring much advantage over simpler systems. SSS with trees does well when the true graph is a tree or a sparse graph. Graph priors improve detection of hubs but need large ranges of probabilities to have much effect. SSS with trees and SSS with forests do better than SSS with decomposable graphs in certain cases. MCMC on forests often fails to mix well and MCMC on trees is much slower than SSS.

## Acknowledgements

I would like to thank my supervisor, Vanessa Didelez, for all her advice and guidance. I would also like to thank staff and students of Bristol University and express my gratitude for the funding I have received through the Engineering and Physical Sciences Research Council.

# Contents

# 1 Introduction

## 1.1 Background

In recent years, high-throughput methods and increases in computing power have seen huge increases in the amount of data on DNA and other biomolecules. Much of this data is amenable to analysis by statistical methods. One example is the use of probabilistic graphical models to analyze gene regulation networks. The key task is to deduce the structure of the graphical model from the numerical expression values of a set of genes, observed in a set of cells. These values are measured using microarrays.

This task involves two types of "sparsity". Firstly, the number of observations is usually much less than the number of variables (which is the number of nodes). This is the issue of "$n < p$", a major topic in statistics. Secondly, the graph is believed to have few edges.

Albieri (2010) considered three frequentist algorithms for learning the structure of Gaussian graphical models (GGMs) from numerical data with $n < p$. She used these algorithms on expression values for a set of *E. coli* genes for which the true graph structure had been deduced by biological experiments, and on several simulated datasets that were generated using known graph structures. She found that when the true graph contained hubs (nodes that are connected to many other nodes), the algorithms tended to produce graphs in which the hub and all the nodes connected to it formed a complete subgraph, making it impossible to tell which node was the hub. Hubs are one of the most notable features of gene regulation networks and other real-world networks, so these results suggest that the frequentist algorithms may be unsatisfactory for learning the structures of these networks.

## 1.2 The subjects of this thesis

The main subject of this thesis is Bayesian structure-learning for GGMs in the cases where attention is restricted to forests or trees. Forests are graphs that contain no cycles, and trees are connected forests (see Figure 1.1). Forests and trees are sparse and they exclude the possibility of the large complete subgraphs produced by the algorithms in Albieri (2010).

One of the main questions addressed by the thesis is whether it is sensible to restrict attention to forests or trees when there are existing methods that work on wider classes of graphs. I have done numerous experiments to answer this question. Another is, how should different algorithms for Bayesian structure-learning be evaluated and compared? The thesis is also about prior distributions on the graph structure. One way to improve the discovery of hubs is to use a prior that gives higher probability to graphs that contain hubs.

Restricting to forests or trees and using prior distributions can both be regarded as ways to overcome the difficulties identified by Albieri (2010). But they also have broader applicability and raise new questions. The thesis is mainly about Gaussian graphical models, though some of the results and algorithms are valid for other types of graphical models.

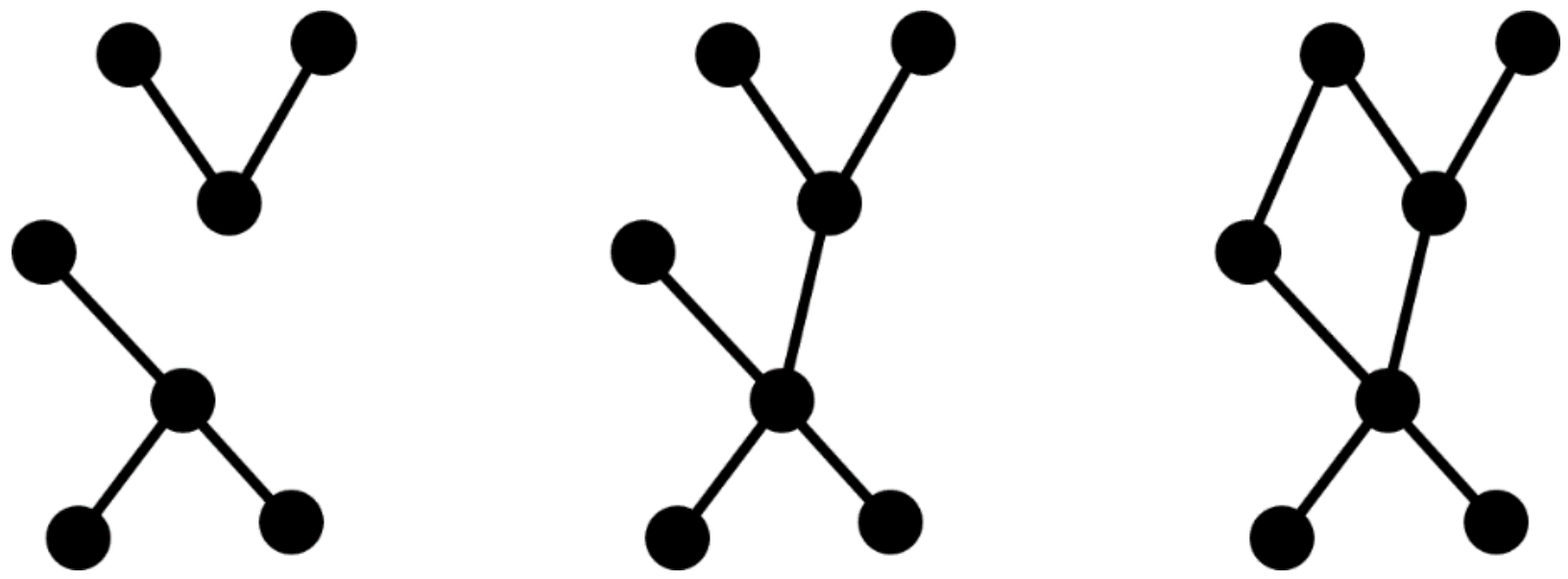

**Figure 1.1.** Left to right: a forest, a tree, and a graph that is neither.

## 1.3 Structure of the thesis

Chapter 2 gives an introduction to graphs and graphical models. Chapter 3 describes the standard Bayesian method for structure-learning of GGMs, in which the number of nodes is fixed and every possible graph has a prior and posterior probability. Next is a review of frequentist methods, including the main ones used by Albieri (2010).

Chapter 4 explains corrections to an algorithm that is used on non-decomposable graphs in Bayesian structure-learning. The purpose of the algorithm is to remove some of a set of extra edges to leave a minimal graph that is still triangulated. I present two corrected versions of the algorithm and detailed discussions of how the original algorithm goes wrong and which of the corrected versions is better.

The prior and posterior distributions on the graph structure are random distributions on the space of graphs with a fixed number of nodes. These random distributions are discussed in depth in chapter 5. Firstly I present two ways of looking at these distributions, “random graph models” and “graph distributions”. I describe the main distributions that have been studied outside the field of graphical models. I then give several definitions of what I call “factored” distributions. These can be used in several of the algorithms for structure-learning that appear in subsequent chapters. (However, they cannot be used as priors that encourage hubs, so I do not use them in my own experiments.) Next I review distributions that have been used as priors in Bayesian structure-learning and discuss the possibility of using graph priors based on random graph models. Finally I present desirable criteria for graph priors and some possible priors that fulfil these criteria.

Chapter 6 is about forest and tree graphical models. I give several reasons why it can be sensible to restrict attention to these relatively small classes of graphs and a detailed and formal consideration of one of these, the notion that sparse graphs are locally tree-like.

Chapters 7 and 8 are about fast algorithms for forest and tree graphical model structure-learning. Chapter 7 is about the Chow–Liu algorithm, which finds the maximum-likelihood tree graphical model. Adaptations of this algorithm can be used to find a forest, using penalized likelihood, or to find the most likely graph in Bayesian structure-learning restricted to trees or forests.

Chapter 8 is about Bayesian structure-learning on trees using methods based on the 19th-century Matrix Tree Theorem. A previously published paper explained how this theorem can be used to find certain quantities exactly in polynomial time. I show how the method works for GGMs and how it can be used to find certain useful quantities such as the posterior expected degrees of the nodes or the expected true-positive rate.

The algorithms in chapters 7 and 8 are fast and produce objects that may be useful in Bayesian structure-learning. But to estimate other quantities and objects, or to produce an estimate of the entire posterior distribution, it is necessary to visit large numbers of individual graphs. With 15 or more nodes, the number of possible graphs is so large that it is computationally infeasible to calculate the posterior probabilities of all of them. This is true even when only forests or trees are considered. Instead, there are algorithms that approximate the posterior distribution by exploring the space of possible graphs.

In chapter 9, I propose new systems for storing forests and trees so that "local moves" to other forests or trees can be made easily and efficiently. For both types of graph, local moves can be chosen uniformly at random from among all possible moves. Section 10.1 describes how to adapt two previously published algorithms for Bayesian structure-learning of GGMs so that they can be used on forests and trees. These algorithms are the reversible-jump MCMC of Giudici & Green (1999) and the stochastic shotgun search (SSS) of Jones et al (2005). Section 10.2 is about how to evaluate and compare frequentist and Bayesian methods for structure-learning. The graph or graphs produced by the algorithm can be compared to the true graph, if that is known.

Chapter 11 mostly consists of computer experiments to evaluate and compare the algorithms and systems in chapters 8–10 and answer the question of whether it is sensible to restrict attention to forests or trees. Section 11.1 gives three new facts about two types of graph, stars and chains, to show that these graphs are extremal in senses to do with the numbers of local moves (equivalently, the numbers of neighbouring graphs). For these reasons stars and chains are used in most of the experiments in the rest of the chapter.

Section 11.2 is about experiments to compare the system for storing trees described in section 9.4 with three alternative systems, using one of my versions of the SSS algorithm. Section 11.3 has experiments on datasets for which the true graph is not a forest but is sparse and locally tree-like, to see whether restricting attention to trees produces good results in this case.

Section 11.4 describes experiments with my versions of the reversible-jump MCMC. Section 11.5 compares SSS restricted to trees with the methods from chapter 8, which calculate exact posterior quantities.

Section 11.6 has experiments with graph prior distributions that are designed to encourage hubs. These priors are compared with the uniform distribution, which has been the most commonly used graph prior in previous research. Finally, section 11.7 compares SSS on trees, SSS on forests, and SSS on decomposable graphs, again to address the question of whether it is sensible to restrict attention to trees or forests.

Chapter 12 presents discussions and possibilities for future research. Appendix I gives the results of some new graph enumerations, including the number of decomposable graphs with 13 nodes, Appendix II is a glossary of terms from graph theory, and Appendix III defines asymptotic notations.

## 1.4 Summary of main contributions

The main contributions of this thesis are as follows.

- (Chapter 4) Corrections to an algorithm for recursive thinning, including explanation of how the algorithm goes wrong and two correct algorithms, with proofs.
- (Section 6.2) Rigorous investigation of the notion that sparse graphs are locally tree-like.
- (Section 8.2) Explanations of how a previously published algorithm can be used for Bayesian structure-learning of tree GGMs and for finding the expected posterior values of certain quantities.
- (Chapter 9) Systems and algorithms for storing forests and trees so that local moves can be made easily and uniformly at random, and numerous propositions related to these.
- (Section 10.1) Modifications of two previously published algorithms for Bayesian structure-learning of GGMs so that they can be used on forests and trees.
- (Chapter 11) Experiments to assess the systems for storing forests and trees, assess how structure-learning with trees performs when the true graph has cycles, compare different graph prior distributions, and compare structure-learning with trees and forests to structure-learning with decomposable graphs.

## 1.5 The meanings of $n$ and $p$

In the Erdős–Rényi random graph model $G(n, p)$, $n$ is the number of nodes and $p$ is the probability of each edge being present. But in multivariate statistics, $n$ is usually the number of data and $p$ is the dimension of the problem, which in graphical models is the number of nodes. Both these systems of notation are very standard in their respective fields.

Both Erdős–Rényi graphs and multivariate statistics arise many times in this thesis, but seldom close to each other. So I use standard notation throughout, except briefly in section 11.3. The meanings of $n$ and $p$ are consistent within individual chapters, but not within the whole thesis. The meanings are stated when the two letters first appear in each chapter and should also be obvious from the context. (Note that $p(\cdot)$ is also used to mean probability density functions, and in chapter 6 the number of data is $m$.)

# 2 Graphs and graphical models

## 2.1 Graphs

### Basic definitions

These definitions are sufficient for this thesis and are not the most general. See also Appendix II, which is a glossary of relevant terms.

- A **graph** $G$ is a pair $(V, E)$, where $V$ is a finite set of nodes, also known as vertices, and $E$ is a set of edges.
- In an undirected graph, the elements of $E$ are unordered pairs $(u, v)$, where $u, v \in V$. (Standard practice is to write unordered pairs using braces, as $\{u, v\}$, but I use regular parentheses, like Edwards et al 2010.)
- In a directed graph, the elements of $E$ are ordered pairs $(u, v)$, where $u, v \in V$.

All graphs considered in this thesis are simple, which means they do not have multiple edges or self-loops. In other words, all the elements of $E$ are distinct, in directed graphs if $(u, v) \in E$ then $(v, u) \notin E$, and in both types of graph if $(u, v) \in E$ then $u \neq v$.

Of course graphs are usually thought of visually. The nodes are dots, and the edges are lines between pairs of dots. A directed edge $(u, v)$ is drawn as an arrow from $u$ to $v$.

- A **subgraph** of $G$ is a graph $H = (V', E')$ where $V' \subseteq V$, $E' \subseteq E$, and $u, v \in V'$ for all $(u, v) \in E'$. The notation $H \subseteq G$ means that $H$ is a subgraph of $G$.
- An **induced subgraph** is a subgraph $(V', E')$ in which $V' \subseteq V$ and $E' = \{(u, v) \in E: u, v \in V'\}$.
- $V_G$ means the set of nodes in $G$ and $E_G$ means the set of edges in $G$. If $(V', E_{V'})$ is an induced subgraph then $E_{V'}$ means the set of edges in the subgraph induced by $V'$.
- The set of **neighbours** of $v$ is $ne(v) = \{u \in V: (u, v) \in E \text{ or } (v, u) \in E\}$.
- The **degree** of $v$ is $\deg v = |ne(v)|$.
- The **size** of $G$ is $|E|$.

### Paths

- In an undirected graph, a **path** is a sequence of distinct nodes $(u_1, u_2, \dots, u_k)$ such that $(u_1, u_2), (u_2, u_3), \dots, (u_{k-1}, u_k) \in E$.
- In a directed graph, a **directed path** is a sequence of distinct nodes $(u_1, u_2, \dots, u_k)$ such that $(u_1, u_2), (u_2, u_3), \dots, (u_{k-1}, u_k) \in E$. It is natural to refer to this as a directed path from $u_1$ to $u_k$. (This is sometimes used as the definition of a "path" in a directed graph—for example, see Lauritzen 1996, page 6.)

- In a directed graph, a **path** is a sequence of distinct nodes $(u_1, u_2, \dots, u_k)$ such that either $(u_1, u_2) \in E$ or $(u_2, u_1) \in E$, either $(u_2, u_3) \in E$ or $(u_3, u_2) \in E$, ..., and either $(u_{k-1}, u_k) \in E$ or $(u_k, u_{k-1}) \in E$. I may refer to $(u_1, u_2, \dots, u_k)$ as a path from $u_1$ to $u_k$, but "from" and "to" do not imply that the path is directed.
- In an undirected graph, a **cycle** is a path $(u_1, u_2, \dots, u_k)$ where $k \geq 3$ and $(u_k, u_1) \in E$.
- The **girth** of a graph is the length of its shortest cycle, or infinity if it has no cycles.
- For a graph to be **connected** means that there is a path between any two nodes.
- Suppose $A$, $B$, and $C$ are induced subgraphs of $G$ with no nodes in common. $C$ **separates** $A$ and $B$ if any path between a node in $A$ and a node in $B$ includes a node in $C$.

## Definitions that only apply to directed graphs

- In an edge $(u, v)$, $u$ is called the **parent** and $v$ is called the **child**.
- The set of **children** of $v$ is $ch(v) = \{u \in V: (v, u) \in E\}$.
- The set of **parents** of $v$ is $pa(v) = \{u \in V: (u, v) \in E\}$.
- The set of **descendants** of $v$ is $de(v) = \{u \in V$: there is a directed path from $v$ to $u\}$.
- The set of **ancestors** of $v$ is $an(v) = \{u \in V$: there is a directed path from $u$ to $v\}$.

## Classes of undirected graph

- A **complete graph** is an undirected graph where $(u, v) \in E$ for all $u, v \in V$. The complete graph on $p$ nodes is called $K_p$.
- A maximal complete subgraph $H$ of $G$ is called a **clique**. (Maximal means there is no complete subgraph $H'$ of $G$ such that $H \subseteq H'$ and $H \neq H'$.)
- A **forest** is an undirected graph that has no cycles. In a forest, any two nodes are connected by at most one path.
- A **tree** is a connected forest. In a tree, any two nodes are connected by exactly one path.
- For a connected graph $G = (V, E)$, a **spanning tree** of $G$ is a tree $T = (V, E')$ such that $E' \subseteq E$.

## Decomposable graphs

Decomposable graphs are a class of undirected graphs that is especially important in graphical models. A **proper decomposition** of $G$ is a pair of induced subgraphs $((A, E_A), (B, E_B))$ such that $V = A \cup B$, $A \neq \emptyset$, $B \neq \emptyset$, $A \neq V$, $B \neq V$, the induced subgraph with node-set $C = A \cap B$ is complete, and $C$ separates $A \setminus C$ from $B \setminus C$. This $C$ is called a separator. It may be possible to decompose $A$ and $B$ further. Following repeated decomposition, the subgraphs that cannot be decomposed any further are called the **prime components** of $G$. If all the prime components are cliques, then the original graph is said to be **decomposable**.

If a graph is decomposable, then its cliques can be put in a perfect sequence (see the definition on pages 14–15 of Lauritzen 1996). As well as the list of cliques, a perfect sequence also gives a list of separators, which are sets of nodes that each induce a complete subgraph.

The lists of cliques and separators are used in expressions for the factorized joint density in graphical models (see section 2.2). The cliques are all distinct, but separators can appear more than once in the list. For this reason the separators will be regarded as a collection in which an element can appear more than once, rather than a set.

The class of decomposable graphs is the same as the class of **chordal graphs**, which have been studied in graph theory. A chordal graph is one in which any cycle of length four or more has a chord—for any cycle $(u_1, u_2, \ldots, u_k)$ where $k \geq 4$ there is an edge $(u_i, u_j)$ where $i, j \in \{1, \ldots, k\}$ and $u_i$ and $u_j$ are not adjacent in the cycle. For a proof that these two classes of graphs are equivalent, see Proposition 2.5 in Lauritzen (1996), which uses "weakly decomposable" instead of "decomposable" and "triangulated" instead of "chordal". In chapter 4 I will use the term "triangulated". Section 2.5 gives other names for this class of graphs.

### Trees and forests

Chapter 9 will use many times the fact that in a tree there is precisely one path between any two nodes. Proposition 2.1 is another simple fact about trees that will be referred to in section 7.2.

***Proposition 2.1.*** Adding one edge to a tree creates a graph that has precisely one cycle.

***Proof.*** Suppose the tree is $T$ and the extra edge is $e = (u, v)$. Being a tree, $T$ contains no cycles. So any cycle in $T + e$, meaning $(V_T, E_T \cup \{e\})$, must consist of $e$ and a path from $u$ to $v$ in $T$. Conversely, any path from $u$ to $v$ in $T$ will give rise to a cycle when $e$ is added. Since $T$ is a tree, there is precisely one path from $u$ to $v$ in $T$, so $T + e$ contains precisely one cycle. □

For another proof, see Theorem 2.1(b) in Even (1979).

All trees and forests are decomposable. The cliques are the pairs of nodes that have edges between them. In trees, the separators are the nodes that have degree 2 or more. In forests, the separators are the nodes that have degree 2 or more and the empty set. So the cliques all have size 2 and the separators all have size 1 or 0.

Definitions and facts to do with rooted trees are given in "Facts about rooted trees", in section 9.2.

## 2.2 Graphical models

### Conditional independence and graphical models

$X$ and $Y$ are **conditionally independent** given $Z$ if $p(x, y \mid z) = p(x \mid z)p(y \mid z)$ for all $x$ and $y$ and for all $z$ such that $p(z) > 0$. In symbols this is written as $X \perp\!\!\!\perp Y \mid Z$.

A graphical model consists of a graph in conjunction with a multivariate statistical model or family of models. Each node $v$ of the graph represents a single variable, $X_v$. In this thesis, it will be assumed that the joint density of these univariate random variables is positive and continuous with respect to a product measure. For more general cases, see chapter 3 of Lauritzen (1996).

The structure of the graph summarizes relations of conditional independence between the variables. In undirected graphical models, if $u, v \in V$ then $(u, v) \notin E \Rightarrow X_u \perp\!\!\!\perp X_v \mid X_{V \setminus \{u,v\}}$. This is called the pairwise Markov property, and with the assumption above it is equivalent to the local and global Markov properties (Lauritzen 1996, section 3.2.1). In directed acyclic graphical models, also known as Bayesian networks, $X_v \perp\!\!\!\perp X_{V \setminus de(v)} \mid X_{pa(v)}$ for all $v \in V$.

Graphical models are used for specifying, analyzing, and interpreting complex relations between random variables. Much of this thesis is about graphical models where the graph is a forest or a tree. For these graphs, there are simple equivalences between directed and undirected graphical models. To explain these requires three definitions. See section 9.2 for the definitions of "rooted tree" and "rooted forest"; and for two graphical models to be "Markov-equivalent" means that they imply the same conditional independence relations. The equivalence between the directed and undirected graphical models is that rooted trees or forests are Markov-equivalent to the undirected trees or forests formed by removing the direction from each edge.

## Gaussian graphical models

This thesis is mainly about Gaussian graphical models (GGMs), which are undirected. These are one of the most widely studied types of graphical model. In a GGM, the variables follow a multivariate normal distribution, $X \sim N_p(\mu, \Sigma)$. One property of this distribution is that $X_i \perp\!\!\!\perp X_j \mid X_{V \setminus \{i,j\}} \Leftrightarrow (\Sigma^{-1})_{ij} = 0$. (Here the nodes are identified with the numbers $\{1, \dots, p\}$.) This can be seen by writing out the joint density and factorizing it, and has been known at least since Wermuth (1976). Using the definition of undirected graphical models, it follows that $(i, j) \notin E \Rightarrow (\Sigma^{-1})_{ij} = 0$. In other words, the edges that are absent from the graph correspond to zeroes in the precision matrix $K = \Sigma^{-1}$ (also known as the concentration matrix).

The object of interest is $\Sigma$ rather than $\mu$, so it is common to set $\mu = 0$. Data can easily be centred so that $\bar{x} = 0$. Suppose $X$ is an $n \times p$ matrix that contains $n$ observations of a $p$-variate Gaussian distribution, and let $1_n$ be an $n \times n$ matrix of 1s. The centred matrix is $(I_n - 1_n/n)X$.

GGMs can be used to model gene regulation networks, as discussed in section 2.4, and financial objects such as currency values (Carvalho et al 2007) and asset returns (Carvalho & Scott 2009). Murray & Ghahramani (2004) state that GGMs are "trivial". This can perhaps be taken to mean that they are simpler than general undirected graphical models.

## Structure-learning

One of the main tasks or problems to do with graphical models is structure-learning. This is the problem of how to infer the graph structure from observations of the random variables. Another is the problem of inference—how to calculate distributions on certain nodes given observations of other nodes.

Maximum-likelihood methods can be used for many statistical problems. But in graphical model structure-learning, the maximum-likelihood graph is always the complete graph, because this implies no restrictions on the variables. The maximum-likelihood estimator of the covariance matrix always exists if $n > p$, but only sometimes exists if $n \le p$ (for details see Lauritzen 1996, section 5.2.1). In microarray experiments (see section 2.4), $n \ll p$, so maximum-likelihood methods cannot be used.

For GGMs there are a variety of frequentist and Bayesian methods for structure-learning. These are described in chapter 3. The main topic of this thesis is Bayesian structure-learning of forests and trees.

## 2.3 The covariance and precision matrices for GGMs

### Possible partial correlations

Firstly, to standardize a matrix $M$ means to replace it by $DMD$, where $D$ is the diagonal matrix whose elements are $d_{ii} = m_{ii}^{-1/2}$. The $(i, j)$ element of the standardized matrix is thus $m_{ij}/\sqrt{m_{ii}m_{jj}}$, so the diagonal elements of the standardized matrix are all 1.

In Gaussian graphical models, not all combinations of partial correlations are possible. Let the precision matrix $\Sigma^{-1}$ be $K$, and consider the standardized precision matrix $C$, where $c_{ij} = k_{ij}/\sqrt{k_{ii}k_{jj}}$ and $-1 \le c_{ij} \le 1$. The partial correlation between $X_i$ and $X_j$ is $r_{ij} = -c_{ij}$ (for $i \ne j$), so $C$ could be called the negative partial correlation matrix. The precision matrix has to be positive-definite, so $C$ also has to be positive-definite.

Sylvester's criterion (Gilbert 1991) states that a matrix is positive-definite if and only if the determinants of all its square upper-left submatrices are positive. These determinants are called the leading principal minors of the matrix. Applying this criterion to $C$ gives a set of algebraic inequalities that must be satisfied by the partial correlations.

For some graphs, these inequalities can be greatly simplified. Consider a graph in which node 1 has edges to all the other nodes, and there are no other edges apart from these. The graph is called a "star" and node 1 is called a "hub". In this case,

$$C = \begin{pmatrix} 1 & c_{12} & c_{13} & \cdots & c_{1p} \\ c_{12} & 1 & 0 & \cdots & 0 \\ c_{13} & 0 & 1 & \cdots & 0 \\ \vdots & \vdots & \vdots & \ddots & \vdots \\ c_{1\mathrm{p}} & 0 & 0 & \cdots & 1 \end{pmatrix}.$$

One of the square upper-left submatrices is the entire matrix. The determinant of this being positive is equivalent to

$$\sum_{j=2}^{p} c_{1j}^2 < 1 \,.$$

If this inequality holds then the other leading principal minors are also positive. So this inequality on its own is a necessary and sufficient condition for $C$ being positive-definite and the distribution being valid. The necessary and sufficient condition on the partial correlations is obviously just $\sum_{j=2}^{p} r_{1j}^2 < 1$ .

It follows, for example, that in a V-shaped graph with three nodes and two edges, at least one of the partial correlations along the edges must have magnitude less than $\sqrt{1/2} \approx 0.707$. More generally, in a star with $s$ "rays", there must be at least one partial correlation on an edge that has magnitude less than $\sqrt{1/s}$.

As a necessary condition, the inequality generalizes to graphs that contain stars as induced subgraphs. This is because the nodes can simply be reordered so that the hub is node 1 and the other $s$ nodes of the star come next. The above argument applied to the upper-left $(s+1) \times (s+1)$ submatrix shows that $\sum_{j=2}^{s+1} r_{1j}^2 < 1$. For example, in any graph that contains a V-shape, which means any graph that does not consist entirely of disjoint cliques, there must be at least one partial correlation on an edge that has magnitude less than 0.707.

As far as I am aware, these conditions on partial correlations in stars have not previously appeared in published research. The closest thing I have found is assumptions A3 and A4 in Kalisch & Bühlmann (2007), which are about the numbers of neighbours of nodes and the magnitudes of the partial correlations in GGMs. These assumptions are also used in Maathuis et al (2009).

For shapes other than stars, it is easy to write down the inequalities that result from Sylvester's criterion, but it is generally not easy to rearrange them into a useful form.

## Possible standard correlations

More widely known than partial correlations, and possibly also of interest, are the standard correlations. These can be found by inverting the standardized precision matrix and standardizing.

As with partial correlations, the conditional independence relations shown by the graph imply conditions on the correlations. However, these conditions are not as simple or notable as the ones for partial correlations. For the V-shaped graph on three nodes,

$$C = \begin{pmatrix} 1 & c_{12} & c_{13} \\ c_{12} & 1 & 0 \\ c_{13} & 0 & 1 \end{pmatrix},$$

which means that the upper triangle of the correlation matrix, found by inverting and then standardizing, is

$$\begin{pmatrix} 1 & \frac{-c_{12}}{\sqrt{1-c_{13}^2}} & \frac{-c_{13}}{\sqrt{1-c_{12}^2}} \\ & 1 & \frac{c_{12}c_{13}}{\sqrt{(1-c_{12}^2)(1-c_{13}^2)}} \\ & & 1 \end{pmatrix}.$$

It can be seen that $corr(X_2, X_3) = corr(X_1, X_2)corr(X_1, X_3)$. The correlation between the two unconnected nodes is the product of the other two correlations.

For arbitrary-sized stars, a standard formula for the inverse of a partitioned matrix can be used to show that

$$corr(X_1, X_i) = -c_{1i}\left[1 - t + c_{1i}^2\right]^{-1/2}$$

$$\text{and } corr(X_j, X_k) = c_{1j}c_{1k}\left[\left(1 - t + c_{1j}^2\right)\left(1 - t + c_{1k}^2\right)\right]^{-1/2} \quad \text{for } j, k \neq 1,$$

where $t = \sum_{m=2}^{p} c_{1m}^2$. Again $corr(X_j, X_k) = corr(X_1, X_j)corr(X_1, X_k)$. There is no simple generalization to graphs that contain stars as induced subgraphs.

## Creating possible covariance matrices

Chapter 11 is about experiments to evaluate and compare algorithms for Bayesian structure-learning of GGMs. These experiments use simulated datasets that each correspond to a particular graph. This subsection is about the issues involved in creating these simulated datasets and several ways of doing it.

Given a covariance matrix $\Sigma$, data from $N_p(0, \Sigma)$ can easily be generated in R or other statistical packages. But creating a possible $\Sigma$ for a given graph is sometimes non-trivial. Necessary and sufficient conditions on $\Sigma$ are that it be symmetric and positive-definite, and that the precision matrix $K = \Sigma^{-1}$ have zeroes in the positions that correspond to absent edges in the graph. Of course the task of creating a possible covariance matrix is equivalent to creating a possible precision matrix or negative partial correlation matrix.

Numerous papers describe experiments that must have involved creating covariance matrices for given graphs, but most do not mention how this was done. It seems likely that the authors chose the partial correlations to all be equal and reasonably large, and then made adjustments as necessary to ensure that the matrix was positive-definite. The papers that do mention how it was done mostly describe specific simple matrices. Meinshausen & Bühlmann (2006, page 1448) generated large random graphs whose nodes have maximum degree 4, and chose all the partial correlations to be 0.245. They state that absolute values less than 0.25 guarantee that the precision matrix is positive-definite. For general graphs, no such statement can be made, as shown in "Possible partial correlations", above. Guo et al (2011, pages 6–7) created precision matrices for "chain" graphs, then added extra edges at random. For each extra edge they set the two corresponding elements of the precision matrix to be a random value from $Unif([-1, -0.5] \cup [0.5, 1])$.

One sure-fire way to create a possible $\Sigma$ is to use the formulas in Appendix A of Roverato (2002). This method was used by Castelo & Roverato (2006). It uses the Cholesky decomposition $K = \Phi^T\Phi$, where $\Phi$ is an upper-triangular matrix. The diagonal elements of $\Phi$, and the elements that correspond to edges in the graph, can be chosen freely, and Roverato calls these the "free" elements. The other elements, which he calls "fixed", have to be calculated according to Roverato (2002)'s equation (10). $K$ and $\Sigma$ can then be calculated from $\Phi$.

For decomposable graphs, the calculations for fixed elements can be avoided, as in Albieri (2010). If the vertices are ordered according to a perfect vertex elimination scheme, then all the fixed elements of $\Phi$ are zero (Roverato 2002, page 408). A perfect vertex elimination scheme is the reverse of a perfect numbering—see Lauritzen (1996, page 15).

The next question is how to choose the free elements of $\Phi$. For very small graphs it is possible to work out explicit formulas for how the elements of $\Phi$ will affect the elements of $K$, but for most graphs it is not. It is undesirable to have partial correlations that are very close to zero, since these edges will be difficult to detect. But in most graphs it is impossible for all the partial correlations to have large magnitude—see "Possible partial correlations" above. The simplest way is to set all the free elements to have the same value, though one's first choice might not give a positive-definite matrix because of hubs or other structures.

An alternative way to create a covariance matrix for a given graph is to first choose any symmetric matrix $K$ such that the diagonal elements are positive and the elements corresponding to absent edges are zero. Find the eigenvalues of $K$, and if any of these are negative, let $-\lambda$ be the lowest one. Replace $K$ with $K + \gamma I$, for some $\gamma > \lambda$. This ensures that all the eigenvalues are positive, so $K$ is positive-definite, without disturbing the off-diagonal zeroes or the symmetry. Schäfer & Strimmer (2005a, pages 757–758) used a similar method, though they added quantities to each of the diagonal elements individually. (The R package "GeneNet", by Schäfer et al 2012, contains a function that performs their method.)

Having created a possible covariance matrix from which to generate simulated data, it is common to standardize the matrix so that the variances are all 1. This ensures that the variables are all on the same scale.

The datasets used in the experiments in chapter 11 mostly correspond to true graphs that are trees. To create the covariances for these datasets, I started by setting $K$ to have 1's on the diagonal and equal values in all the positions that correspond to edges. I then inverted $K$ and standardized to create $\Sigma$.

## 2.4 Biomolecular networks

### Modelling biomolecular networks

The ultimate intended application of my work on GGMs is gene regulation networks. Each gene corresponds to a node, and the numerical value for each node is the logarithm of the expression level of that gene. The networks arise because genes are transcribed to form molecules of mRNA, which are then translated to form proteins, and some of these proteins are transcription factors that promote or inhibit the transcription of other genes (Pournara & Wernisch 2007).

The idea of using GGMs to model gene regulation networks was first proposed in Friedman et al (2000). This paper used Bayesian networks and the Bayesian structure-learning methods described in Heckerman et al (1995). It contains several significant ideas, for example that most of the difficulties arise from the number of variables being much greater than the number of observations—that is, $n \ll p$—and the idea of using prior biological knowledge about the network structure. GGMs have subsequently been used to model gene regulation networks in Castelo & Roverato (2006, 2009), Albieri (2010), Edwards et al (2010), and probably many others—as of February 2013, Google Scholar says that Friedman et al (2000) has been cited 2285 times.

There are numerous public databases that contain the results of experiments to measure gene expression levels, for example the National Center for Biotechnology Information's Gene Expression Omnibus (Barrett et al 2007) and M3D (Faith et al 2008). These databases usually have $n \ll p$.

For details of the preliminary statistical analysis of microarray experiments, including experimental design and how the data is processed and cleaned, see Wit & McClure (2004). Sections 6.2.1 and 6.2.2 of this book present arguments for and against the use of the multivariate Gaussian distribution to model log gene expression levels.

Measurements of gene expression levels are produced using DNA microarrays. These are an example of a high-throughput method—a method that can quickly produce data on large numbers of biomolecules. GGMs can also be used to model other large biomolecular networks. For an overview of this topic see Markowetz & Spang (2007).

It is believed that gene regulation networks and other biomolecular networks tend to have certain properties related to the degrees of the nodes and other features of the graph. These properties are the topic of the next few subsections. Some of the research in this area has not been mathematically rigorous. For example, in Barabási & Albert (1999) the description of the growth of "scale-free" graphs (see below) is not a full definition of a random process, as pointed out in Bollobás et al (2001). In this section I will just report the properties without attempting to state or discuss them in a fully mathematical way. Some of them are discussed further in chapter 5.

## Hubs

Biomolecular networks tend to contain a few nodes, called hubs, that are connected to a large number of other nodes. In gene regulation networks, the biological meaning of a hub is that one gene codes for a protein that regulates the expression of many other genes. Hubs are probably the most notable and widely recognized characteristic of biomolecular networks. For example, Barabási & Oltvai (2004) report that the transcriptional gene regulation networks of *E. coli* and *S. cerevisiae* (yeast) contain disproportionately many hubs. In Alterovitz & Ramoni (2006), Figure 1 shows several hubs in the *E. coli* gene regulation network that have very large numbers of neighbours. Royer et al (2008) states that hubs are also a feature of protein networks, and that the abundance of hubs can be explained by models of evolution.

## Other motifs

Certain small-scale motifs also seem to be common in biomolecular networks. Motifs are subgraphs, or induced subgraphs, that appear more often in real networks than in "randomized networks" (Milo et al 2002). Milo et al (2002) set out to find three- and four-node motifs in various well-studied directed networks including the *E. coli* and *S. cerevisiae* gene regulation networks. They compared these real networks to random graphs where each node had the same number of incoming and outgoing edges as in the real networks, and found that the feed-forward loop and the bi-fan—see Figure 2.1—occur far more frequently in the gene regulation networks. The $Z$-values for these motifs ranged from 10 to 41.

Royer et al (2008) states that protein interaction networks tend to have cliques and bicliques. A biclique is two sets of nodes where every node in one set is connected to every node in the other—see Figure 2.1. Alon (2007) shows a directed biclique from the transcription regulation network of *E. coli* (in his Figure 6).

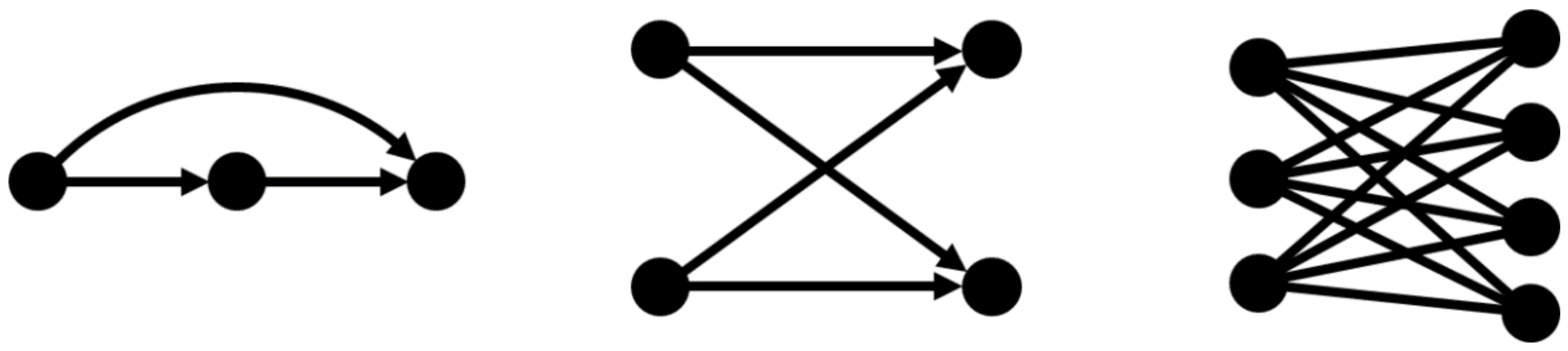

**Figure 2.1.** Left to right: a feed-forward loop, a bi-fan, and a biclique. These motifs have been found to be common in real-world networks.

## Sparsity

It is widely believed that biomolecular networks are sparse, meaning that they have few edges. The precise meaning of this statement is discussed in depth in section 6.2. For example, Leclerc (2008) reports the numbers of nodes and edges in gene networks for Arabidopsis, Drosophila, and three other widely studied model organisms. All these networks are sparse. Pournara & Wernisch (2007) state that gene regulation networks are sparse because most genes are known to be regulated by a small number of transcription factors and most transcription factors regulate a small number of genes. Protein interaction networks are also believed to be sparse (Spirin & Mirny 2003). Many papers simply assert without elaboration that biomolecular networks are sparse—for example Wille & Bühlmann (2006) and Han et al (2007).

## Scale-free networks

Barabási & Oltvai (2004) state that the most striking feature of biomolecular networks, as well as social and technological networks, is that they are approximately "scale-free". This term was introduced by Barabási & Albert (1999) and means that, over a large range, the degrees of the nodes follow a power law; that is, the probability that a node has degree $k$ is proportional to $k^{-\gamma}$. In most cases, $2 < \gamma < 3$.

In scale-free graphs most nodes have small degrees, but a few have very high degrees. (In other words, the power-law distribution is long-tailed.) So for biomolecular networks both the sparsity and the relative preponderance of hubs could be regarded as consequences of being scale-free.

Jeong et al (2001) reports that the *S. cerevisiae* protein network follows a power law, and Jeong et al (2000) studied metabolic networks of 43 species and found strong evidence of power laws. However, Barabási & Oltvai (2004) state that in the transcriptional gene regulation networks of *E. coli* and *S. cerevisiae*, the degree distributions are mixtures of power laws and exponential distributions.

### Log-transformation

Gene expression data needs to be log-transformed before being modelled by the multivariate Gaussian distribution. But websites, online databases, and papers that present this kind of data do not always state whether this has been done.

For example, Albieri (2010) used a gene expression dataset with 100 nodes and 43 observations that was a subset of the EcoliOxygen dataset in the R package "qpgraph" (Castelo & Roverato 2009). The EcoliOxygen dataset is reported in Covert et al (2004) and available from the National Center for Biotechnology Information's Gene Expression Omnibus (*http://www.ncbi.nlm.nih.gov/geo/*, Barrett et al 2007), where it is record number GDS680. Without looking at the numerical values themselves, it is not obvious whether the values in EcoliOxygen have been log-transformed. Covert et al (2004) mentions a $t$-test on log-transformed data, but that is all.

I found a different database of *E. coli* gene expression data, M3D (*http://m3d.bu.edu/*, Faith et al 2008), where it is stated that the values are log-transformed. I plotted a histogram of all the expression levels from M3D and a histogram of all the EcoliOxygen data. The two distributions looked similar, which suggests that the EcoliOxygen data have been log-transformed.

## 2.5 Supplementary notes: alternative terms and the history of graphical models

Books about graphical models include Pearl (1988), Whittaker (1990), Edwards (1995), Lauritzen (1996), Cowell et al (2007), and Koller & Friedman (2009). Graphical models are also known as probabilistic graphical models or graphical Markov models (Wermuth 1998, Wermuth & Cox 2001). Undirected graphical models are sometimes called Markov random fields, and directed acyclic graphical models are often called Bayesian networks (Bayes nets for short) or belief networks.

For a brief history of graphical models see Wermuth (1998), in which their origins are traced back to the early twentieth century. Gaussian graphical models originate in Dempster (1972). But Dempster did not mention graphs or conditional independence. What he proposed was to simplify the multivariate normal distribution $N_p(\mu, \Sigma)$ by setting some elements of $\Sigma^{-1}$ to zero. Dempster called this "covariance selection", and as a result Gaussian graphical models are also known as covariance selection models. They are occasionally called concentration graph models (Wermuth & Cox 2001).

The task of inferring the graph structure from data is often called "structural learning", though I prefer "structure-learning". It is also referred to as "model selection", "reverse engineering" (Alon 2003, Castelo & Roverato 2009, Maathuis et al 2010), "topology discovery" (Anandkumar et al 2011), and "estimation of structure" (Lauritzen 2012). For GGMs it is sometimes called covariance selection. Structural learning contrasts with "quantitative learning", which means estimating the numerical parameters of the probability distribution (Giudici 1996).

In the field of graphical models it is common to talk about "decomposable" graphs. In graph theory these are called chordal graphs (Gavril 1974, Diestel 2005, Bondy & Murty 2008). Sometimes they are called triangulated graphs (Rose 1970, Rose 1972, Berge 1973, Lauritzen 1996, Diestel 2005). They have also been called rigid circuit graphs (Dirac 1961), perfect elimination graphs (Rose et al 1976), and monotone transitive graphs (Rose 1972).

Regarding the terms “star” and “hub”, it is not ideal to use words that are unrelated in the real world for mathematical objects that are closely related. But “hub” often refers to the centre of a network, so its use in describing graphs is natural; and it is useful to have the separate word “star” for the hub and the nodes connected to it. Both are commonly used—“hub” in Barabási & Oltvai (2004) and Albieri (2010), for example, and “star” in Royer et al (2008) and Yuan & Lin (2007).

# 3 Structure-learning for GGMs

## 3.1 Bayesian methods

### The standard Bayesian method

Bayesian learning of graphical model structure involves a likelihood, a prior distribution, some data, and a posterior distribution. The prior and the posterior are both distributions on the set of all graphs with the appropriate number of nodes, or in the case that only a certain subset of graphs is considered, they are distributions on that set of graphs. The prior is specified by the user or researcher and the posterior is calculated from the prior and the data.

The most widely used Bayesian method for learning Gaussian graphical model structure requires a prior distribution on $\Sigma$ as well as the prior on the graph structure. Suppose there are $p$ nodes. Let $x$ be the $n \times p$ matrix ($n$ rows, $p$ columns) of the $n$ observed data:

$$x = \begin{pmatrix} x_1^T \\ x_2^T \\ \vdots \\ x_n^T \end{pmatrix},$$

and let $U = x^T x$. ($U$ is usually called $S$, but I use $S$ to mean separators.) The likelihood is

$$p(x \mid G_i, \Sigma) = (2\pi)^{-np/2} |K|^{n/2} \exp\left[-\frac{1}{2}\sum_i x_i^T K x_i\right]$$

$$= (2\pi)^{-np/2} |K|^{n/2} \exp\left[-\frac{1}{2}\operatorname{tr}(KU)\right],$$

where $K = \Sigma^{-1}$, and the posterior probability of $G_i$ being the true graph is

$$p(G_i \mid x) = \frac{p(x \mid G_i) p(G_i)}{\sum_j p(x \mid G_j) p(G_j)}.$$

The meaning of $p(\cdot)$ changes according to its arguments. Of the terms on the right-hand side, $p(G_i)$ is the prior probability of $G_i$ and $p(x \mid G_i)$ is the marginal likelihood:

$$p(x \mid G_i) = \int_{\Sigma^{-1} \in M^+(G_i)} p(x \mid G_i, \Sigma)\, p(\Sigma \mid G_i)\, d\Sigma\,.$$

Here $M^+(G_i)$ is the set of positive-definite matrices that have zeroes in the positions that correspond to absent edges in $G_i$. So the integral is over all values of $\Sigma$ that are possible for $G_i$.

For $p(\Sigma \mid G_i)$ it is common to use the generalized hyper inverse Wishart (HIW) distribution (Dawid & Lauritzen 1993), which is conjugate. This and other priors for $\Sigma$ are described in the next few subsections. Graph priors are discussed in chapter 5.

## Complete graphs

If the graph is known to be the complete graph, $K_p$, then the distribution of $x$ is just the multivariate Gaussian distribution with no conditional-independence restrictions, and the conjugate prior for $\Sigma$ is the inverse Wishart distribution. This is defined as follows.

If $X$ is an $m \times p$ matrix where each row is an independent sample from the $p$-variate Gaussian distribution with zero mean and covariance matrix $V$, then the $p \times p$ matrix $U = X^T X$ has the Wishart distribution with scale matrix $V$ and $m$ degrees of freedom. I will write this as $U \sim W(m; V)$. For $U$ to be invertible with probability 1, it is necessary that $m \geq p$. The distribution of $\Sigma = U^{-1}$ is then the inverse Wishart distribution with inverse scale matrix $D = V^{-1}$ and $m$ degrees of freedom. I will write this as $IW(\delta, D)$, where $\delta = m - p + 1$.

The only restrictions on the parameters for the Wishart distribution are that $V$ be positive-definite and $m$ be positive. The only restrictions on the parameters for the inverse Wishart distribution are that $D$ be positive-definite and $m \geq p$, which means $\delta \geq 1$. Obviously under these characterizations $m$ and $p$ are both positive integers.

If $\Sigma \sim IW(\delta, D)$, then the density of $\Sigma$ is

$$p(\Sigma) = \frac{|D|^{(\delta+p-1)/2} \exp\left[-\frac{1}{2}\operatorname{tr}(D\Sigma^{-1})\right]}{2^{(\delta+p-1)p/2}\, |\Sigma|^{p+\delta/2}\, \Gamma_p\big((\delta+p-1)/2\big)}$$

$$= \frac{\left|\frac{D}{2}\right|^{(\delta+p-1)/2} \exp\left[-\frac{1}{2}\operatorname{tr}(D\Sigma^{-1})\right]}{|\Sigma|^{p+\delta/2}\, \Gamma_p((\delta+p-1)/2)}$$

(Giudici & Green 1999, page 787; Roverato 2002, page 396). Here $\Gamma_p$ is the multivariate gamma function (James 1964), defined by

$$\Gamma_p(a) = \pi^{p(p-1)/4} \prod_{j=1}^{p} \Gamma[a + (1-j)/2]\,.$$

The "normalizing constant" for the inverse Wishart distribution is the part of the formula for the density that does not involve $\Sigma$ (Jones et al 2005). There is no problem with terms like $|\Sigma|^{p+\delta/2}$, where the exponent can be non-integer, since $\Sigma$ and $D$ are both positive-definite and so their determinants are positive.

## Decomposable graphs

For a decomposable graph, the conjugate prior for $\Sigma$ is the hyper inverse Wishart (HIW) distribution. This was defined by Dawid & Lauritzen (1993) and also described in detail in Giudici & Green (1999).

For a given decomposable graph, suppose the cliques have covariances $\Sigma_C$ and the prior on each $\Sigma_C$ is $IW(\delta, D_C)$, where $\delta$ is some positive number that is the same for all cliques. Dawid & Lauritzen (1993) showed that these distributions on the cliques induce a unique hyper Markov distribution on $\Sigma$, the covariance for the whole graph. In this distribution, $\Sigma$ is constrained so that its inverse has zeroes in the appropriate places, which means the distribution is Markov on the graph. They called this the hyper inverse Wishart distribution and showed that it is conjugate for the family of multivariate Gaussian distributions that are Markov on the graph.

Two issues that arise in specifying the $D_C$'s are hyperconsistency and compatibility. Hyperconsistency means that the distributions of the clique covariances have to be the same where they overlap, so $\left(D_{C_1}\right)_{ij} = \left(D_{C_2}\right)_{kl}$ whenever $(i,j)$ and $(k,l)$ identify the same edge. Compatibility between the distributions on two graphs means that any clique that appears in two graphs has the same distribution in both cases. Hyper-consistency is essential but compatibility is merely desirable. Probably the simplest way to ensure hyperconsistency and compatibility is to choose a single $p \times p$ matrix $D$, and for each graph let each $D_C$ or $D_S$ be the appropriate submatrix of $D$. For full details of these issues see Dawid & Lauritzen (1993) or Giudici & Green (1999). For incomplete graphs not every element of $D$ is used.

I will parameterize the HIW distribution using a $p \times p$ matrix $D$ and write it as $HIW_G(\delta, D)$. If $\Sigma \sim HIW_G(\delta, D)$ then the density of $\Sigma$ is

$$p(\Sigma) = \frac{\prod_C p(\Sigma_C \mid G)}{\prod_S p(\Sigma_S \mid G)},$$

where $\Sigma_C \sim IW(\delta, D_C)$ and $\Sigma_S \sim IW(\delta, D_S)$. The product in the numerator is over the set of cliques, and the product in the denominator is over the collection of separators. A separator may appear more than once in this collection.

For the HIW prior to be proper, it is sufficient that $\delta > 2$ (Roverato 2002, page 402; Jones et al 2005, page 390) and $D^{-1} \in M^{+}(G)$ (Atay-Kayis & Massam 2005, page 322). If the prior on $\Sigma$ is $HIW_G(\delta, D)$, then the posterior is $HIW_G(\delta + n, D + U)$, where $U = x^T x$ is a sufficient statistic for the data $x$.

The marginal likelihood $p(x \mid G)$ can be found explicitly as follows. In the following expressions, $|\Sigma|$ is the determinant of $\Sigma$ but $|C|$ is the number of elements in $C$:

$$p(x \mid G) = \int p(x \mid G, \Sigma) p(\Sigma \mid G)\, d\Sigma$$

$$= \int \frac{\prod_C (2\pi)^{-n|C|/2} |\Sigma_C|^{-n/2} \exp\left[-\frac{1}{2} tr(U_C \Sigma_C^{-1})\right]}{\prod_S (2\pi)^{-n|S|/2} |\Sigma_S|^{-n/2} \exp\left[-\frac{1}{2} tr(U_S \Sigma_S^{-1})\right]}$$

$$\cdot \frac{\prod_C \dfrac{\left|\frac{D_C}{2}\right|^{\frac{\delta+|C|-1}{2}}}{\Gamma_{|C|}\left(\frac{\delta+|C|-1}{2}\right)} |\Sigma_C|^{-\frac{\delta+2|C|}{2}} \exp\left[-\frac{1}{2}\mathrm{tr}(D_C\, \Sigma_C^{-1})\right]}{\prod_S \dfrac{\left|\frac{D_S}{2}\right|^{\frac{\delta+|S|-1}{2}}}{\Gamma_{|S|}\left(\frac{\delta+|S|-1}{2}\right)} |\Sigma_S|^{-\frac{\delta+2|S|}{2}} \exp\left[-\frac{1}{2}\mathrm{tr}(D_S\, \Sigma_S^{-1})\right]} d\Sigma$$

$$= \frac{\prod_C \left|\frac{D_C}{2}\right|^{\frac{\delta+|C|-1}{2}} / \, \Gamma_{|C|}\left(\frac{\delta+|C|-1}{2}\right)}{\prod_S \left|\frac{D_S}{2}\right|^{\frac{\delta+|S|-1}{2}} / \, \Gamma_{|S|}\left(\frac{\delta+|S|-1}{2}\right)}$$

$$\cdot\, (2\pi)^{-np/2} \int \frac{\prod_C \, |\Sigma_C|^{-\frac{\delta+n+2|C|}{2}} \exp\left[-\frac{1}{2}\mathrm{tr}(\{D_C+U_C\}\,\Sigma_C^{-1})\right]}{\prod_S \, |\Sigma_S|^{-\frac{\delta+n+2|S|}{2}} \exp\left[-\frac{1}{2}\mathrm{tr}(\{D_S+U_S\}\,\Sigma_S^{-1})\right]} d\Sigma\,.$$

In these integrals the measure $d\Sigma$ can be taken to be the product of the Lebesgue measures on the elements of the incomplete covariance matrix, which contains only the elements of $\Sigma$ that correspond to edges in the graph (Giudici & Green 1999). The exponent of $2\pi$ is simplified using $\sum_C |C| - \sum_S |S| = p$, which follows from the definition of a perfect sequence. In the second large expression, the first big fraction is the normalizing constant for the HIW prior density (the part of this density that does not involve $\Sigma$) and the integrand is the HIW posterior density without its normalizing constant. It follows that

$$p(\,x \mid G\,) = (2\pi)^{-np/2} \frac{\prod_C \frac{k(C,\delta,D)}{k(C,\delta+n,D+U)}}{\prod_S \frac{k(S,\delta,D)}{k(S,\delta+n,D+U)}},$$

where $k$ is the normalizing constant for each clique or separator:

$$k(C,\delta,D) = \frac{\left|\frac{D_C}{2}\right|^{\frac{\delta+|C|-1}{2}}}{\Gamma_{|C|}\left(\frac{\delta+|C|-1}{2}\right)}.$$

Once the marginal likelihood of a graph has been calculated, it is easy to find its unnormalized posterior probability, since $p(\,G \mid x\,) \propto p(\,x \mid G\,)p(G)$.

## General graphs

For graphs that may or may not be decomposable, the conjugate prior for $\Sigma$ is the generalization of the HIW distribution given by Roverato (2002). This is called the *G*-Wishart distribution in Atay-Kayis & Massam (2005), Lenkoski & Dobra (2011), and Wang & Li (2012).

As in the decomposable case, the density can be written as the product of densities on the prime components divided by the product of densities on the separators (Roverato 2002, Proposition 2). For the separators and complete prime components, the density is the inverse Wishart distribution, as before. For any incomplete prime components, the density is

$$p(\Sigma_P^E) \propto |\Sigma_P|^{-\frac{\delta-2}{2}} J(\Sigma_P^E) \exp\left[-\frac{1}{2}\mathrm{tr}(\Sigma_P^{-1} D_P)\right].$$

Here $E$ is the edge-set of the prime component $P$. The reason for writing the density as a function of $\Sigma_P^E$, rather than just $\Sigma_P$, is to emphasize that its dimension equals the number of free (unconstrained) elements in $\Sigma_P$. (In contrast, with cliques and separators the dimension of the random variable is $|C|(|C|+1)/2$, or the same with $S$—the full number of elements in the Cholesky square root.) Some of the non-free elements of $\Sigma_P$ appear in the expression to the right of the proportional symbol. It would also be possible to just write $\Sigma_P$ throughout. The term $J(\Sigma_P^E)$ is the Jacobian for the transformation from $K_P^E$ to $\Sigma_P^E$.

To find the marginal likelihood, let $k(P,\delta,D)$ be the normalizing constant in the expression for $p(\Sigma_P^E)$, so that

$$k(P,\delta,D)^{-1} = \int_{\Sigma_P^E|P} |\Sigma_P|^{-\frac{\delta-2}{2}} J(\Sigma_P^E) \exp\left[-\frac{1}{2}\mathrm{tr}(\Sigma_P^{-1}D_P)\right] d\Sigma_P^E.$$

This integral cannot be calculated exactly and is discussed in the next subsection. As with decomposable graphs, the marginal likelihood factorizes according to the decomposition of the graph:

$$p(x \mid G) = (2\pi)^{-np/2} \frac{\prod_P \frac{k(P,\delta,D)}{k(P,\delta+n,D+U)} \prod_C \frac{k(C,\delta,D)}{k(C,\delta+n,D+U)}}{\prod_S \frac{k(S,\delta,D)}{k(S,\delta+n,D+U)}}.$$

The three products are over the incomplete prime components, the cliques, and the separators. The $k$'s in the first product in the numerator are defined by the equation with the integral, and the $k$'s in the other two products are as in the previous subsection.

## Calculating the normalizing constant for incomplete prime components

The problem with the above expression for the marginal likelihood is that $k(P,\delta,D)$, the normalizing constant for incomplete prime components, cannot be calculated exactly. For calculating it approximately, Roverato (2002) presents a method that uses importance sampling and Atay-Kayis & Massam (2005) give a method that uses simple Monte Carlo. Lenkoski & Dobra (2011) use a Laplace method that is quicker but less accurate. Moghaddam et al (2009) give two other Laplace-type methods.

Moghaddam et al (2009) describe Monte Carlo methods as the "gold standard" for this problem. Section 4.2 of Atay-Kayis & Massam (2005) presents their Monte Carlo method as a step-by-step algorithm. First, change variables from $\Sigma$ to $K$, and then to $\Phi$, the Cholesky square root of $K$. Then change variables to $\Phi$ post-multiplied by the inverse of the Cholesky square root of $D^{-1}$. Next, manipulate this expression into the form of a multiple of the expectation of a function with respect to chi-squared and univariate normal random variables (the former corresponding to the diagonal elements of the matrix, the latter corresponding to the edges that are present in the graph). Finally, approximate the integral using simple Monte Carlo.

Roverato (2002) used the generalized HIW distribution to analyze Fisher's iris data. This is a well-known set of multivariate data with $p = 4$ that was published in Anderson (1935) and used in Fisher (1936). I have done the same analysis of this dataset, using the same values of the HIW hyperparameters as Roverato (2002) and the same uniform graph distribution, using Java. Instead of Roverato's importance-sampling method, I used Atay-Kayis & Massam (2005)'s simple Monte Carlo method. The posterior distribution that I found was very close to Roverato's—see Figure 3.1. The reasons it was not exactly the same were probably that both methods are random and that Roverato (2002) only used 15,000 samples for the importance sampling whereas I used a billion for the Monte Carlo method. The top graph is a four-cycle, which is of course non-decomposable.

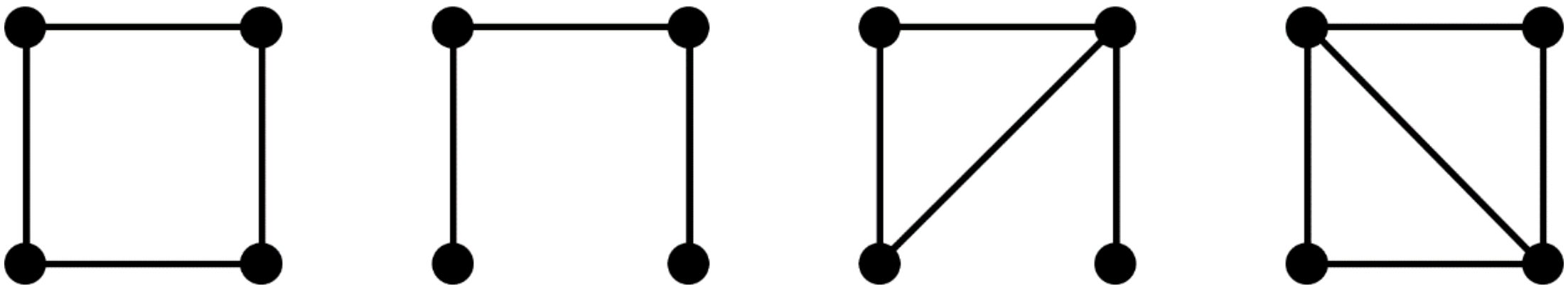

| | | | |
|---|---|---|---|
| 0.14797 | 0.13481 | 0.10590 | 0.10583 |
| 0.147 | 0.135 | 0.106 | 0.106 |

**Figure 3.1.** The top four graphs for the iris data. Below each graph is its posterior probability according to my program, with 1 billion iterations of the Monte Carlo method, and according to Roverato (2002).

## Exploring the posterior distribution

For small $p$ it is possible to calculate the posterior probability for every possible graph. For $p$ larger than about 10, this is computationally infeasible, because there are too many graphs, even if attention is restricted to only decomposable ones. The solution is to somehow explore the space of graphs, moving from one graph to another repeatedly. Madigan & Raftery (1994) presented methods for doing this in an ad-hoc way, for both directed and undirected graphical models.

Giudici & Green (1999) gave a reversible-jump MCMC algorithm for approximating the posterior distributions of $\Sigma$ and the graph structure, in the case that attention is restricted to decomposable graphs. The dimension-changing proposals consist of adding or deleting a single edge to the graph structure. The posterior graph distribution is taken to be the proportion of time spent at each graph. Asymptotically the Markov chain gives a sample from the exact true posterior distribution. Brooks et al (2003) give an adaptation of this method, and Green & Thomas (2013) give another MCMC algorithm for the same problem, which stores and manipulates not graphs but junction trees.

As an alternative to MCMC, Jones et al (2005) proposed a "stochastic shotgun search" algorithm for exploring either the space of all possible graphs or the space of all decomposable graphs. At each step, this calculates the unnormalized posterior probability of several neighbouring graphs, and then chooses which one to move to according to a certain distribution based on those unnormalized probabilities.

Section 10.1 gives full descriptions of how the algorithms of Giudici & Green (1999) and Jones et al (2005) can be adapted to the cases where attention is restricted to forests or trees. Chapter 11 is about experiments to assess how well these adapted algorithms do.

Moghaddam et al (2009) propose a "neighbourhood fusion" method for exploring the posterior graph distribution for general GGMs. To do this, for each node use lasso regression or a similar method to estimate its neighbourhoods of all possible sizes, and calculate a probability for each neighbourhood. Then repeatedly sample from these possible neighbourhoods, combine them to create a graph, and calculate its score. Dobra et al (2011) give an MCMC method for general graphs. For non-decomposable graphs, the method avoids the need to find the posterior normalizing constant of the HIW distribution, which is the most time-consuming part of the calculations.

### An alternative conjugate prior

One possible weakness of the HIW prior is that it only has one scalar parameter (a "shape" parameter). Letac & Massam (2007) define an alternative prior, for decomposable graphs, that has a scalar parameter for each possible clique and separator and is thus more flexible. This is a generalization of the HIW distribution and is still conjugate. Rajaratnam et al (2008) give a reference prior (in other words, a non-informative prior—see Rajaratnam et al 2008, page 2819, or Gelman et al 2004, page 61) that is an improper special case of Letac & Massam's.

### An alternative method that just uses a prior for the covariance matrix

The rest of this thesis uses the HIW prior on $\Sigma$ (though many sections are more general and not directly related to GGMs or $\Sigma$). But this is not the only Bayesian method for learning GGM structure. Wong et al (2003) give a prior for $\Sigma^{-1}$ that enables its off-diagonal elements to be zero with positive probability. In effect this combines the priors for $\Sigma$ and the graphs into a single distribution. The prior is constructed as follows. Firstly they write $\Sigma^{-1}$ as $TCT$, where $C$ is the negative partial correlation matrix and $T$ is diagonal. $T_{ii}^2$ is given an uninformative gamma prior. For $C_{ij}$ they use a hierarchical prior: each element is zero (corresponding to the edge being absent) with a certain probability, and then $C$ is distributed uniformly in the space of possible values. They describe a reversible-jump MCMC scheme for generating values of $\Sigma^{-1}$. The proposal distributions are the full conditional distributions of $T_{ii}$ and $C_{ij}$, both approximated by normal distributions. The distribution for $C_{ij}$ is a mixture that uses the indicator function $\mathbb{I}[C_{ij} = 0]$.

This method removes the need for a separate graph prior. It also applies to all graphs in one go, whereas the HIW distribution described above is a separate distribution on $\Sigma$ for every graph. Experiments in Wong et al (2003) suggest that when $\Sigma^{-1}$ is sparse the method works well compared to the maximum likelihood estimator of $\Sigma$ and two estimators proposed by Yang & Berger (1994). The comparisons used two loss functions from the same paper.

With this method it is not possible to use any detailed prior beliefs about the graph structure. The user can only specify a prior distribution for $\psi$, the probability that each edge is present. Another possible disadvantage is that it is not possible to calculate anything about the posterior distribution exactly. In contrast, with the HIW prior there is an explicit formula for the posterior probabilities of decomposable graphs.

## 3.2 Frequentist methods

### Preamble

There are also various frequentist methods for GGM structure-learning. These produce a single graph rather than a distribution over a set of graphs. Albieri (2010) is a review and comparison of some of the main frequentist methods. Three of these are described below. See also Dobra et al (2004), Castelo & Roverato (2006), and chapter 20 of Koller & Friedman (2009).

Some of the many methods for DAG structure-learning may be suitable for GGMs. See for example chapter 18 of Koller & Friedman (2009) or Gasse et al (2012). There has also been various research on estimating the covariance matrix that makes little or no mention of graphs or graphical models, for example Yang & Berger (1994), Liechty et al (2004), or Bickel & Levina (2008).

### The simple frequentist method

This method is described by Albieri (2010) on pages 20–21. Find the sample covariance matrix, invert it to find the sample precision matrix, and then standardize (see section 2.3) to find the sample negative partial correlation matrix. Draw an edge between each pair of nodes if and only if the magnitude of their sample partial correlation is above a certain threshold. The appropriate threshold can be calculated using the fact that if the true partial correlation between two nodes is zero, then the sample partial correlation follows a $t$-distribution (Lauritzen 1996, section 5.2.2; Albieri 2010, section 3.3.3), and using multiple-testing procedures as described by Drton & Perlman (2007).

The problem is that when $n < p$, the sample covariance matrix is singular and cannot necessarily be inverted. Formerly, the standard methods for graphical model structure-learning were greedy stepwise forward-selection and backward-elimination—see Whittaker (1990, section 8.4) or Edwards (1995, sections 6.1–6.2). But these fail to account for multiple testing (Edwards 1995, page 138).

### The shrinkage / empirical Bayes method

This method was proposed in Schäfer & Strimmer (2005a,b). To estimate the covariance matrix, they use a linear “shrinkage” of the unbiased estimator towards a diagonal estimator in which the variances are not necessarily equal (Schäfer & Strimmer 2005b). This shrinkage estimator is always positive-definite, so it can be inverted to find estimators of the precision and partial correlation matrices.

The next step is to test the partial correlations. The distribution of the estimated partial correlations is claimed to be similar to the exact distribution (Hotelling 1953), which appears in the maximum likelihood method. The number of degrees of freedom for this distribution is estimated from the data—this is the “empirical Bayes” step (Schäfer & Strimmer 2005a). This ultimately gives a threshold to which the estimated partial correlations are compared to decide which edges are present in the graph.

## Lasso-type methods

The "lasso" (Tibshirani 1996) is a method for estimating coefficients in standard linear models. Let $\{y_i\}$ be the observations, $\{x_{ij}\}$ be the observed covariates, and $\{\beta_j\}$ be the regression coefficients, and assume that $\bar{y} = 0$. The coefficients are chosen to minimize the residual sum of squares

$$\sum_i \left( y_i - \sum_j \beta_j x_{ij} \right)^2$$

subject to $\|\beta\|_1 \leq t$. Here $t$ is a tuning parameter and $\|\beta\|_1$ is the $L_1$ norm of $\beta$, which is $\sum_j |\beta_j|$. This method often gives coefficients that are exactly zero, which means that the corresponding covariates do not appear in the model.

Several methods inspired by the lasso have been proposed for GGM structure-learning. Meinshausen & Bühlmann (2006) proposed a "neighbourhood selection" method. For each node $i$, do lasso regression with $X_i$ as the observation and all the other nodes $X_{V\setminus\{i,j\}}$ as covariates; the nodes for which the regression coefficients are non-zero are taken to be the estimated neighbourhood of $i$ in the graph. To estimate the whole graph structure, the edge $(i, j)$ is claimed to be present if and only if $i$ is in the estimated neighbourhood of $j$ and vice versa—alternatively, the same thing but with "or vice versa".

Friedman et al (2007) present a method that gives estimates of the graph structure and the whole of the precision matrix. The idea is to maximize the log-likelihood penalized by the $L_1$ norm of $K$,

$$\log|K| - \mathrm{tr}(SK) - \rho\|K\|_1 ,$$

over non-negative-definite matrices $K$. Here $K = \Sigma^{-1}$, $S$ is the empirical covariance matrix, $\rho$ is a tuning parameter, $|K|$ is the determinant, and $\|K\|_1 = \sum_{i,j}|K_{ij}|$ (this sigma means a sum). This is equivalent to a minimization problem that resembles a lasso problem as in Tibshirani (1996)—see Banerjee et al (2008) for details. Friedman et al (2007)'s contribution is the "graphical lasso algorithm" for solving the minimization problem. This gives an estimate of $\Sigma$ that can be inverted reasonably fast to give an estimate of $K$. Their experiments suggest that their algorithm is much faster than the rival one in Banerjee et al (2008), but the computation time depends greatly on $p$.

Yuan & Lin (2007) set out to maximize the same penalized log-likelihood, except that they omit the diagonal elements of $K$ from the penalty. Meinshausen (2008) shows that this method is not consistent for estimating the graph structure. For a certain graph and covariance matrix, it gives the wrong graph structure in the "population case", where the MLE of the covariance equals the true covariance, and with positive probability in the case of finite samples.

## Finding hubs

Albieri (2010) compared the shrinkage / empirical Bayes method, the graphical lasso, and the PC algorithm of Kalisch & Bühlmann (2007), which is for structure-learning of directed acyclic graphical models. She found that none of these algorithms was good at discovering hubs. Instead of finding hubs, these algorithms found that the hub and all the nodes it is connected to were all connected, making a large complete subgraph.

# 4 Corrections to an algorithm for recursive thinning

## 4.1 Maximal prime decomposition and minimal triangulation

This chapter presents corrections to a graph-manipulation algorithm known as recursive thinning. First it is necessary to explain minimal triangulation.

Section 3.1 described Bayesian structure-learning of GGMs with the generalized hyper inverse Wishart (G-Wishart) prior distribution on $\Sigma$. In this framework, finding the marginal likelihood of a given graph requires finding its maximal prime decomposition. Olesen & Madsen (2002) is about how to find the maximal prime decomposition of a directed graph. The same process works for undirected graphs, except that one step, "moralization", is omitted.

The first step in finding the maximal prime decomposition is to find a minimal triangulation, which is defined as follows. Let $(V, E)$ be a finite undirected graph. A triangulation of $(V, E)$ is a set of extra edges $T$, often called fill edges, such that $E \cap T = \emptyset$ and $(V, E \cup T)$ is triangulated. As stated in section 2.1, triangulated graphs are the same as decomposable or chordal graphs. A minimal triangulation is one such that removing any edge makes it no longer a triangulation. (Minimal triangulation is not necessary for graphs that are already decomposable, but I am describing the general process.) Minimal triangulations are not the same as minimum triangulations; the latter are triangulations for which there are no triangulations with fewer edges. Finding minimum triangulations is NP-hard (as proved in Yannakakis 1981).

There are numerous algorithms to find minimal triangulations. Heggernes (2006) is a history and survey of these algorithms. She divides them into two main categories, based on two different characterizations of triangulated graphs: (a) they have perfect elimination orders, and (b) every minimal separator is a clique. She gives brief explanations of five or so algorithms in each category.

The first algorithms for minimal triangulation were published in 1976. Two of these take time $O(mn) = O(n^3)$, where $n$ is the number of nodes and $m$ is the number of edges of the untriangulated graph. Algorithms based on the separator-based characterization started to appear in the 1990s, and many more algorithms have appeared since then. Heggernes (2006) makes no mention of graphical models or statistics, except for a cursory citation of a 1988 paper by Lauritzen and Spiegelhalter. The main applications that she mentions are sparse matrix computations (not in a way that is directly relevant to graphical models) and solving systems of sparse linear equations.

Heggernes (2006) also discusses a third class of ways to create a minimal triangulation: create a triangulation that is not necessarily minimal, and then remove excess edges to create a minimal triangulation. In this approach, the usual way to create a triangulation is elimination (which she calls Elimination Game). This works as follows. Put the nodes in some order, the "elimination ordering"; for each node in turn, add edges as necessary to make all the neighbours of the node be connected to each other, and then remove the node and all its incident edges. The triangulation consists of all the edges that are added during this loop. Heggernes gives four algorithms for finding a minimal triangulation by removing excess extra edges from a triangulation that was created using elimination.

There are many possible ways to choose an elimination ordering. It can even be chosen as the algorithm progresses. One popular ordering is "minimum degree", where at each step you choose the remaining node that has the smallest degree (or one of these nodes, if there are more than one). This often creates minimal triangulations straight away, but not always.

For a slightly different use of triangulation in statistics or machine-learning, see Meilă & Jordan (1997).

## 4.2 Recursive thinning

The R package "gRbase" (Dethlefsen & Højsgaard 2005) includes a function called minimalTriang, which performs minimal triangulation. The main argument to this function is the graph for which a minimal triangulation is desired. As an optional argument, a triangulation can be supplied; if it is not, then one is created using a function called triangulate. The main body of minimalTriang is an algorithm that removes excess extra edges from the triangulation to create a minimal triangulation.

The documentation for minimalTriang cites Olesen & Madsen (2002)—this is true as of February 2013, when the most recent version of gRbase was version 1.6-7. The relevant part of Olesen & Madsen (2002) is 2, and the source of the algorithm is cited as Kjaerulff (1993). The relevant part of Kjaerulff (1993) is chapter 1, which is the same as chapters 1 and 2 of Kjaerulff (1990), so I will just refer to the earlier document.

Kjaerulff (1990) and Olesen & Madsen (2002) both call the algorithm "recursive thinning". Only Kjaerulff (1990)'s version of it is recursive, meaning that it calls itself. Olesen & Madsen (2002)'s version uses a Repeat loop and is not recursive, but it is essentially the same. In this chapter I will present non-recursive versions of algorithms, because I think these are easier to understand.

Both Kjaerulff (1990) and Olesen & Madsen (2002) claim that the algorithm works on any triangulation, not just ones created by elimination. However, the algorithm as given in these two publications is not correct, even for triangulations created by elimination. This chapter is concerned with correcting the algorithm for recursive thinning.

The next sections present the incorrect recursive thinning algorithm, Algorithm I, and then two corrected versions, Algorithms II and III, and proofs that these are correct. The R function minimalTriang actually performs Algorithm III, not the incorrect algorithm cited in its documentation. Algorithm II is a simplified version of Algorithm III.

It appears that Kjaerulff (1990) is not well known. Heggernes (2006) states that "In 1996, Blair et al … posed and solved the problem of making a given triangulation minimal by removing edges." The reference is to Blair et al (2001), but this is precisely the problem addressed by Kjaerulff (1990).

## 4.3 Notation

I use a simplified version of the notation in Kjaerulff (1990) and Olesen & Madsen (2002). The list below gives my notation, the notation used in these two papers, and the variable names in the R code for minimalTriang, to make it easy to compare the different versions of the algorithms.

- The given graph is $(V, E)$.
- $T$ is the triangulation. Its initial value is the triangulation that is given as input to the algorithm. (This is called $T$ in Olesen & Madsen 2002 and TT in minimalTriang.)
- $G = (V, E \cup T)$ is the triangulated version of the graph.
- In Algorithm I, $U$ is the set of edges that get removed on this iteration of the Repeat loop. (This is called $T'$ in Kjaerulff 1990 and Olesen & Madsen 2002.)
- In Algorithm III, $R \subseteq T$ is the set of edges that are candidates for removal. Edges are sometimes added to $R$. (This is called $R'$ in Kjaerulff 1990 and Olesen & Madsen 2002, and Rn in minimalTriang.)
- In Algorithm III, $B$ is the set of nodes at the end of edges that have been removed on this iteration of the Repeat loop. (In minimalTriang, exclT is first the set of edges that have been removed on this iteration, and then this set of nodes.)

$T$, $G$, $R$, and $B$ all change during the algorithm. $T$ and $G$ always change at the same time, so it is always true that $G = (V, E \cup T)$.

## 4.4 The incorrect algorithm

Algorithm I is the incorrect algorithm as given in Olesen & Madsen (2002). (Kjaerulff's algorithm starts with $G = (V, E \cup T)$, whereas Olesen & Madsen start with $G = (V, E)$; I think this is a minor oversight in the latter.)

### Algorithm I: an incorrect method for recursive thinning

1. Set $R = T$.
2. Repeat
3. Set $U = \{(x, y) \in R : G(ne(x) \cap ne(y))$ is complete (in $G$)$\}$.
4. Set $T = T \setminus U$ (and update $G$, which is $(V, E \cup T)$).
5. Set $R = \{e_1 \in T : \exists e_2 \in U$ such that $e_1 \cap e_2 \neq \emptyset\}$.
6. Until $U = \emptyset$.
7. Return $T$.

Line 5 sets $R$ to be the set of remaining extra edges that share a node with one or more of the edges that was removed on this iteration of the Repeat loop. Line 9 in Algorithm III does the same thing, though it is written differently. (There is no analogous line in Algorithm II.)

## 4.5 How the incorrect algorithm goes wrong

The fundamental problem with Algorithm I is that it removes more than one edge at a time, instead of updating the graph after each individual edge-removal and checking whether the other edges can still be removed. Algorithms II and III work correctly because they update $T$ and $G$ after each individual edge-removal.

The simplest example of a triangulation for which Algorithm I does not work is Figure 4.1(a), where the solid lines are the edges in $E$ and the dashed lines are the edges in $T$. The algorithm removes both edges at the first step.

Kjaerulff (1990) and Olesen & Madsen (2002) state that the algorithm works on any triangulation, but they seem to have in mind triangulations produced by elimination. The simplest such triangulation for which it does not work is shown in Figure 4.1(b). Any elimination ordering that starts with the node at the bottom would produce this triangulation.

This is not minimum-degree elimination—the bottom node has the highest degree. However, the algorithm can also fail on triangulations created by minimum-degree elimination. An example graph can be constructed as follows. Start with the five-node graph in Figure 4.1(b). For each node except the bottom one, add a clique of 10 nodes that intersects with the original graph only at that node. The new nodes have degree 9, the bottom node still has degree 4, and the other four nodes now have degree 12. Minimum-degree elimination will start with the bottom node and produce the two extra edges shown in Figure 4.1(b) (as well as many others), and Algorithm I will fail.

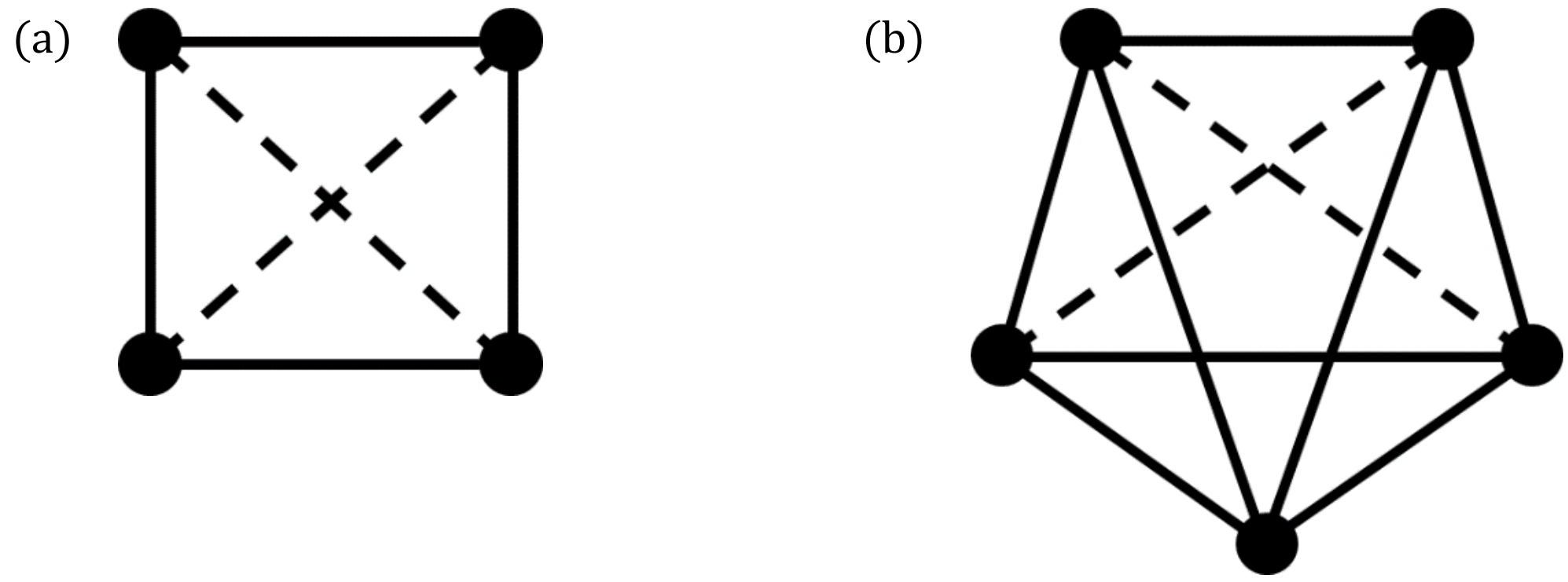

**Figure 4.1.** Two graphs, shown with solid lines, and triangulations of them, shown with dashed lines. (a) A triangulation for which Algorithm I does not work. (b) A triangulation produced by elimination for which Algorithm I does not work.

Similarly, probably any other rule for choosing an elimination ordering will in some cases lead to the failure of the algorithm. For example, the algorithm will fail in any graph where the graph in Figure 4.1(b) appears as an induced subgraph and the bottom node comes first in the elimination ordering.

Incidentally, although Kjaerulff (1990) makes it clear, using unambiguous English and standard notation, that his algorithm checks all the extra edges on the first run, in one example (on pages 11–12) he checks the edges one at a time.

## 4.6 A correct algorithm

### Algorithm II: a correct method for recursive thinning

```
1. Put the edges in T in some arbitrary order.
2. Repeat
3.     For each edge (x, y) ∈ T in turn, in order,
4.         If G(ne(x) ∩ ne(y)) is complete (in G)
5.             Remove (x, y) (from T and G).
6. Until "no edges were removed this time".
7. Return T.
```

### A preliminary result for proving the correctness of Algorithm II

Assuming that $G$ is triangulated, say that the edge $(x, y) \in T$ is "removable" from the current $T$ if removing it does not make $G$ become untriangulated.

***Proposition 4.1.*** $(x, y) \in T$ is removable if and only if the condition in line 4 of Algorithm II is fulfilled.

***Proof.*** Suppose the condition in line 4 is not fulfilled. There must be nodes $a$ and $b$ such that $\{(x,a),(a,y),(y,b),(b,x)\} \subseteq E \cup T$ and $(a,b) \notin E \cup T$. Removing $(x,y)$ would make there be a chordless cycle of length four, $x$–$a$–$y$–$b$–$x$, which would mean that $G$ would no longer be triangulated. So the condition in line 4 is necessary for $(x,y)$ to be removable.

Now suppose the condition in line 4 is fulfilled. Firstly, suppose that removing $(x,y)$ makes there be a chordless cycle of length 5 or more. Then there must have been a chordless cycle of length 4 or more before $(x,y)$ was removed. But $G$ was triangulated, so this is impossible. Secondly, suppose that removing $(x,y)$ causes the appearance of a chordless cycle of length 4, say $x$–$a$–$y$–$b$–$x$, where $(a,b) \notin E \cup T$. This is impossible, because it contradicts the condition in line 4. So removing $(x,y)$ does not lead to the appearance of any chordless cycles of length 4 or more. This shows that the condition in line 4 is sufficient for $(x,y)$ to be removable. ☐

So in Algorithm II, the For loop simply checks each edge in $T$ in turn, and removes the edge if it is removable.

### Proof of correctness for Algorithm II

I will use the word "run" to refer to a single iteration of the Repeat loop. It suffices to prove that (a) the final $T$ is a triangulation, (b) this triangulation is minimal, and (c) the algorithm finishes in finite time.

(a) $G$ is triangulated to start with, by definition. It is only ever modified by the removal of an edge, which happens when the condition in line 4 is fulfilled. $G$ remains triangulated after every such removal, by Proposition 4.1. Therefore $G$ is always triangulated and $T$ is always a triangulation.

(b) On the final run, the algorithm checks the condition in line 4 for every edge in $T$, and finds that it is not fulfilled for any of them. By Proposition 4.1, this means that removing any of the edges in $T$ would make $G$ become untriangulated. In other words, the triangulation is minimal.

(c) Let $t$ be the initial number of edges in $T$. On each run except the last, the Repeat loop removes at least one edge. So the largest number of times that the Repeat loop can be carried out is $t + 1$, which is finite. On each run, the For loop checks all the remaining edges in $T$. On the $i$th run, the remaining number of edges in $T$ is at most $t - i + 1$. This is also finite, so the algorithm is certain to finish in finite time. ☐

## 4.7 A second correct algorithm

### Algorithm III: a second correct method for recursive thinning

1. Put the edges in $T$ in some arbitrary order (minimalTriang uses lexicographic order).
2. Set $R = T$.
3. Repeat
4.     Set $B = \emptyset$.
5.     For each edge $(x, y) \in R$ in turn, in order,
6.         If $G\big(ne(x) \cap ne(y)\big)$ is complete (in $G$)
7.             Remove $(x, y)$ (from $T$ and $G$)
8.             Add $x$ and $y$ to $B$
9.     Set $R = \{(x, y) \in T : x \in B \text{ or } y \in B\}$
10. Until $B = \emptyset$.
11. Return $T$.

### An example of how Algorithms II and III are different

Figure 4.2 shows an example of how Algorithms II and III sometimes do not remove the same edges as each other on every run. On the first run, Algorithm III does not include ①–③ in $R$, so on the second run it misses the chance to remove this edge.

### The intention of Algorithm I

In Algorithm I, the idea of $R$ was that you can save time by not checking edges that you know cannot be removed (Kjaerulff 1990). The same concept is used in Algorithm III, which is a corrected version of Algorithm I. The idea is that on the next run there is no point checking edges for which the sets of neighbours of the two nodes did not change on this run—because even if you check these edges, they will not be removed.

But this idea is mistaken, because the sets of neighbours sometimes change from one iteration of the For loop to the next. And if the sets of neighbours change, then it may become possible to remove the edge. This is illustrated in Figure 4.2. For both algorithms, when the second run starts, the sets of neighbours for ①–③ are the same as in the first run. But in Algorithm II when the For loop gets round to checking ①–③, the neighbours have changed and the edge gets removed.

Figure 4.2 also shows that in Algorithm III it is possible for an edge to be excluded from $R$ but later reappear in it and be removed.

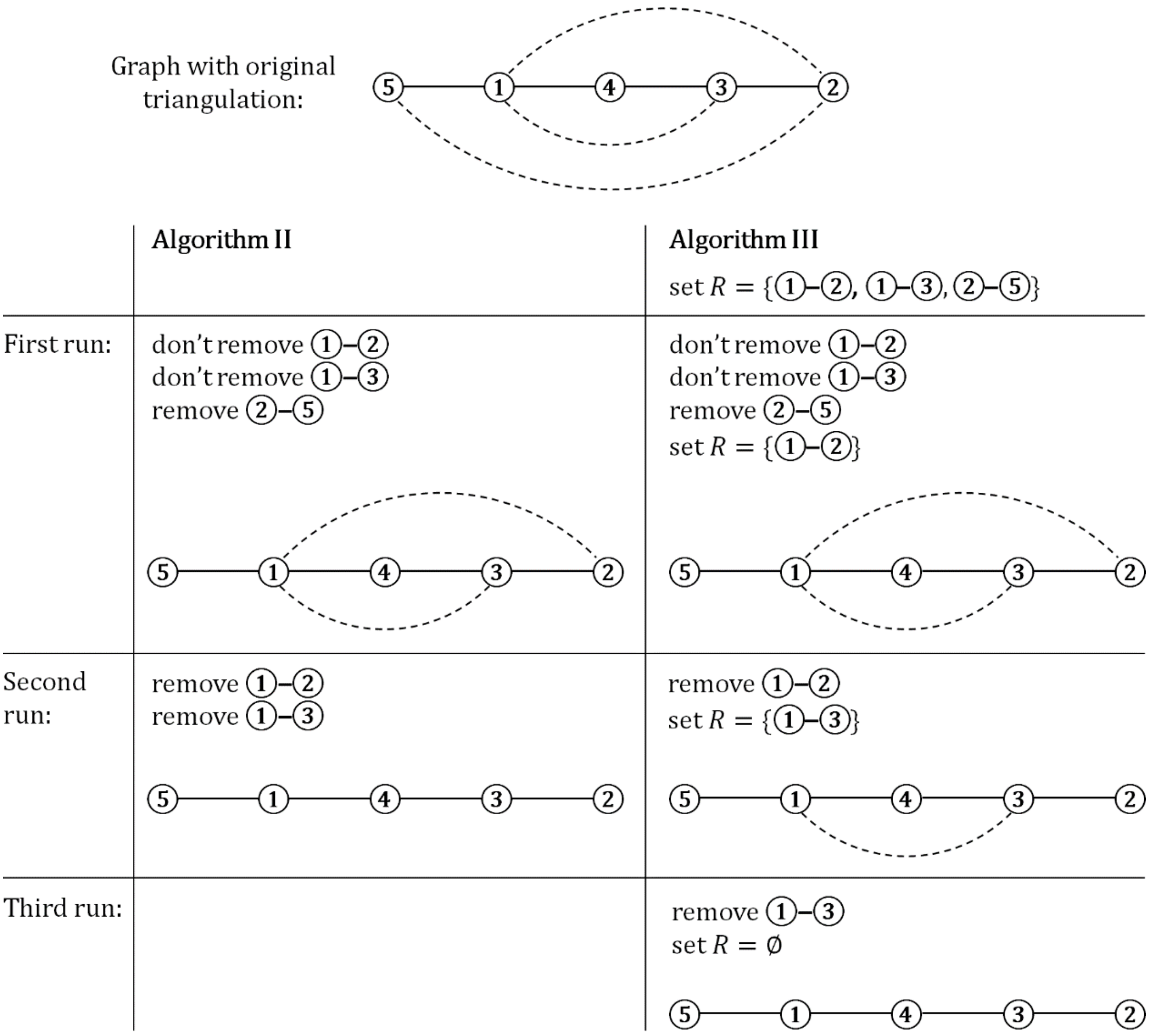

**Figure 4.2.** A graph and a triangulation for which Algorithms II and III do not remove the same extra edges on each run. The solid edges are the graph and the dashed edges are the extra edges. On each run the edges are checked are checked in the order ①–②, ①–③, ②–⑤. Each algorithm also does one final run, which is not shown, in which no edges are removed. (The graph itself is already triangulated, so the triangulation is pointless, but this is just an example for illustration.)

## Proof of correctness for Algorithm III

Again I will use "run" to refer to a single iteration of the Repeat loop.

Algorithm III is the same as Algorithm II, except that on each run Algorithm II checks all the edges in $T$, whereas Algorithm III only checks the edges in $R$, which is always the same as $T$ or a subset of it. When the For loop finishes, $B$ is the set of nodes such that edges incident to them have been removed during the current run. So line 9 has the effect of ensuring that if $(x, y) \in T \setminus R$ then neither $ne(x)$ nor $ne(y)$ changed during the current run, which means that $ne(x) \cap ne(y)$ did not change during the current run. The contrapositive of this is that if $ne(x) \cap ne(y)$ changed during the current run, then $(x, y) \in R$. (The converse is not true—sometimes line 9 puts $(x, y)$ in $R$ even though $ne(x) \cap ne(y)$ did not change during the current run.)

Line 10 in Algorithm III is the same as line 6 in Algorithm II. The equivalence of the other parts of the two algorithms is obvious.

It suffices to prove that on the final run of Algorithm III, the edges that are not checked are not removable. The proof will work by considering an edge $(x, y)$ that does not get checked on the final run, and looking back through the runs to find the last run where it was checked. When $(x, y)$ was last checked, it was obviously found to be unremovable. It will be shown that after that $ne(x) \cap ne(y)$ never changed, from which it follows that $(x, y)$ is not removable during the final run.

Suppose that $(x, y)$ does not get checked on the final run, and say this run was the $i$th. Just after line 9 on the $(i-1)$th run, $(x, y)$ must have been in $T \setminus R$. This means that $ne(x) \cap ne(y)$ did not change during the $(i-1)$th run.

Either (a) $(x, y)$ was checked during the $(i-1)$th run, or (b) it was not. If (a), then $G\big(ne(x) \cap ne(y)\big)$ must have been found to be incomplete, otherwise $(x, y)$ would have been removed from $T$. Since $ne(x) \cap ne(y)$ did not change during this run, even if $(x, y)$ was checked on the final run then $G\big(ne(x) \cap ne(y)\big)$ would still be found to be incomplete (since edges are never added to $G$). In other words, during the final run, $(x, y)$ is not removable.

If (b), then $(x, y)$ must have been in $T \setminus R$ just after line 9 in the $(i-2)$th run. This means that $ne(x) \cap ne(y)$ did not change during the $(i-2)$th run. Either (a2) $(x, y)$ was checked during the $(i-2)$th run, or (b2) it was not. If (a2), then $G\big(ne(x) \cap ne(y)\big)$ must have been found to be incomplete during the $(i-2)$th run. It is now known that $ne(x) \cap ne(y)$ did not change during the $(i-2)$th or $(i-1)$th runs. It follows that $G\big(ne(x) \cap ne(y)\big)$ is still incomplete after line 9 in the $(i-1)$th run, so on the final run $(x, y)$ is not removable.

If (b2), then $(x, y)$ must have been in $T \setminus R$ just after line 9 in the $(i-3)$th run. This reasoning can be continued backwards through the runs. All the edges were checked on the first run, so eventually this search backwards through the runs is certain to find a run where $(x, y)$ was checked and $G\big(ne(x) \cap ne(y)\big)$ was found to be incomplete; and $ne(x) \cap ne(y)$ has not changed since this run—if it had changed, $(x, y)$ would have been put in $R$ and checked on the next run. It follows that $G\big(ne(x) \cap ne(y)\big)$ is still not complete after line 9 in the $(i-1)$th run, so even if $(x, y)$ were checked on the $i$th run it would not be removed. ☐

## 4.8 Comments on the two correct algorithms

Underlying the two proofs is the result, quoted in Heggernes (2006), that "if $G \subset H$ for two chordal graphs $G$ and $H$ on the same vertex set, then there is a sequence of edges that can be removed from $H$ one by one, such that the resulting graph after each removal is chordal, until we reach $G$." This explains why it is sensible to remove one edge at a time.

The edges can be checked in different orders on different runs. The proofs make no assumptions about these orders. However, it is natural in writing a computer program to make it check the edges in the same order on every run.

The R function minimalTriang performs Algorithm III plus various checks. For example, it checks whether $(V, E)$ is triangulated at the start and whether $G$ is triangulated at the end. The test in line 10 of Algorithm III is actually done before line 9, and if the result is true then the process breaks out of the Repeat loop.

## 4.9 Which of the correct algorithms is faster?

In the example in Figure 4.2, Algorithm II is faster than Algorithm III. Algorithm II can also be faster for a triangulation produced by minimum-degree elimination. An example can be constructed from the original graph in Figure 4.2 (the graph with the four solid edges). Add two nodes that are connected to each other and node 5, and add two nodes that are connected to each other and node 2. One possible minimum-degree elimination begins by eliminating nodes 4, 3, and 1 and creates the same three extra edges as in Figure 4.2. Algorithms II and III proceed in the same way as in Figure 4.2, and Algorithm II is faster.

On the other hand, Algorithm III is sometimes faster than Algorithm II, even with triangulations produced by minimum-degree elimination ordering. Again consider the graph and the triangulation shown at the top of Figure 4.2, or the ones described in the previous paragraph. But this time suppose the edges are checked in the order 1–3, 1–2, 2–5. On the first run both algorithms remove 2–5, on the second run they both remove 1–2, and on the third run they both remove 1–3. But on the second run, Algorithm III saves time by not checking 1–3. This makes it faster overall, assuming that creating $B$ and $R$ does not take any time.

Kjaerulff (1990) recommends checking the edges in the reverse of the order in which they were added during the elimination-ordering algorithm. It is not clear whether one of the two correct algorithms is always faster than the other if this advice is followed.

I carried out an experimental comparison of the two correct algorithms in R, using simplified versions of minimalTriang. To test the two algorithms it is necessary to create triangulations. The R function triangulate uses a version of minimum-degree elimination ordering that very often creates a minimal triangulation straight away, so that the recursive thinning algorithm has nothing to do. So I wrote a simpler triangu-lation function that does elimination with the nodes in their natural order. The idea was that this would create non-minimal triangulations more often.

I created 100,000 random non-decomposable graphs on 30 nodes, by choosing them uniformly at random from among all such graphs, with replacement. I then created triangulations of all these graphs and ran the two programs on the graphs and their triangulations. Algorithm II took 767 seconds of CPU time and Algorithm III took 830 seconds.

## 4.10 What is the best algorithm for minimal triangulation?

Heggernes (2006) reports that the fastest known algorithm for minimal triangulation is $o(n^{2.376})$, where $n$ is the number of nodes. This algorithm appears in Heggernes et al (2005). The fastest algorithms are rather complicated, and their high asymptotic speeds rely on the detailed manipulations being done in specially fast ways.

If a fast algorithm was wanted, my personal choice would be MCS-M (Berry et al 2004). This has asymptotic speed $O(mn)$, where $m$ is the number of edges of the untriangulated graph. MCS-M is not especially simple, since it requires searching along paths where the nodes fulfil a certain condition. This is more complicated to program than merely checking whether the neighbours of a node are connected, which is how Algorithms II and III work. Unless speed is paramount, it seems sensible to use minimum-degree elimination followed by Algorithm II or III to remove excess edges, especially as minimum-degree elimination often produces minimal triangulations straight away.

R is much slower than general-purpose programming languages, so there would be no point in rewriting minimalTriang to use an algorithm that is theoretically or asymptotically superior. If speed was important, it would be more sensible to rewrite triangulate or the main body of minimalTriang in C or Fortran and call these from R, or abandon R and use a different programming language.

# 5 Random graph distributions

## 5.1 Two ways of looking at graph distributions

Bayesian structure-learning of graphical models involves probability distributions on sets of graphs. One of the first steps for the user is to specify a prior distribution that accords with their beliefs about which graphs are more or less likely. This chapter is about probability distributions on graphs and in particular about prior distributions in Bayesian structure-learning.

There are two ways of looking at or defining probability distributions on sets of graphs. The first is that you have a set of graphs and a formula that can be applied to any of these graphs to give its probability, or its unnormalized probability. Obviously the probabilities are all non-negative and sum to 1. This I will call a "graph distribution". The second is a "random graph model", which is essentially a random or partly random procedure for constructing a graph.

Any probability distribution on a set of graphs can be defined in either of these two ways. But in practice the two ways of looking at these distributions are different. If you are given a graph distribution, it may be difficult to generate a graph from it. Conversely, if you are given a random graph model, it may be difficult to calculate the probability of a given graph.

If you have a graph distribution, MCMC can be used to generate a sample of graphs that approximately follow the distribution. The acceptance probability for moving from $G$ to $G'$ is

$$\min\left\{1,\ \frac{p(G')}{p(G)}\frac{q(G' \to G)}{q(G \to G')}\right\},$$

where $p$ is the graph distribution and $q(G_1 \to G_2)$ is the probability of proposing to move to $G_2$ if the current graph is $G_1$. In principle, the proposal distribution $q$ can be chosen arbitrarily as long as the Markov chain is irreducible and aperiodic.

When dealing with prior distributions for graphical model structure-learning, it is necessary to calculate the probability of a given graph, so it is natural to work with graph distributions rather than random graph models. However, you might also want to be able to generate from the distribution, for example to empirically evaluate whether it encourages hubs.

Graph distributions have been the subject of some research in the context of graphical model structure-learning. But random graph models have been the subject of far more research, in other contexts, as described in the next two sections. In this chapter, the number of nodes in the graph is $n$.

## 5.2 Erdős–Rényi random graphs

The first random graph models to be studied in depth were Erdős–Rényi graphs. These two models will be referred to in several sections of this chapter and in section 6.2. The first Erdős–Rényi model is $G(n,p)$, in which there are $n$ nodes and each edge appears independently with probability $p$, and the second is $G(n,M)$, where there are $n$ nodes and $M$ edges and all such graphs have equal probability. $M$ and $p$ are usually $M(n)$ and $p(n)$, functions of $n$. The first Erdős–Rényi model was introduced in Gilbert (1959) and the second was introduced in Erdős & Rényi (1959).

Erdős–Rényi random graph theory is covered in depth by Bollobás (2001). It is mainly concerned with approximating the proportion of graphs that have a certain property and seeing what happens as $n \to \infty$. In many cases, either almost every graph has the property (in other words, the proportion of graphs with the property tends to 1 as $n \to \infty$) or almost every graph does not have the property. The preface of Bollobás (2001) says that the main omission from this book is probably random trees, which are covered in chapter 7 of Moon (1970).

The notations $G(n,p)$ and $G(n,M)$ may seem ambiguous, because the second parameter has two possible meanings, but it is uncommon to write specific numbers or formulas inside the brackets. Alternative notations include $\mathcal{G}_p$, $\mathcal{G}(p)$, $\mathcal{G}_{n,p}$, and $\mathcal{G}_{n,m}$. $G(n,p)$ is sometimes called the Bernoulli random graph or the binomial model, and $G(n,M)$ is sometimes called the uniform model (Janson et al 2000, page 2). For many questions, results about these two different models are very similar. The two models are in certain senses equivalent, as shown by Theorem 2.2 in Bollobás (2001) and a stronger result in Łuczak (1990).

Erdős–Rényi graphs are clearly random graph models as defined in section 5.1. However, given an Erdős–Rényi model, it is also easy to calculate the probability of any given graph. So they could also be regarded as graph distributions.

## 5.3 Complex networks

"Random graphs" is sometimes taken to mean Erdős–Rényi random graphs. An example of this usage is in Watts & Strogatz (1998). In Erdős–Rényi graphs, the node degrees follow an approximate Poisson distribution, which means that most nodes have similar degrees (Barabási & Oltvai 2004, Jeong et al 2000). Since the late 1990s a consensus has emerged that these graphs are usually unsuitable for modelling networks in the real world.

Random graph models that are intended to model real-world networks have come to be known as "complex networks". Some research on complex networks is not mathematically rigorous and instead demonstrates properties by means of experiments on computer. Examples of this type of research are Watts & Strogatz (1998) and Barabási & Albert (1999), which have both been highly influential. The range of random graph models or complex networks that have been proposed, studied, and used is discussed in depth in Newman (2003).

One random graph model discussed in Newman (2003) is the "scale-free" graphs of Barabási & Albert (1999). These are described in section 2.4, about biomolecular networks.

Another random graph model that is currently the subject of research is the **configuration model**. This starts with a fixed degree for each vertex. A random graph is generated by creating the appropriate number of half-edges for each node, and then joining the half-edges in pairs uniformly at random. Doing this can lead to self-edges and multi-edges, but asymptotically the proportion of these is small (Molloy & Reed 1995).

The configuration model is used to model social networks or networks of human contact. These networks are then used for modelling the spread of epidemics—see for example Andersson (1998) or Britton et al (2007, 2011). The configuration model has been used to prove asymptotic mathematical theorems about graphs that are chosen uniformly at random from among all those that have a given degree sequence (Molloy & Reed 1995).

Similar to the configuration model is the **expected-degree model** (Chung & Lu 2002a,b, 2006; Chung et al 2003), in which each node $v_i$ has a weight $w_i$, and the edge $(v_i, v_j)$ is present with probability $w_i w_j / \sum w_k$, independent of all the other edges. If self-edges are permitted then $\mathbb{E}(\deg(v_i)) = w_i$. For this model, the probability of a given graph can easily be calculated. The expected-degree model is an example of a factored distribution—see the next section.

## 5.4 Factored distributions

### Definitions

This section is about a certain class of graph distributions that appears in several contexts. I will refer to these as "factored" distributions (following Meilă & Jaakkola 2006). The number of nodes is fixed. A factored distribution on a set of graphs is one where each edge has a weight, $w_e$, and the probability of each graph is proportional to the product of the weights of the edges in that graph. In symbols,

$$\mathbb{P}(G) \propto \prod_{e \in E_G} w_e \, , \tag{1}$$

for $G \in \mathcal{G}$, where $\mathcal{G}$ is the set of graphs under consideration; and $\mathbb{P}(G) = 0$ for $G \notin \mathcal{G}$. It will be useful later to write the definition with an equals sign:

$$\mathbb{P}(G) = \frac{\prod_{e \in E_G} w_e}{\sum_{H \in \mathcal{G}} \prod_{e \in E_H} w_e} \, . \tag{2}$$

Let $E_{all}$ be the set of all $\binom{n}{2}$ possible edges. Any distribution where

$$\mathbb{P}(G) \propto \prod_{e \in E} p_e \prod_{e \in E_{all} \setminus E} (1 - p_e)\,, \tag{3}$$

for some $\{p_e\}$, is also factored, since this expression can be written as

$$\prod_{e \in E} \frac{p_e}{1 - p_e} \prod_{e \in E_{all}} (1 - p_e) \propto \prod_{e \in E} \frac{p_e}{1 - p_e} = \prod_{e \in E} w_e\,,$$

where $w_e = p_e/(1 - p_e)$ (which is the odds that $e \in E$ in the case described in the next subsection).

## The set of all graphs

Let $\mathcal{G}_{all}$ be the set of all $2^{\binom{n}{2}}$ graphs. If $\mathcal{G} = \mathcal{G}_{all}$, then factored distributions are ones where each edge is present or absent with a fixed probability and all these events are independent. Moreover, $p_e$ is the probability that $e \in E$ and $w_e$ is the odds of the same event. To see these facts, use definition (3), and note that

$$\sum_{G \in \mathcal{G}_{all}} \left\{ \prod_{e \in E_G} p_e \prod_{e \in E_{all} \setminus E_G} (1 - p_e) \right\} = \prod_{e \in E_{all}} \big(p_e + (1 - p_e)\big) = 1\,.$$

It follows that the proportional-to symbol in definition (3) can be replaced by an equals sign:

$$\mathbb{P}(G) = \prod_{e \in E} p_e \prod_{e \notin E} (1 - p_e)\,.$$

This is essentially the definition of the presence or absence of each edge being independent.

## Trees and forests

If $\mathcal{G}$ is the set of forests or trees, then in a factored distribution the edges are not present or absent independently of each other, because the graph is constrained to be a forest or tree. This was pointed out by Meilă & Jaakkola (2006).

If $\mathcal{G}$ is the set of trees, then the products in the numerator of (2) all have the same number of terms (namely $n - 1$). So

$$\mathbb{P}(G) = \prod_{e \in E_G} \frac{w_e}{\left(\sum_{H \in \mathcal{G}} \prod_{e \in E_H} w_e\right)^{1/(n-1)}}\,.$$

This now has the very simple form $\mathbb{P}(G) = \prod_{e \in E} w_e$. (To convert to this form, replace each of the original $w_e$'s with $w_e/(\sum_{H \in \mathcal{G}} \prod_{e \in E_H} w_e)^{1/(n-1)}$.) However, the form with the proportional-to symbol is more natural, since this is how factored distributions arise as prior distributions that are inferred from expert knowledge, as suggested by Madigan & Raftery (1994), and as posterior distributions.

### Uses of factored distributions

Factored distributions are graph distributions, rather than random graph models. However, in the case that $\mathcal{G} = \mathcal{G}_{all}$, it is easy to generate from them, since all the edges are independent, so they could be regarded as random graph models.

The use of factored distributions as prior distributions for graphical model structure-learning seems to have been first proposed by Madigan & Raftery (1994). They suggest getting an expert to estimate the probability of each edge being present and assuming that the presences of the edges are mutually independent.

The results in Meilă & Jaakkola (2006) are all about factored distributions on trees. For Bayesian structure-learning of discrete-valued tree graphical models, they prove that under certain assumptions the graph posterior is a factored distribution. Chapter 8 describes methods for analyzing factored distributions for trees, based on Meilă & Jaakkola (2006), and methods for generating from these distributions. Section 7.4 describes how the Chow–Liu algorithm can be used if the prior is a factored distribution on trees.

## 5.5 Graph priors that have been proposed

### Priors for undirected graphs

In Bayesian structure-learning it is necessary to be able to calculate the probability of a given graph, so in practice prior distributions are invariably defined as graph distributions, rather than random graph models.

The simplest type of graph prior distribution is the uniform distribution, where each of the graphs under consideration is equally likely. Priors of this type have been used by Cooper & Herskovits (1992), Madigan & Raftery (1994), Giudici (1996), Giudici & Green (1999), Roverato (2002), Atay-Kayis & Massam (2005), Dobra et al (2011), Wang & Li (2012), and others. If you are considering all graphs, or all decomposable graphs, then the uniform distribution gives higher probability to medium graph sizes—the "size" of a graph is the number of edges it has—than to small or large sizes (Giudici & Green 1999, Jones et al 2005, Carvalho & Scott 2009, Armstrong et al 2009).

For undirected graphical models, several alternatives have been proposed in published research. One is the "size-based prior" of Armstrong et al (2009). In this distribution, non-decomposable graphs have probability zero, all sizes are equally likely, and all decomposable graphs of the same size are equally likely. They also propose a more general hierarchical prior distribution in which the size has a binomial distribution,

$$\mathbb{P}(\text{size} = k) = \binom{\binom{n}{2}}{k} \psi^k (1-\psi)^{\binom{n}{2}-k} \quad \text{for } k = 0,1,\dots,\binom{n}{2},$$

the binomial parameter $\psi$ has a beta distribution, and again all decomposable graphs of the same size are equally likely. If $\psi < 0.5$ then more probability is given to sparser graphs.

Part of the motivation for the size-based prior distribution was the belief that the graph is sparse. But the prior fails to reflect this belief in a sensible way, because it does not take account of how many graphs there are that have each size. For example, it often gives lower probability to each of the graphs of size $\binom{n}{2} - 1$ than to the complete graph. More probability is assigned to the size $\binom{n}{2} - 1$, but it gets shared out among many graphs. Any sensible "sparsity-encouraging" prior would surely give higher probability to any graph of size $k - 1$ than to any graph of size $k$, especially in the case $k = n$. Consequently it does not seem sensible to assign a probability to a size without taking into account how many graphs have that size.

Dobra et al (2004) and Jones et al (2005) use $\mathbb{P}(G) = \beta^{|E_G|}(1-\beta)^{\binom{n}{2}-|E_G|}$, where $\beta \in [0,1]$. When all graphs are being considered, this is the first Erdős–Rényi graph model, where each edge is present independently with probability $\beta$. When only decomposable graphs are being considered, it is not. Jones et al (2005) call this the Bernoulli prior but I will call it the "binomial prior". Carvalho & Scott (2009) say that this prior distribution is "rapidly becoming the standard," and they use an adaptation of it in which there is a hierarchical prior distribution on $\beta$.

Bornn & Caron (2011) propose a class of priors for decomposable graphs that are calculated using the cliques and separators. The main one they suggest is

$$\mathbb{P}(G) \propto \frac{\prod_C a(|C|-1)!}{\prod_S b(|S|-1)!}.$$

The product in the denominator is not over the collection of separators, as in section 3.1, but over the collection of non-empty separators. (Their more general prior has functions $\psi_C(C_j)$ and $\psi_S(S_j)$ in the numerator and denominator, "with the convention that $\psi_S(\emptyset) = 1$," but the rest of the paper makes it clear that $S_j$ is never $\emptyset$.) The parameters $a$ and $b$ can be adjusted to encourage or discourage cliques and non-empty separators respectively. The main aim of these priors is to express the belief that the nodes should be clustered in cliques, especially non-overlapping cliques, rather than spread out in long lines.

Thomas et al (2008) use undirected graphical models to analyze residue positions in proteins. They describe a "contact graph prior", which only permits edges between pairs of residue that are within a certain physical distance of each other.

As mentioned in section 5.4, Madigan & Raftery (1994) proposed factored priors, where each edge has a fixed probability of appearing in the graph and the presences of all the edges are independent. These can be used for both undirected graphical models and directed acyclic ones.

## Priors for DAGs

For directed acyclic graphs, Heckerman et al (1995) assume that the user can express their prior beliefs in the form of a single graph. They assign prior probabilities to graphs by penalizing them according to the number of edges that are different compared to the user's graph. Buntine (1991) describes how to convert an expert's beliefs on the probability of each edge into a prior distribution. Heckerman et al (1997/1999/2006) use a uniform prior.

Mukherjee & Speed (2008) propose a more elaborate class of graph priors based on the properties that the graph is believed to have. These priors are of the form

$$\mathbb{P}(G) \propto \exp\left(\lambda \sum_i w_i f_i(G)\right),$$

where each $f_i$ is a “concordance function” that increases as $G$ matches the prior belief more closely, and the $w_i$ are weights. By using several $f_i$’s it is possible to combine several prior beliefs. They give possible $f_i$’s for several types of prior belief, for example that certain edges are likely to be present or absent, that edges between two groups of vertices are unlikely (which they say is common in molecular biology), or that the nodes are unlikely to have many edges into them. They also give an $f_i$ for the belief that the degree distribution is likely to be scale-free (see section 2.4). Also given are three references (19–21) that discuss informative priors for biomolecular networks.

Chapters 4 and 5 of Byrne (2011) are about “structural Markov properties” for decomposable undirected graphical models and DAG graphical models. These are properties of graph distributions that are analogous to the standard Markov properties and the hyper and meta Markov properties from Dawid & Lauritzen (1993). The basic idea is that two components of the graph are conditionally independent given a separating component. Graph distributions that have the structural Markov property are conjugate priors in certain situations.

## 5.6 Graph priors based on random graph models

This section is about what kind of prior distribution should be used for Bayesian learning of graphical model structure. Of course the graph prior should encapsulate the researcher’s prior beliefs about the graph structure.

These beliefs for biomolecular networks were discussed in section 2.4. A good graph prior for biomolecular networks might give high probability to sparse graphs, encourage structures such as hubs and cliques, or induce an approximate power law on the node degrees. The presence or absence of all of these features except cliques can be assessed by looking at a graph’s degree sequence. (To put it mathematically, the presence or absence of these features can be expressed as a function of the degree sequence.) For example, the question of whether there are any hubs is simply about whether there are any degrees that are much higher than the others. Conversely, if you are free to choose the degree sequence then you can choose it so as to encourage or discourage most of these features. An alternative way to enforce sparsity is to consider only forests or trees.

In any case it seems sensible for the degree sequence to be the main feature of graphs that is used in specifying the prior distribution, or the sole feature that is used. One possible random graph model is the configuration model, described in section 5.3. For this, you have to specify the degree of each node. Sparsity could be enforced by simply choosing low degrees, or low total degree. The configuration model might be appro-priate if you thought that a specific node was a hub and none of the others were. But it cannot be used to express the belief that an unspecified node is a hub, or the belief that each node has a small but positive probability of being a hub.

Using the configuration model leads to several complications. It generates not simple graphs but configurations, which may include multiple edges and self-edges. Multiple edges would have to be replaced with single edges, and self-edges would have to be discarded, so the eventual degree of each node might be lower than intended. Chapters 6–11 are about forest and tree graphical models. In the configuration model, if the degrees are suitably low then the graph is likely to be a forest, so you could generate forests by choosing low degrees and rejecting any graphs that were not forests. However, there does not seem to be an easy and practical way to generate trees from the configuration model.

In connection to the configuration model the question arises of whether a given set of numbers is a possible degree-sequence. For this question see Theorem 6.4 in section 6.2.

Perhaps the biggest drawback of the configuration model is that, using the terminology from section 5.1, it is a random graph model rather than a graph distribution, and calculating the probability of a given graph is difficult. The probability of any given configuration is $2^{D/2}(D/2)!/D!$, where $D = \sum_{v \in V} \deg(v)$. But the probability of a given graph is more complicated, and if you are restricting to forests or trees then it is more complicated again.

The expected-degree model, also described in section 5.3, has the advantages that it is a factored distribution, the probability of a given graph is easy to calculate (since for each edge there is a simple formula for the probability that it is present), and it is more flexible than the configuration model in the sense that the degree of each node is not specified exactly. But it is still no use for the scenario where you believe that an unspecified node is a hub.

The configuration model or expected-degree model could be adapted by using a hierarchical distribution. For example,

$$\begin{aligned}
&\mathbb{P}(v \text{ is a hub}) = \theta \\
&\deg(v) \mid (v \text{ is a hub}) \sim Poisson(\lambda_1) \\
&\deg(v) \mid (v \text{ is not a hub}) \sim Poisson(\lambda_2).
\end{aligned}$$

Obviously $\lambda_1 > \lambda_2$. To generate from this model, you would choose whether each node is a hub according to the Bernoulli distribution with parameter $\theta$, then choose the degree of each node according to the appropriate Poisson distribution, and finally generate the graph according to the configuration model or the expected-degree model.

This still suffers from the drawback that it is complicated and inelegant to calculate the probability of a given graph, since you have to sum over all the $2^n$ possibilities of which nodes are hubs and then the range of the Poisson distribution, which is the non-negative integers. You could decide as *a priori* knowledge that there is only one hub, in which case the first sum would only have $n$ terms; or when exploring the graph posterior distribution (see section 10.1) you could have a separate “move” that consists of deciding anew which nodes were hubs, in other words resampling from the $Bernoulli(\theta)$ distribution for each node. But there are still the problems of multiple edges and self-edges, and the complications of restricting to forests or trees if that is required.

Instead of having a separate Bernoulli distribution for each node, you could have

$$\mathbb{P}(v_{i_1}, \dots, v_{i_m} \text{ are hubs, and the other nodes are not})$$

$$= \binom{n}{m}^{-1} \text{ for all } \{i_1, \dots, i_m\} \subseteq \{1, \dots, n\}.$$

Here $m$ nodes are chosen to be hubs, and these are chosen uniformly at random from all the nodes. With this distribution it would be simpler to calculate the probability of a given graph, since there are only $\binom{n}{m}$ possibilities to sum over.

Another alternative would be to use a Pareto distribution to choose the degrees. Pareto distributions are long-tailed. Most values are small but there is some chance of getting a large value.

## 5.7 Practical graph prior distributions

In this section I propose seven criteria for a graph prior. I explain how the priors in section 5.5 fail to satisfy these criteria and propose one possible prior that does fulfil them. In section 11.6 I will use this prior in experiments to see whether it gives better results than the uniform graph prior.

The seven criteria, listed below, are intended to reflect the beliefs that the graph is sparse and some nodes are likely to be hubs, but it is not known which ones. They also include certain criteria that are useful in practice.

1. There has to be an explicit formula for the probability of any given graph. This probability can be unnormalized, because Bayesian structure-learning usually produces unnormalized posterior probabilities anyway.
2. It is desirable that there be a computationally efficient method for generating from the prior distribution, as with random graph models. This would enable you to check whether the prior produces graphs that look right and accord with your beliefs. However, this is not as important as criterion number 1.
3. The formula should give higher probabilities to sparse graphs. The precise meaning of this criterion is deliberately not specified, and it is not needed if you are restricting attention to trees or forests.
4. The formula should give higher probabilities to graphs with hubs, and higher probabilities to graphs with hubs that have larger degrees. The precise meaning of this criterion is deliberately not specified.
5. The formula should be a function of the unordered degree sequence, or equivalently the sorted degree sequence. This means the prior is symmetric in the nodes and the degrees are exchangeable. It corresponds to not knowing which nodes are hubs. The reason for this criterion is that, as mentioned in section 5.6, the beliefs that the graph is sparse and some unspecified nodes are likely to be hubs can be expressed as beliefs about the unordered degree sequence.

6. If only trees are under consideration, then it is desirable but not essential for the prior to be a factored distribution, because there are various fast methods for analyzing factored distributions on trees.
7. All other things being equal, the formula should be simple, because this will make it easier to understand and work with.

None of the graph prior distributions in section 5.5 satisfy these criteria fully. The size-based prior of Armstrong et al (2009) and the binomial prior of Jones et al (2005) and others do not fulfil criterion 4, about hubs—they would give the same probability to a graph with 100 edges in which all the degrees are below 5 as to a graph with 100 edges in which one node had degree 50. Bornn & Caron (2011)'s general class of priors might include ones that fulfil criterion 4, but it is difficult to see how, and such priors would not fulfil criterion 7. The class of priors proposed by Mukherjee & Speed (2008) certainly does contain priors that fulfil criteria 1–5, but actually creating one of these would be tantamount to inventing a prior from scratch, because their class of priors is so broad.

I will describe one possible graph prior distribution, which I will call the "hub-encouraging prior". This has two parameters, $\chi \in \mathbb{Z}^+$ and $\psi \in \mathbb{R}^+$. Given a graph, subtract $\chi$ from all the degrees, and retain only the positive ones. The probability of the graph is proportional to the sum of these values plus $\psi$. In symbols,

$$\begin{aligned}\mathbb{P}(G) \;&\propto\; \psi + \sum_{v \in V} \max\{0, \deg(v) - \chi\} \\ &= \psi + \sum_{v:\, \deg(v) > \chi} (\deg(v) - \chi)\,.\end{aligned}$$

The idea is that a node is regarded as a hub if and only if its degree is greater than $\chi$. All graphs that have no hubs are equally likely, and any graph that has a hub is more likely than any graph that does not. A hub contributes more if its degree is higher. How much hubs affect a graph's probability also depends on $\psi$—the larger $\psi$, the smaller the effect. Obviously $\chi$ should be a reasonably large positive integer, for example 10, and $\psi$ should be positive so that even graphs with no hubs still have positive probability. The range of unnormalized probabilities is $[\psi, \psi + p - 1 - \chi]$, assuming that the graphs under consideration include at least one that has no hubs and at least one in which one node has the maximum possible degree.

This distribution is not easy to generate from, but its simplicity means that it is easy to interpret and understand. For example, if $\psi = 10$ then all graphs in which no node has degree greater than 10 have the same probability. If one wanted to generate graphs from this distribution and look at them, MCMC could be used to generate from it approximately, or a rejection or importance-sampling method could be used to sample from it exactly, though this might be cumbersome or slow.

As described in chapters 7 and 8, there are numerous useful methods that can be used to analyze factored distributions on trees. Unfortunately, priors that fulfil criteria 4–5 cannot be expressed as factored distributions. Consider all the trees that contain the edges $(v_2, v_3), (v_2, v_4), \dots, (v_2, v_n)$. Only one edge is unspecified, and this edge must include $v_1$. According to criterion 4, the tree with $(v_1, v_2)$ ought to have higher probability than the tree with $(v_1, v_3)$, because $v_2$ is already a hub. So in a factored distribution, $w_{(v_1,v_2)}$, the factor for $(v_1, v_2)$, would have to be higher than $w_{(v_1,v_3)}$, the factor for $(v_1, v_3)$. But a similar argument about a different set of trees shows that $w_{(v_1,v_3)}$ has to be higher than $w_{(v_1,v_2)}$, which is impossible. On the other hand, the belief that a particular set of nodes are hubs can be expressed as a factored distribution—the weights on the edges from those nodes should simply be high.

# 6 Forest and tree graphs and graphical models

## 6.1 Why consider forest and tree graphical models?

### Preamble

Forests are graphs that have no cycles, and trees are connected forests. This definition of trees makes it sound as though they are bigger than forests. In graphical models this is appropriate, since the set of nodes is usually fixed, and so trees have more edges than forests (except for forests that are themselves trees). The alternative and equivalent definition of forests is that they are graphs whose connected components are all trees. In the literature of machine learning, where a lot of research on graphical models appears, forests are sometimes referred to as trees (Meilă & Jaakkola 2006, Bradley & Guestrin 2010, Bach & Jordan 2003).

In Bayesian learning of GGM structure, it is common to restrict attention to decomposable graphs (in other words, to set the prior probability of all non-decomposable graphs to zero), because the marginal likelihoods of these graphs can be calculated exactly using the explicit formula from section 3.1. It is also possible to restrict attention even further, to forests or trees, by setting the prior probability of all other graphs to zero. All forests and trees are decomposable.

Forests and trees are very restricted classes of graphs, and no doubt these graphs are too simple to be realistic models of biological or other networks, as mentioned in Edwards et al (2010). But there are several reasons why it might be sensible and desirable to consider only forests or trees. These reasons are the subject of this chapter.

### Computational tractability

One of the main reasons for restricting attention to forests or trees is that they are much more computationally tractable than general or even decomposable graphs. This is essentially because the joint density factorizes in terms of marginal densities on nodes and pairs of nodes. Viewing forests or trees as decomposable graphs, the cliques are the edges and the separators are the individual nodes. (In unconnected forests, the separators also include the empty set, but this can be ignored because it contributes a factor of 1 to the density and other quantities.) The multiplicity of each non-empty separator is the degree of that node minus one. So the factorization of the joint density using cliques and separators is

$$p(x) = \frac{\prod_{(u,v)\in E} p(x_u, x_v)}{\prod_{v\in V} p(x_v)^{\deg(v)-1}} = \prod_{v\in V} p(x_v) \prod_{(u,v)\in E} \frac{p(x_u, x_v)}{p(x_u)p(x_v)},$$

and the likelihood given $m$ independent and identically distributed observations is $\prod_{i=1}^{m} p(x^i)$.

For graphical-model structure-learning, the algorithm of Chow & Liu (1968), described in chapter 7, gives the maximum-likelihood tree in time that is polynomial in the number of nodes. Let the number of nodes be $n$. The time taken is $O(n^2 \log n)$ according to Acid et al (1991), Eaton & Murphy (2007), and Meyer et al (2007). But these papers either cite no sources for this claim or cite sources that do not make the claim. The only publication I have found that actually calculates the asymptotic time is Meilă (1999), which proves that it is $O(n^2(m + \log n))$, where $m$ is the number of observations. The first term is for calculating the edge-weights, and the second term is for doing Kruskal's algorithm. Goldberger & Leshem (2009) state that the time taken is $O(n^2)$ if Prim's algorithm is used.

For general graphs, structure-learning is believed to be computationally intractable. For example, Anandkumar et al (2012) and Tan et al (2010b) both assert that structure-learning of general graphs is NP-hard. However, it is not clear exactly what they mean, since the maximum-likelihood graph is always the complete graph. Both papers cite Karger & Srebro (2001), which shows only that finding the maximum-likelihood decomposable graph with bounded clique-size is NP-hard. Anandkumar et al (2012) also cite Bogdanov et al (2008), which is about the case where the node-degrees are bounded.

The other major computational task with graphical models is inference—finding the marginal distributions on one set of nodes, given data on another set. This is also fast on trees. For general graphs, inference is done using the junction-tree algorithm (Lauritzen & Spiegelhalter 1988), whose running time is exponential in the size of the largest clique. But for trees and forests, the junction-tree algorithm simplifies to the sum-product algorithm, also called belief propagation, which takes time proportional to the number of edges (Pearl 1988, sections 4.2–4.3).

## Sparsity

Another key justification for restricting attention to forests or trees is that biomolecular networks are sparse and "sparse graphs are locally tree-like". A detailed discussion and investigation of this notion appears in section 6.2. This is the second type of sparsity that arises in graphical model structure-learning (the first type of sparsity is the number of variables or nodes being much greater than the number of observations).

## Informal justifications

For structure-learning of biomolecular networks, Edwards et al (2010) express the belief that forests can give some idea of the structure and be useful in several ways. Firstly, they suggest that if you select a forest then this can be used as the initial model in a search algorithm through a wider space of graphs, for example decomposable graphs. For algorithms that examine decomposable graphs, it is not usually possible to start from graphs produced by procedures such as the graphical lasso of Friedman et al (2007), since these are not guaranteed to be decomposable. A suitable forest can be found quickly using the Chow–Liu algorithm, described in section 7.1.

Edwards et al (2010) also suggest regarding some of the properties of the selected forest as properties of the true graph. For example, you could assume that the true graph has the same connected components as the forest—in graphical models, separate connected components are marginally independent. If there is more than one component, this will reduce the dimension of the problem, which is a great advantage for computational efficiency in multivariate statistics. Edwards et al (2010) also claim that analysis of forests could be used to identify hubs, which are one of the most important features of biomolecular networks.

### Tree and forest graphical models in use

Tree and forest graphical models have been used for a wide variety of applications. Kundaje et al (2002) uses trees because they can be learnt quickly and are appropriately sparse for time-series gene regulation networks. Costa et al (2008) use tree GGMs to model gene expression levels at different stages of cell differentiation. Each tree corresponds to a group of genes, each node corresponds to a known stage of cell differentiation, and the edges are directed forwards in time.

Ihler et al (2007) gives several examples of how inference on tree graphical models can be used in climate science. Willsky (2002, from page 1399) reviews how tree graphical models have been used in a very wide variety of fields including oceanography, analysis of network traffic, and numerous aspects of image analysis.

Tree networks appear naturally in biology as phylogenetic trees, which show how different species have evolved from each other. Phylogenetics uses genetic information to infer this tree structure. Given their evolutionary predecessors, organisms are genetically independent of their predecessors' predecessors, so these trees can be regarded as probabilistic graphical models, as mentioned for example in Friedman (2004). But phylogenetics is a major field of research in its own right, and the models used are more elaborate. See chapter 7 of Durbin et al (1998) for an overview of learning phylogenetic trees from genetic data using clustering, bootstrapping, and other techniques from statistics. Incidentally, the full evolutionary tree of life on earth is not a tree. Genetic material is not only passed from organisms to their offspring, but is sometimes also transferred laterally, especially between bacteria. This phenomenon was first identified in Freeman (1951).

## 6.2 The claim that sparse graphs are locally tree-like

### Preamble

The notion that sparse graphs are locally tree-like is an important justification for studying tree and forest graphical models. If the true graph is locally tree-like, then a forest structure might give useful information about small parts of the graph, even if it is unlikely to be accurate across large sets of nodes. Anandkumar et al (2011) state that sparse graphs are locally tree-like and cite Bollobás (1985), which is the first edition of the book *Random Graphs*, about Erdős–Rényi graphs.

Speaking very informally, it seems believable that sparse graphs are locally tree-like, since if there are not many edges then there is not much chance of them being close to each other and forming short cycles. One example of an informal statement on this question appears in Macris (2006), which is a physics paper. The author claims that in sparse graphs with $n$ nodes the typical size of loops is $O(n)$; rephrasing in the caption of a figure, he says the loops are of size $O(n)$ with high probability. There is no citation or comment on how this is known, and there is no formal definition of “sparse” or “typical”.

## Interpretations of “sparse”

As discussed in section 2.4, it is often claimed that biomolecular networks are sparse, meaning that they have few edges, but a precise definition of “sparse” in this context is elusive. It may be useful to distinguish “degree-sparsity”, where the degrees of the nodes are small in some sense, from “edge-sparsity”, which means only that the total number of edges is small. But obviously the degrees and the number of edges are closely related. Some definitions refer to the average degree, which is $2|E|/n$. Clearly such definitions could be expressed in terms of the total number of edges.

It is possible to imagine three types of precise definition of sparse graphs. Firstly, “sparse” could be defined for a given graph, so that if someone gives you a graph you can examine it and say whether it is sparse or not. For example, you could say that a sparse graph is one with at most $2n$ edges ($n$ being the number of nodes). With graphical models it is natural to want to say whether a specific graph is sparse, since one deals with specific graphs that have specific numbers of nodes and edges.

Secondly, “sparse” could be defined for a random graph model with a fixed number of nodes. For example, $G(n,p)$ is sparse if $p \leq 0.2$. (For the definition of $G(n,p)$ see section 5.2.)

Thirdly, sparsity could be defined for a random graph model where $n \to \infty$, explicitly or implicitly. This would be an asymptotic definition that only makes assertions about all $n > N$, for some unspecified $N$, in terms of probabilities. Several definitions of this type appear in the literature. In Erdős–Rényi random graph theory and extremal graph theory, “sparse” usually means the number of edges is $O(n)$ (Diestel 2005, page 163; Bollobás 2001, pages 221 and 303). Bollobás & Riordan (2011) is about sparse graphs that have $\Theta(n)$ edges; it describes these graphs as “extremely” sparse and says they are the sparsest graphs that are interesting to study. Sudakov & Verstraëte (2008) say that dense graphs have average degree linear in $n$, which means they have $\Theta(n^2)$ edges and presumably implies that sparse graphs have $O(n)$ edges. A different asymptotic definition is that sparse graphs have average degree close to 2 (Bollobás & Szemerédi 2002).

These three types of definition are separate in that none of them imply any of the others. The first is about fixed graphs, the second is about fixed $n$, and the third is about $n \to \infty$.

For pure mathematicians, the first two types of definition are rather arbitrary. It is more natural to think of adjectives such as “sparse” in terms of asymptotic expressions such as $O(n)$, and so the third type of definition is the most common. The theory of Erdős–Rényi random graphs mostly uses this kind of definition, and the same is true of the rigorous results that have been proved about Barabási & Albert (1999)'s scale-free graphs—see chapter 4 of Durrett (2007). Janson et al (2000, page 2) say that the whole of random graph theory is asymptotic in nature. Of course the problem with asymptotic definitions is that graphs in the real world are finite, and these definitions say nothing about specific finite graphs.

Here are three other definitions of graph sparsity from the literature. The first two use degree-sparsity. Firstly, Dobra et al (2004) say a sparse graph is one where each node has only a small number of neighbours compared to the total number of nodes. This is convenient for their purposes but excludes the possibility of hubs, even though the paper is about GGMs for gene expression networks. Secondly, Meinshausen & Bühlmann (2006) use the condition that $\max_{v \in V} \deg(v) = O(n^{\kappa})$ for some $\kappa \in [0,1)$, for the purpose of proving asymptotic results. This condition means that $|E| = \frac{1}{2}\sum_v \deg(v)$ can be as large as $\frac{n}{2}\max \deg(v) = O(n^{1+\kappa})$, so it is similar to the asymptotic definitions above but less sparse than $|E| = O(n)$. If $\kappa$ is small, it would imply the condition of Dobra et al (2004). Thirdly, Wille & Bühlmann (2006) state simply that "if the number of … edges is much smaller than $p(p-1)/2$ [they use $p$ instead of $n$, so this is the maximum possible number of edges], a graph is generally referred to as being sparse."

## Interpretations of "locally tree-like"

The phrase "locally tree-like" appears in the literatures of physics, computer science, and pure mathematics, as well as statistics, with a wide variety of interpretations. Many definitions of it are brief, informal, and only given in passing.

Perhaps the simplest interpretation of "locally tree-like" is that it means the graph has few short cycles (of course this is still vague and not a formal mathematical idea). This is the interpretation that appears in Anandkumar et al (2011), Forney (2003), Brummitt et al (2012), and Sly (2010). With this interpretation, the claim that sparse graphs are locally tree-like can be justified to some extent using the theorems about Erdős–Rényi random graphs that are described in the next section. Using these theorems will mean interpreting the vague words "few" and "locally" to have asymptotic meanings, as with the third type of definition of "sparse" in the previous subsection.

One alternative to investigating the property of having few short cycles is to investigate the property of having none. This would mean investigating the girth of random graphs. Neither Bollobás (2001) nor Janson et al (2000), whose preface describes it as an update of Bollobás (1985), give any results about girth. But some facts can be deduced from the results on small cycles. For example, $\mathbb{P}(\text{girth} \geq 5) = \mathbb{P}(X_3 = 0 \cap X_4 = 0)$, where $X_i$ is the number of cycles of length $i$, and the theorems in the next subsection give information about $\mathbb{P}(X_i = 0)$.

As with "sparse", it would be possible to make a precise definition of "locally tree-like" for given graphs or given $n$. For example, the girth of the graph has to be at least 5, or $n/4$.

Most of the other interpretations of "locally tree-like" refer to the girth. For example, Miller (2008) uses a sequence of graphs $G_1, G_2, \ldots$ and takes locally tree-like to mean that $G_m$ has girth greater than $2m$, Coja-Oghlan et al (2009) take it to mean there are either no short cycles or the girth is $\Omega(\log n)$ (the relevant sentences are conjectures or informal observations), Chandar (2010) takes it to mean the girth is $\Omega(\log n)$, and Vontobel (2003) simply says that it means there are "no short cycles". Durrett (2007, page 134) says that being locally tree-like is the same thing as having no triangles—which has the practical advantage that it is completely precise and applies to actual specific graphs. Karrer & Newman (2010) state that it means "all small connected subsets of vertices within the network are trees." Cooper & Frieze (2010) give a formal definition for whether a node, rather than a graph, is locally tree-like, based on whether any cycles appear when you explore up to a certain distance away from the node.

## Small cycles in Erdős–Rényi random graphs

The most relevant rigorous results are the set of theorems about the numbers of small cycles in an Erdős–Rényi random graph. For definitions of Erdős–Rényi random graphs and notations, see section 5.2. The present question of the number of small cycles is one of the many questions for which results about the two Erdős–Rényi models are very similar.

Following Bollobás (2001), the canonical text in this field, I will use the notation $a(n) \sim b(n)$ to mean that $\lim_{n\to\infty} \frac{a(n)}{b(n)} = 1$ (as mentioned in Appendix III). This subsection does not present new results but rather presents and uses theorems and corollaries that have been proved elsewhere.

***Theorem 6.1 (Theorem 3a in Erdős & Rényi 1960).*** In $G(n, M)$, suppose that $M(n) \sim cn$, where $c > 0$. Let $X_i$ be the number of cycles of length $i$ in $G$. Then

$$\mathbb{P}(X_i = j) \sim \frac{\lambda^j e^{-\lambda}}{j!},$$

where $\lambda = (2c)^i/2i$. In other words, $X_i$ converges in probability to Poisson($\lambda$) as $n \to \infty$.

***Theorem 6.2.*** With the same assumptions as in Theorem 6.1,

$$\mathbb{P}(X_i = j_i \text{ for } i = 3,4,\dots,t) \sim \prod_{i=3}^{t} \frac{\lambda_i^{j_i} e^{-\lambda_i}}{j_i!}$$

where $\lambda_i = (2c)^i/2i$. In other words, the joint distribution of the $X_i$'s converges in probability to the joint distribution of independent Poisson random variables with the given means.

***Proof.*** This follows from Theorem 4 of Bollobás (1981), using the fact that the automorphism group of the cycle of length $i$ has size $2i$.

Theorem 6.1 is about a single cycle-length. Theorem 6.2 is just Theorem 6.1 plus the additional result that the numbers of different-sized cycles are asymptotically independent.

***Theorem 6.3 (Corollary 4.9 in Bollobás 2001, Corollary 9 in Bollobás 1985).*** In $G(n, p)$, suppose that $p(n) \sim c/n$, where $c > 0$. Let $X_i$ be the number of cycles of length $i$ in $G$. Then

$$\mathbb{P}(X_i = j_i \text{ for } i = 3,4,\dots,t) \sim \prod_{i=3}^{t} \frac{\lambda_i^{j_i} e^{-\lambda_i}}{j_i!}$$

where $\lambda_i = c^i/2i$.

Obviously Theorem 6.2, about $G(n,M)$, is almost exactly the same as Theorem 6.3, about $G(n,p)$. But in the former, the Poisson parameters have a factor of 2 that does not appear in the latter. This is to be expected, since in $G(n,M)$ with $M(n) \sim cn$ the expected number of edges is asymptotically $cn$, whereas in $G(n,p)$ with $p(n) \sim c/n$ the expected number of edges is $c(n-1)/2$, or equivalently $cn/2$; and the two models correspond most closely when the expected numbers of edges are the same.

The main results in Bollobás (1981) and section 4.1 of Bollobás (2001) are more general than the ones given above. They state that if a possible subgraph is "strictly balanced", then the number of copies of it that appear in a random graph is asymptotically distributed as a Poisson random variable. "Strictly balanced" is a property possessed by all cycles, and "copies" of a subgraph means subgraphs that are isomorphic to it.

## The values of the Poisson parameters for Erdős–Rényi graphs

The above results on Erdős–Rényi random graphs use the assumption that $M(n) \sim cn$ or $p(n) \sim c/n$. The number of possible edges is $\sim n^2/2$. So both the assumptions imply that the graph is sparse in the sense that the number of edges is $O(n)$—as discussed above, this is the meaning of "sparse" that is most often used in graph theory. For fixed $c$ and any given $i$, the Poisson parameters do not depend on $n$. This means that for large $n$ they are "small", in the sense that as $n \to \infty$ the proportion of cycles of length $i$ that appear in the graph will tend to zero. (This proportion is the number of cycles of length $i$ that appear in the graph divided by the number of cycles of length $i$ in $K_n$.) In this sense the results imply that for large $n$ sparse graphs have few short cycles—in other words, they are locally tree-like.

A secondary question is whether short cycles are commoner or rarer than long ones. In Theorem 6.2, $\lambda_i = (2c)^i/2i$. For large $i$ this is dominated by $(2c)^i$. If $c > 0.5$, then long cycles are more likely than short ones, but if $c \leq 0.5$, then short cycles are more likely than long ones. However, the latter case corresponds to extremely sparse graphs where $|E| \leq |V|/2$ (at least asymptotically). These have so few edges that long cycles are not even possible. In Theorem 6.3, $\lambda_i = c^i/2i$, so the situation is essentially the same but the boundary between the two cases is $c = 1$.

The next question is whether the Poisson parameters $\lambda_i$ are actually reasonably small for sparse graphs with realistic $n$. The obvious way to investigate this is to choose $n$, and $p$ or $M$, and find $\lambda_i$ using the equations given above. If $\lambda_i$ is small for, say, $i \in \{3,4,5,6\}$, then it can be said that the graph is locally tree-like. Table 6.1 shows the values of the Poisson parameters for small $i$ with a sparse model from $G(n,p)$, a sparse model from $G(n,M)$, and the random graph model $G(n,p = 0.5)$, where all graphs are equally likely. Of course the Poisson parameters are also the asymptotic mean numbers of cycles.

The first two random graph models have the same Poisson parameters. But the parameters for $G(100{,}0.5)$ are much bigger. Based on this it seems reasonable to state that for the sparse random graph models the Poisson parameters are small.

| **Random graph model** | **Poisson parameters for numbers of cycles of lengths from 3 to 8** | | | | | |
|---|---|---|---|---|---|---|
| | **3** | **4** | **5** | **6** | **7** | **8** |
| $G(n,p)$ with $p = 10/n$<br>Example: $G(100{,}0.1)$ | 167 | 1 250 | 10 000 | 83 333 | 714 286 | 6 250 000 |
| $G(n,M)$ with $M = 5n$<br>Example: $G(100{,}500)$ | 167 | 1 250 | 10 000 | 83 333 | 714 286 | 6 250 000 |
| $G(100, p = 0.5)$ | 20 833 | 781 250 | $3.1 \times 10^7$ | $1.3 \times 10^9$ | $5.6 \times 10^{10}$ | $2.4 \times 10^{12}$ |

**Table 6.1.** Poisson parameters for the numbers of cycles for two sets of sparse Erdős–Rényi random graph models and for $G(100{,}0.5)$. These are from Theorems 6.2 and 6.3. (In $G(100{,}0.1)$, $G(100{,}500)$, and $G(100{,}0.5)$, the parameter $c$, defined in Theorems 6.2 and 6.3, was chosen so that the asymptotic conditions were satisfied exactly; so for $G(n,p)$ it was chosen to be $np$, and for $G(n,M)$ it was chosen to be $M/n$.)

## Using simple Monte Carlo to approximate the numbers of small cycles

There is often a tacit assumption that asymptotic results will be approximately true for realistic-sized problems. But graphical models usually have fixed $n$, and for any fixed $n$ it is quite possible that a given asymptotic result does not hold, even approximately.

The obvious avenue of approach is to look at actual random graphs, identify all their cycles, and examine the numbers of cycles of each length. For very small $n$ it would be possible to generate all the possible graphs for either of the Erdős–Rényi random graph models. For $G(n,p)$, this would involve generating all $2^{\binom{n}{2}}$ possible graphs and weighting them according to their probabilities. This is only feasible for $n$ up to about 8. For $G(n,M)$, it would only be necessary to generate the $\binom{\binom{n}{2}}{M}$ graphs that are possible under this model, but even for $n = 16$ and $M = 10$ this is $1.2 \times 10^{14}$, which is very large. So it seems more sensible to look at a sample of randomly generated graphs from $G(n,p)$ or $G(n,M)$. This is a type of simple Monte Carlo method. If the sample is reasonably big then the empirical distributions of the numbers of cycles should be similar to the true distributions.

Identifying all the cycles in a given undirected graph can be done by considering all the possible cycles and then checking if they are present. Coming up with a more efficient method is far from trivial. Much research on this problem appeared in the late 1960s and early 1970s. Mateti & Deo (1976) give a summary of all the algorithms that were known at that time, classified into four basic types. The simplest way I have found is to use the algorithm in Paton (1969) to find a "fundamental set" of cycles, which is a basis of the vector space of all the cycles, and then use the algorithm in Gibbs (1969) to generate the other cycles. I have not found any single document that gives a complete account of how to find all the cycles in an undirected graph.

I wrote a program to generate a simple Monte Carlo sample of random graphs from $G(n, p)$ or $G(n, M)$ and then count all their cycles. Some results are shown in Figure 6.1. The most relevant cycles are the shortest ones, and the bar-charts show that there are indeed few of these. The asymptotic means are close to the true values for the very shortest cycles.

## Small cycles in graphs with given degree sequences

As discussed above, claims that sparse graphs are locally tree-like have usually been made with reference to theorems about cycles in Erdős–Rényi random graphs, or with reference to nothing. But, as discussed in section 2.4, Barabási & Oltvai (2004) and other papers have claimed that real networks tend to be "scale-free", meaning that over a large range the degrees of the nodes follow a power law. This would mean that the Erdős–Rényi random graph models are not appropriate. It might be more sensible to consider theorems about random graph models that are scale-free or approximately so.

The most relevant results seem to be the ones about cycles in random graphs with given degree sequence. This subsection will present the results but not a detailed investigation. A degree sequence is a list of $n$ non-negative integers. For a graph to have a given degree sequence means that the degrees of its nodes are the same as the integers in this list, in some order. The random graph model is that all graphs with the given degree sequence are equally likely. Random graphs with given degree sequence could be made scale-free by choosing the degree sequence to follow a power law.

The other main feature of biomolecular networks mentioned in section 2.4 was that they contain hubs. Small cycles in graphs that contain hubs could be investigated using the theorems on random graphs with given degree sequence. The degree sequence should simply be chosen to contain a small number of large degrees.

In either case it is necessary to specify a degree sequence. Given a degree sequence, the theorems below can be used to calculate the asymptotic numbers of short cycles. But not just any sequence of integers can be a degree sequence. For a start, they have to be in $\{0, \ldots, n-1\}$ and their sum has to be even. But this is not sufficient—for example, $\{2,2,0\}$ is not the degree sequence of any graph. Necessary and sufficient conditions are given by Theorem 6.4.

***Theorem 6.4 (Erdős & Gallai 1960).*** Let $\{d_1, \ldots, d_n\}$ be a non-increasing sequence of non-negative integers. This is the degree sequence of some graph if and only if $\sum_{i=1}^{n} d_i$ is even and $\sum_{i=1}^{k} d_i \leq k(k-1) + \sum_{i=k+1}^{n} \min\{k, d_i\}$ for $1 \leq k \leq n$.

***Proof.*** For a proof in English see Tripathi et al (2010).

Theorem 6.5 gives the asymptotic distributions of the numbers of cycles.

***Theorem 6.5 (Theorem 2 in Bollobás 1980).*** Let $\{d_1, \ldots, d_n\}$ be a given degree sequence, and suppose all graphs with this degree sequence are equally likely. Let $m = ½ \sum_{i=1}^{n} d_i$ be the number of edges in each of these graphs, and assume that $2m - n \to \infty$ as $n \to \infty$. Then the numbers of cycles $X_3, \ldots, X_k$ are asymptotically independent Poisson random variables with means $\lambda_i = \lambda^i / 2i$, where $\lambda = \frac{1}{m} \sum_{i=1}^{n} \binom{d_i}{2}$.

(a) $G(22,0.1)$

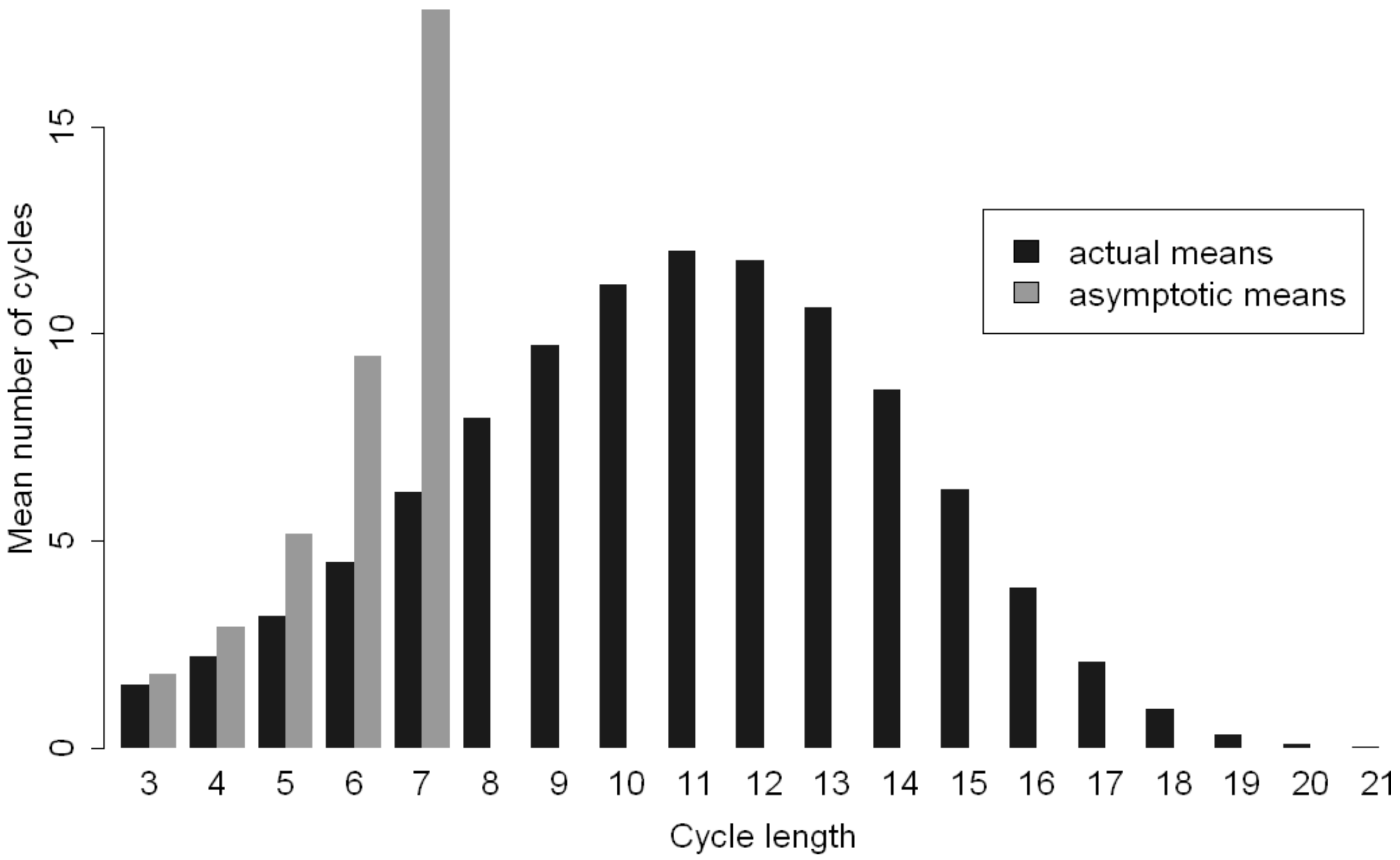

(b) $G(30,40)$

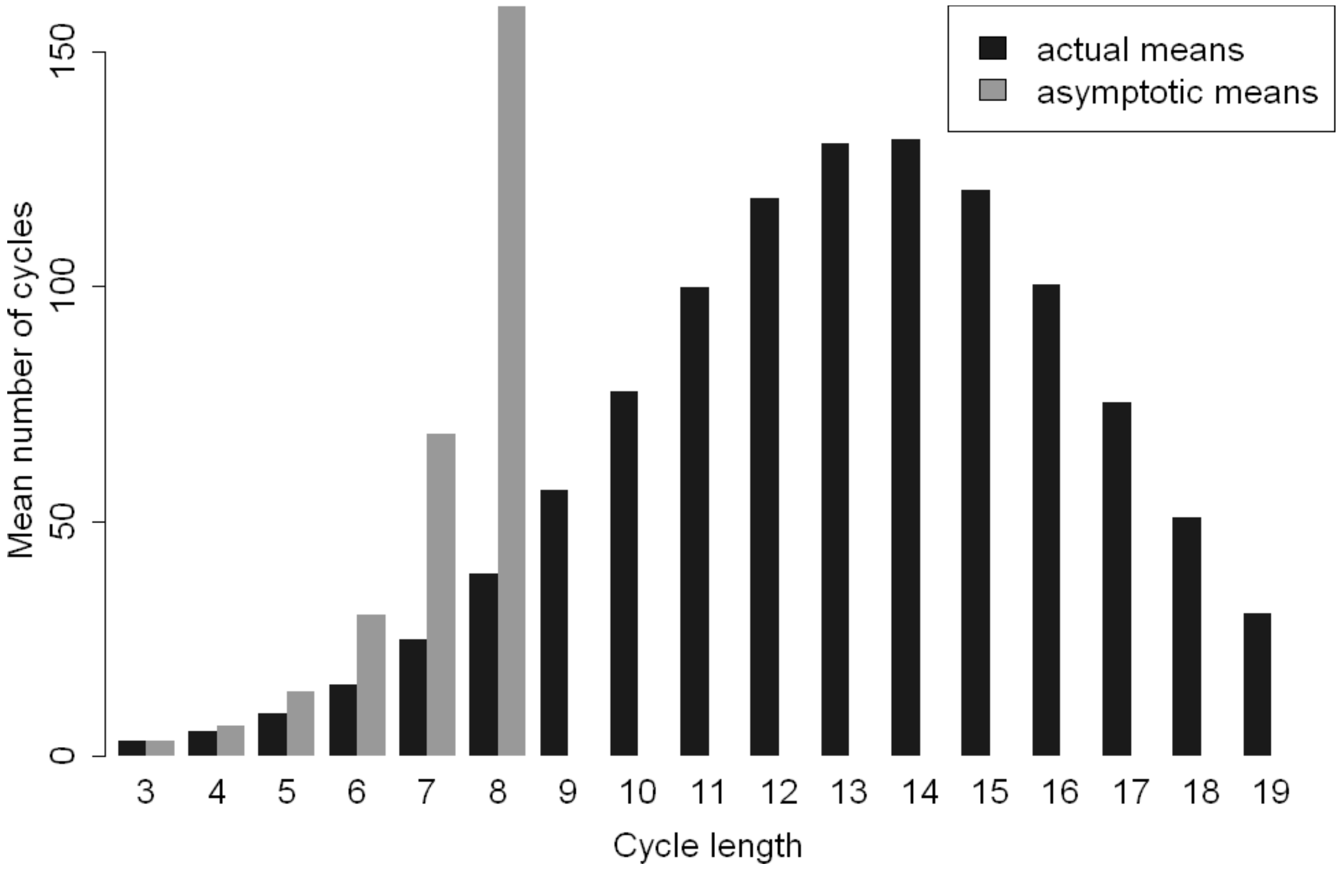

**Figure 6.1.** Numbers of cycles in (a) $G(22,0.1)$ and (b) $G(30,40)$. The "actual" means were found by generating simple Monte Carlo samples of 5000 graphs and then counting all the cycles. In (a) the asymptotic means are the Poisson parameters from Theorem 6.3, with $c = np$; in (b) they are the Poisson parameters from Theorem 6.2, with $c = M/n$. The asymptotic means are only shown for short cycles, since they get very big, and in (b) the largest cycle-lengths are omitted altogether.

Whether these Poisson parameters can be regarded as small depends on the degree sequence, or more accurately the sequence of degree sequences, since the theorem is about asymptotic behaviour as $n \to \infty$. (Theorem 6.5 was also proved independently as Corollary 1 in Wormald 1981. Wormald's result is slightly different, being about cycles of an arbitrary set of lengths.)

A special case of graphs with given degree sequences is regular graphs. Regular graphs are ones where every node has the same degree, and $d$-regular graphs are ones where every node has degree $d$. All $d$-regular graphs have $nd/2$ edges, so if $d$ is fixed they are sparse in the sense that the number of edges is $O(n)$. Many of the results on regular graphs assume that $d \geq 3$. This is not a restrictive assumption, since if $d = 2$ then the graph is one big loop, and if $d = 1$ then it consists of nothing but connected components of size 2—these cases are both trivial.

Regular graphs definitely do not have hubs, so they are not likely to be good models for biomolecular networks. In any case, for regular graphs Theorem 6.5 simplifies to give the following result.

***Theorem 6.6. (Wormald 1981; Bollobás 1980; Bollobás 2001, page 56).*** Assume graphs are chosen uniformly at random from the set of $d$-regular graphs. For fixed $k \geq 3$, the numbers of cycles $X_3, \dots, X_k$ are asymptotically independent Poisson random variables with means $\lambda_3, \dots \lambda_k$, where $\lambda_i = (d-1)^i/2i$.

As with Theorems 6.2 and 6.3, about Erdős–Rényi graphs, these Poisson parameters do not depend on $n$, so as $n \to \infty$ the proportion of cycles of length $i$ that appear in the graph tends to zero. In this sense the graphs have few short cycles and are locally tree-like.

Bollobás (2001, page 84) comments as follows on the similarity between this result for regular graphs and the results for Erdős–Rényi random graphs. A random $d$-regular graph is in many ways similar to either $G(n,p)$ with $p(n) = d/n$ or $G(n,M)$ with $M(n) = dn/2$. In a random $d$-regular graph the expected number of cycles of length $i$ is $(d-1)^i/2i$, but in the two Erdős–Rényi models it is $d^i/2i$. This means that short cycles are slightly less likely in regular graphs than in Erdős–Rényi random graphs.

Slightly stronger than Theorem 6.6 is Theorem 6.7, in which $d$ is allowed to grow slowly as a function of $n$.

***Theorem 6.7 (Theorem 1 in McKay et al 2004).*** Assume graphs are chosen uniformly at random from the set of $d$-regular graphs. Allow $d = d(n)$ and $g = g(n)$ to increase with $n$, so long as $(d-1)^{2g-1} = o(n)$. Let $\{c_1, \dots, c_k\}$ be a set of cycle-lengths that is a non-empty subset of $\{3, \dots, g\}$. Then $X_{c_1}, \dots, X_{c_k}$, the numbers of cycles of these lengths, are asymptotically independent Poisson random variables with means $\lambda_i = (d-1)^{c_i}/2c_i$.

***Corollary 6.8 (Corollary 1 in McKay et al 2004).*** With the same assumptions as in Theorem 6.7, the probability that the girth is greater than $g$ is

$$\exp\left(-\sum_{r=3}^{g}(d-1)^r/2r + o(1)\right).$$

If the restriction to the class of regular graphs is lifted, and instead $d$ is the maximum degree of the graph, then it might be possible to use one of the above results on asymptotic Poisson distributions as a sort of "upper bound" and thus show that a further class of sparse graphs is locally tree-like. Removing edges from a regular graph can only decrease the number of cycles of any given length.

## Summary

The purpose of this section was to consider the notion that sparse graphs are locally tree-like. Probably the most natural way to interpret "sparse" is asymptotically (with $n \to \infty$), and probably the most natural way to interpret "locally tree-like" is that there are few short cycles, where "few" and "short" are also to be interpreted asymptotically. When these interpretations are used, theorems about the two Erdős–Rényi random graph models can be used to justify the notion that sparse graphs are locally tree-like.

The first drawback of this is that asymptotic theorems say nothing about real graphs with fixed $n$. The only remedies are to generate actual graphs and count the cycles, as I did above, or to come up with new theorems about the numbers of short cycles in graphs with specific $n$.

The second drawback is that, as mentioned in section 2.4 and 5.3, real networks are not well modelled by Erdős–Rényi graphs. For scale-free graphs or other complex networks it may be possible to use the theorems about cycles in graphs with given degree sequence, though some of these random graphs are not defined as precisely as Erdős–Rényi graphs.

## Supplementary notes: extremal graph theory

The field of extremal graph theory (Bollobás 1978, Diestel 2005) addresses problems that are somewhat related to the question of being locally tree-like. Extremal graph theory is about "all graphs" or "no graphs"; it is not concerned with "most graphs" or random graphs. The archetypal question is to find the number of edges, as a function of the number of nodes, such that all graphs with that many edges contain a certain subgraph. The biggest graphs that do not contain the subgraph are called the extremal graphs. For example, Turán (1941) found the graph that has the maximum number of edges among graphs with $n$ nodes that do not contain any copies of $K_r$. This graph, known as the Turán graph, is the extremal graph for this problem.

Extremal graph theory is not just about subgraphs but also about other properties. The most relevant property to the question of being locally tree-like is girth. As mentioned above, "locally tree-like" could be taken to mean that a graph has large girth. Bollobás & Szemerédi (2002) show that the girth of a graph with $n$ nodes and $n + k$ edges is at most $2(n + k)(\log k + \log\log k + 4)/3k$, for $n \geq 4$ and $k \geq 2$. Section III.1 of Bollobás (1978) is about graphs with large minimal degree and large girth.

Sudakov & Verstraëte (2008) prove various results about the set of lengths of cycles in a graph. For example, given the average degree and the girth of the graph, they give an asymptotic lower bound for the size of this set and show that it contains a certain number of consecutive even integers.

# 7 The Chow–Liu algorithm

## 7.1 Finding the optimal tree

Suppose you are given a joint distribution on a set of discrete-valued random variables, and each variable corresponds to a node. From the distributions on these variables that are Markov with respect to a tree on these nodes, you have to find one that is closest to the given distribution in terms of Kullback–Leibler distance. An elegant algorithm for solving this problem was given in Chow & Liu (1968), a much-cited paper that appeared before probabilistic graphical models were widely studied.

The paper also addressed a dual problem. Suppose the true distribution on the discrete-valued random variables is unknown but observations from it are available. From the distributions that are Markov with respect to a tree, find one that has maximum likelihood. The algorithm for doing this is essentially the same as the one for the first problem, except that it uses empirical distributions rather than true ones.

The algorithm for the second problem is described below. For both problems it is possible for there to be more than one optimal tree, but I will sometimes refer to "the" optimal tree as this is easier to read.

Chow & Liu describe their algorithms using directed graphical models in the shape of rooted trees. Each of their arrows points from a dependent variable to the variable it depends on, and towards the root—the opposite direction to what is now standard. Any rooted-tree DAG is Markov-equivalent to an undirected tree, so the algorithms and results can also be stated using undirected graphs.

Given observations from a discrete-valued multivariate distribution, the obvious way of finding the optimal tree and distribution would be to first find the optimal distribution for each possible tree, and then look through all the trees and find the optimal one. It turns out that these two steps can be done in one. Let the nodes be $\{1, \ldots, p\}$, and without loss of generality let node 1 be the root of the tree. Chow & Liu write the density as

$$p(x) = p(x_1) \prod_{i=2}^{p} p\big( x_i \mid x_{pa(i)} \big),$$

where $pa(i)$ is the parent of node $i$ (they would draw the arrow from $i$ to $pa(i)$). If there are $n$ independent and identically distributed data, the likelihood is $\prod_{k=1}^{n} p\big(x^k\big)$. Chow & Liu (1968) show that maximizing the log-likelihood over all tree distributions is equivalent to maximizing

$$\sum_{(u,v)\in E} I_{u,v}\,,$$

where $(u, v)$ is an unordered pair and $I_{u,v}$ is the empirical mutual information, also called the sample mutual information (Chow & Liu 1968) or the empirical cross-entropy (Lauritzen 2006):

$$I_{u,v} = \sum_{x_u,\, x_v} \frac{n(x_u, x_v)}{n} \log \frac{n(x_u, x_v)/n}{n(x_u)n(x_v)/n^2}\,.$$

Here for example $n(x_u, x_v)$ is the number of observations of $X_u = x_u$ and $X_v = x_v$, and if any of the $n(\cdot)$'s is zero then the summand is taken to be zero. Recall that the variables are all discrete. $I_{u,v}$ is always non-negative.

As an aside, note that the elements in the expression for $I_{u,v}$ are maximum-likelihood estimators. For example, $n(x_u, x_v)/n$ is the MLE of $\mathbb{P}(X_u = x_u, X_v = x_v)$. This is just the observation that in discrete decomposable graphical models the MLE of each probability on a clique is simply the observed proportion of the data that take that set of values.

The first part of the Chow–Liu algorithm is to calculate $I_{u,v}$ for all possible edges $(u, v)$. In the second part of the algorithm, the $I_{u,v}$'s are regarded as weights on the possible edges. The task that remains is to find the tree with maximum total weight. This is a maximum-weight spanning tree (MWST) problem. In this case what is sought is a maximum-weight tree that spans the complete graph $K_p$.

There are several simple algorithms that can solve MWST problems in polynomial time and are thus efficient in high dimensions. Chow & Liu (1968) and most papers based on it use Kruskal's algorithm (Kruskal 1956), which is described in the next section. For another discussion of the Chow–Liu algorithm see Pearl (1988, section 8.2.1). The asymptotic time that the algorithm takes is discussed in section 6.1.

## 7.2 Kruskal's algorithm

### The algorithm as used in the Chow–Liu algorithm

Given a complete graph on a set of nodes, and real-valued weights on each edge, the following algorithm produces a maximum-weight spanning tree—in other words, a maximum-weight tree on the same set of nodes.

#### Algorithm IV: Kruskal's algorithm

1. Start with the empty graph on the given set of $p$ nodes.
2. From among the unused edges whose addition would not lead to the appearance of a cycle, add the one with largest weight. (If there are two or more such edges, add any one of them.)
3. Repeat step 2 until you have $p - 1$ edges.

## Different versions of the algorithm

Algorithm IV gives a maximum-weight spanning tree for the complete graph. Kruskal's algorithm is usually stated in a different form, for finding the minimum-weight spanning tree for a given connected graph (Kruskal 1956). Bondy & Murty (2008) say that Kruskal's algorithm first appeared in Borůvka (1926a,b), which are in Czech, and that Kruskal's discovery of it was independent.

Whether the total weight has to be minimized or maximized is obviously trivial. To minimize instead of maximize, simply replace "largest" with "smallest" in step 2.

Moreover, as pointed out by Kruskal (1956), there is no loss of generality in considering only the complete graph. Suppose you want to find a maximum-weight spanning tree for a connected graph $G = (V, E)$ that is not complete. In the complete graph $K_{|V|}$, set the weight of each edge $e$ to be its weight in $G$ if $e \in E$, or $-\infty$ if $e \notin E$, and do Algorithm IV. $G$ spans $K_{|V|}$ and thus contains at least one spanning tree. This means that $K_{|V|}$ has at least one spanning tree with no infinite-weight edges. It follows that the algorithm will never add an infinite-weight edge and will produce a tree that spans $G$. Any proof of correctness for Algorithm IV will therefore also suffice as a proof of correctness for the more usual form of Kruskal's algorithm, and vice versa.

The most well-known alternative to Kruskal's algorithm is Prim's algorithm, for which a full proof appears in section 2.2 of Even (1979).

## Proofs

Proofs that the usual form of Kruskal's algorithm is correct can be found in Bondy & Murty (1976, page 39) or Aldous & Wilson (2000, pages 190–191). Kruskal (1956) only proved that the algorithm is correct when the weights are all distinct. All these proofs use the same basic idea of identifying a cycle and then modifying a tree by adding an edge and removing one—see also Theorem 2.3 in Even (1979). They use Proposition 2.1, which stated that adding an edge to a tree creates a graph with precisely one cycle.

Kruskal (1956) proved that if all the edge-weights are distinct then the minimum-weight spanning tree is unique. (Actually this was the main purpose of the paper; the algorithm was merely a way to prove this fact.) If the edge-weights are not distinct, then the minimum- or maximum-weight spanning tree is not necessarily unique.

# 7.3 Relevant developments since Chow–Liu

Chow & Liu's method can be adapted to the case of the multivariate Gaussian distribution. Again it gives either the tree that minimizes the Kullback–Leibler distance to the true distribution or the maximum-likelihood tree. As in the discrete case, the weight on each edge is the mutual information. This equals $-\frac{1}{2}\log(1-\rho_e^2)$, where $\rho_e$ is the actual or empirical correlation coefficient along edge $e$. Transforming the edge-weights by any monotone-increasing function gives the same result in Kruskal's algorithm, so it is also possible to use just $\rho_e^2$ (Goldberger & Leshem 2009). This method for GGMs appears in Lauritzen (2006), Goldberger & Leshem (2009), and Tan et al (2010a).

Edwards et al (2010) gives two adaptations of Chow & Liu's method. The first deals with the drawback that even if the true graph is a forest, Chow & Liu's method will always produce a tree. This is analogous to the facts that in graphical models the maximum-likelihood graph is always the complete graph and in regression problems the maximum-likelihood model always includes all the covariates. They adapt Chow & Liu's method to optimize a penalized likelihood criterion such as AIC or BIC. In general, this produces a forest, not a tree. The formula for the edge-weights is changed by subtracting a certain quantity. This means that the edge-weights can be negative. To find the optimal forest, you remove all the edges whose weights are negative and then do Kruskal's algorithm on all the connected components. Their second adaptation is an extension to mixed graphical models where some nodes are discrete and some are Gaussian. Obviously GGMs are a special case of these models. The paper also includes several example applications of the methods. One of these uses the breast cancer data from the R package "gRbase". The analysis took 18 seconds and located several nodes that seem to be hubs. It is also described in section 7.3 of Højsgaard et al (2012).

The two algorithms in Edwards et al (2010) have been implemented in the R function minForest, in the package "gRapHD" (Abreu et al 2010). See "Finding the MAP forest in R" in section 7.4.

## 7.4 Finding the MAP forest

### Using a uniform graph prior

In Bayesian structure-learning, the graph that has highest posterior probability is called the MAP (maximum *a posteriori* probability) graph. An adaptation of the Chow–Liu algorithm can be used to find the MAP forest for discrete random variables, assuming that the graph prior is uniform on the set of forests (Højsgaard et al 2012, section 7.7). The weight of each edge is taken to be the logarithm of the Bayes factor for the presence of that edge, and the version of Kruskal's algorithm in Edwards et al (2010) is then used to find the MAP forest. (Remove the edges that have negative weights, then do Kruskal's algorithm on the graph that remains.) Forests like this are sometimes called spanning forests, where "spanning" just means that the forest has the same set of nodes as the original graph.

This method can be adapted to GGM structure-learning with the hyper inverse Wishart prior on the covariance matrix. This adaptation appears in lectures 8 and 9 of Lauritzen (2006). Let $G = (V, E)$ and $p = |V|$. Assume the graph prior distribution is uniform, so $p(G)$ is the same for every graph and $p(\, G \mid x\,) \propto p(x \mid G)$. The method is best explained by rearranging the formula for the marginal likelihood $p(x \mid G)$. In section 3.1 I wrote the formula for the marginal likelihood of a decomposable graph as

$$p(\, x \mid G\,) = (2\pi)^{-np/2} \frac{\prod_C \frac{k(C, \delta, D)}{k(C, \delta + n, D + U)}}{\prod_S \frac{k(S, \delta, D)}{k(S, \delta + n, D + U)}},$$

where

$$k(C,\delta,D) = \frac{\left|\frac{D_C}{2}\right|^{\frac{\delta+|C|-1}{2}}}{\Gamma_{|C|}\left(\frac{\delta+|C|-1}{2}\right)}.$$

In the decomposition of a forest the cliques are the edges (strictly, the pairs of nodes that have edges between them) and the isolated nodes (the nodes that have degree zero). The separators are a subset of the individual nodes plus, if the forest is not connected, the empty set. The number of times each node appears as a separator is its degree minus one; any empty separators contribute a factor of 1 and can thus be ignored.

So for forests the marginal likelihood is

$$p(x \mid G) = (2\pi)^{-np/2} \frac{\prod_{(u,v)\in E} \frac{k(\{u,v\},\delta,D)}{k(\{u,v\},\delta+n,D+U)}}{\prod_{v\in V}\left[\frac{k(\{v\},\delta,D)}{k(\{v\},\delta+n,D+U)}\right]^{\deg(v)-1}}.$$

Let

$$K(v_1,v_2,\dots,v_m) = \frac{k(\{v_1,v_2,\dots,v_m\},\delta,D)}{k(\{v_1,v_2,\dots,v_m\},\delta+n,D+U)}.$$

($K$ is a "variadic" function that takes any number of arguments.) Then

$$\begin{aligned} p(x \mid G) &= (2\pi)^{-np/2} \frac{\prod_{(u,v)\in E} K(u,v)}{\prod_{v\in V} K(v)^{\deg(v)-1}} \\ &= (2\pi)^{-np/2} \prod_{v\in V} K(v) \prod_{(u,v)\in E} \frac{K(u,v)}{K(u)K(v)}. \end{aligned}$$

In this last expression, the first product is over the nodes, which are the same for all graphs. So to choose among forests with different edges it is sufficient to maximize the second product. This can be done using an adaptation of Chow & Liu's algorithm. Simply let the weight of edge $(u,v)$ be $\log\frac{K(u,v)}{K(u)K(v)}$, and as in Edwards et al (2010) omit all edges that have negative weights.

After describing how to use penalized likelihoods with the Chow–Liu algorithm, Edwards et al (2010) states that Panayidou (2011) "finds the Bayesian MAP tree/forest in a similar way". This probably refers to the method in Højsgaard et al (2012) or the method I have described. (Panayidou 2011 can only be viewed by travelling to Oxford.)

This method for finding the MAP forest is fast, like the standard Chow–Liu algorithm. Similar methods appear in Meilă-Predoviciu (1999, page 59) and Heckerman et al (1995, pages 226–227). Lauritzen (2006) poses the question of whether there a feasible algorithm for finding the MAP decomposable graph, and states that the answer is probably no since this task is probably NP-complete.

## Factored priors

The method described in the previous section requires the graph prior distribution to be uniform, so that $p( G \mid x ) \propto p(x \mid G)$. But it can also be adapted to work with factored distributions, which were defined in section 5.4. For factored distributions, each edge $(u, v)$ has associated with it a quantity $w_{uv}$, and the probability of $G = (V, E)$ is

$$p(G) \propto \prod_{(u,v)\in E} w_{uv} \,.$$

With a graph prior distribution of this form, the posterior probability of graph $G$ is

$$p( G \mid x ) \propto p( x \mid G )p(G)$$

$$= (2\pi)^{-np/2} \prod_{v\in V} K(v) \prod_{(u,v)\in E} \frac{K(u,v)}{K(u)K(v)} \prod_{(u,v)\in E} w_{uv} \,.$$

The only parts of this formula that depend on the forest are the second and third products. So the weight of edge $(u, v)$ should be set to

$$\log\left(\frac{K(u,v)}{K(u)K(v)}\, w_{uv}\right).$$

Kruskal's algorithm can now be used as before to find the MAP forest. Edges whose weights are negative should be removed before doing Kruskal's algorithm.

This method cannot be adapted to work with general graph prior distributions. In the general case the posterior probability of a graph $G$ is $p( G \mid x ) \propto p( x \mid G )p(G)$, and the prior probability of the graph, $p(G)$, cannot be factorized into a contribution from each edge. So it is impossible to write $p( G \mid x )$ in a way that reduces the problem to that of finding a maximum-weight spanning forest. For example, with the size-based priors of Armstrong et al (2009), adding an edge does not have a fixed multiplicative effect on the probability of the graph; the effect depends rather on how many edges there are in total.

## Finding the MAP forest in R

The MAP forest for GGMs can be found using the function minForest, from the R package "gRapHD" (Abreu et al 2010; see also Edwards et al 2010 or Højsgaard et al 2012, section 7.4). You have to write a custom function to calculate the edge-weights and then pass this function to minForest as the "stat" argument. The only other arguments you need to provide are the data (an $n \times p$ matrix) and the prior values of the two HIW hyperparameters. I tested this method on the iris data used by Roverato (2002) and it gave the right answer.

Despite its name, minForest finds the forest that maximizes the given edge-weights. In its documentation the Arguments section states that the default value of stat is "LR", but actually it is "BIC".

## 7.5 Supplementary notes

### Methods for finding the top few trees

Kruskal's algorithm finds the optimal tree. There are also fast algorithms that can find the top $k$ trees, for any $k \in \{1, \dots, p^{p-2}\}$, or all the spanning trees in order from best to worst. See Gabow (1977), Camerini et al (1980), Eppstein (1990), Sörensen & Janssens (2005), or Climaco et al (2008). Cowell (2013) uses the algorithm of Sörensen & Janssens (2005) to find the most likely pedigree charts for a group of animals or humans.

### Other edge-weights in the Chow–Liu algorithm

The algorithm in section 7.1 does not have to be done with these exact weights on the edges. Transforming the edge-weights by any monotone increasing function makes no difference. For more on this see Acid et al (1991).

### Improvements to the Chow–Liu algorithm

Numerous papers have proposed improvements to the Chow–Liu algorithm and adaptations of it. For example, Alcobe (2002) gives an "incremental" method to implement the Chow–Liu algorithm in the case where data come in one at a time and you need to update the tree after each item of data. His computer experiments suggest that it is much faster than running the Chow–Liu algorithm again from scratch each time. Choi et al (2011) gives two algorithms for learning "latent" tree graphical models where some variables are unobserved, which is NP-hard. They prove that the algorithms are asymptotically consistent and report the results of numerical experiments. Wang (2009) is similar.

Meilă (1999) presents a way of speeding up the Chow–Liu algorithm if the data are sparse, by comparing some mutual informations without actually calculating them. Zaffalon & Hutter (2005) is an adaptation of the Chow–Liu algorithm that apparently gives results that are more robust to the random variation in the data, by using the "imprecise Dirichlet model" to model the prior uncertainty about the data, which are discrete-valued.

Pelleg & Moore (2006) present a way of speeding up the Chow–Liu algorithm for large datasets by maintaining confidence intervals on the edge-weights. Their maximum-weight spanning tree algorithm works down from the complete graph to a tree. When two edges need to be compared but their confidence intervals overlap, they look at more data to shrink one of them and make a decision. Naturally this method does not always find the optimal tree.

### Other research based on the Chow–Liu algorithm

Gupta et al (2010) is about learning forest graphical models using non-parametric kernel density estimates on each node and pair of nodes, using a method based on Chow & Liu's. Fleischer et al (2005) consider the NP-hard problem of finding the minimum spanning tree where both the edges and the "inner nodes" (the nodes that are not leaves) have weights.

In machine learning there has been research on “mixture of trees” models (Meilă & Jordan 2000), in which the joint density consists of a weighted sum of the densities of several tree distributions. Mixture-of-trees models can be regarded as having an unobserved variable that chooses one of the trees; each separate tree distribution is a conditional distribution given that the unobserved variable chose that tree. Meilă & Jordan (2000)’s method for learning a mixture-of-trees model from data is a combination of the EM algorithm, for the unobserved variable, and Chow–Liu, for each of the trees. See also Kollin & Koivisto (2006) or Kumar & Koller (2009).

Vincent Tan and his collaborators have produced several papers based on the Chow–Liu algorithm. In Tan et al (2010a) they calculate the “error exponent” for the maximum-likelihood estimator of the tree structure. Measuring the difficulty of learning a graph by the error exponent, they prove that the star graph (consisting of a hub and its spokes only) is the hardest to learn and the chain graph (with all the nodes in a line) is the easiest. See section 11.1 for other senses in which these graphs are extremal. Tan et al (2010c) is about hypothesis tests to decide which of two trees or forests a sample comes from, Tan et al (2010b) is about learning two tree graphical models for the purpose of classifying future observations into one of two categories, and Tan et al (2010d) is about learning forests for discrete graphical models by removing edges from the Chow–Liu tree.

# 8 Methods for factored distributions on trees

## 8.1 Introduction and the Matrix Tree Theorem

This chapter is about methods for analyzing factored distributions on trees, in particular factored posterior distributions. For a factored distribution on trees, Meilă & Jaakkola (2006) showed how to find the normalizing constant and certain other quantities in polynomial time, rather than by calculating the unnormalized probabilities of all the possible trees and summing them, which would be much slower. (Factored distributions were discussed in section 5.4. The main ideas in the chapter were mentioned briefly and without details in two presentations, Lauritzen 2006 and 2012.)

Meilă & Jaakkola (2006) presented their theorems in the context of Bayesian structure-learning for discrete-valued graphical models. Section 8.2 gives a summary of their relevant results and several new examples of questions they can be used to answer. In section 8.3 I show how these methods can be used for GGM structure-learning. Section 8.4 is a review of methods for generating trees and forests from factored distributions.

The methods in this chapter are based on the Matrix Tree Theorem (MTT), or more precisely a version of it that I will call the Weighted Matrix Tree Theorem (WMTT). MTT gives a way to calculate how many spanning trees a given graph has, and WMTT gives an explicit way of finding all the spanning trees. Section 8.5 describes the origins of these two theorems and includes references to publications that contain proofs of them.

***Matrix Tree Theorem (MTT).*** For an undirected graph $G = (V, E)$, where $V = \{1, \dots, p\}$, define the Laplacian matrix $L$ by

$$L_{ij} = \begin{cases} -1 & \text{if } (i,j) \in E \\ \deg(i) & \text{if } i = j \\ 0 & \text{otherwise.} \end{cases}$$

The number of spanning trees of $G$ equals the absolute value of any minor of $L$. (A minor of a matrix is the determinant of the matrix formed by removing one row and one column.)

***Weighted Matrix Tree Theorem (WMTT).*** Let $G$ be as above. On each edge $(i, j)$, put an indeterminate variable $x_{ij}$. Define $L$ by

$$L_{ij} = \begin{cases} -x_{ij} & \text{if } (i,j) \in E \\ \sum_{k:(i,k)\in E} x_{ik} & \text{if } i = j \\ 0 & \text{otherwise.} \end{cases}$$

Let $M$ be the absolute value of any of the minors of $L$, and for a spanning tree $T = (V, E_T)$, let $h(T) = \prod_{(i,j)\in E_T} x_{ij}$. Then $M = \sum_T h(T)$, where the sum is over all spanning trees of $G$.

In other words, each monomial term in $M$, when it is simplified, corresponds to one possible spanning tree. (A monomial is a product of powers of variables.) If $x_{ij} = 1$ for all $(i,j) \in E$ then $L$ is just the Laplacian matrix and WMTT reduces to MTT.

Here is an example of WMTT. It can also be regarded as an example of MTT, by replacing all the $x_{ij}$'s by 1. The graph is shown on the left in Figure 8.1. The matrix is

$$L = \begin{pmatrix} x_{12} + x_{14} & -x_{12} & 0 & -x_{14} \\ -x_{12} & x_{12} + x_{23} + x_{24} & -x_{23} & -x_{24} \\ 0 & -x_{23} & x_{23} + x_{34} & -x_{34} \\ -x_{14} & -x_{24} & -x_{34} & x_{14} + x_{24} + x_{34} \end{pmatrix},$$

and the absolute value of any of its minors will simplify to give

$$M = x_{12}x_{14}x_{23} + x_{12}x_{14}x_{34} + x_{12}x_{23}x_{24} + x_{12}x_{23}x_{34}$$
$$+ x_{12}x_{24}x_{34} + x_{14}x_{23}x_{24} + x_{14}x_{23}x_{34} + x_{14}x_{24}x_{34}.$$

For instance, the first monomial, $x_{12}x_{14}x_{23}$, corresponds to the spanning tree shown on the right in Figure 8.1. Setting all the $x_{ij}$'s to 1 gives $M = 8$, which is the number of spanning trees of the graph.

As another example, if $G$ is $K_3$, the complete graph on 3 nodes, then $M$ simplifies to $x_{12}x_{13} + x_{12}x_{23} + x_{13}x_{23}$. The first monomial, $x_{12}x_{13}$, corresponds to the tree with edges (1,2) and (1,3).

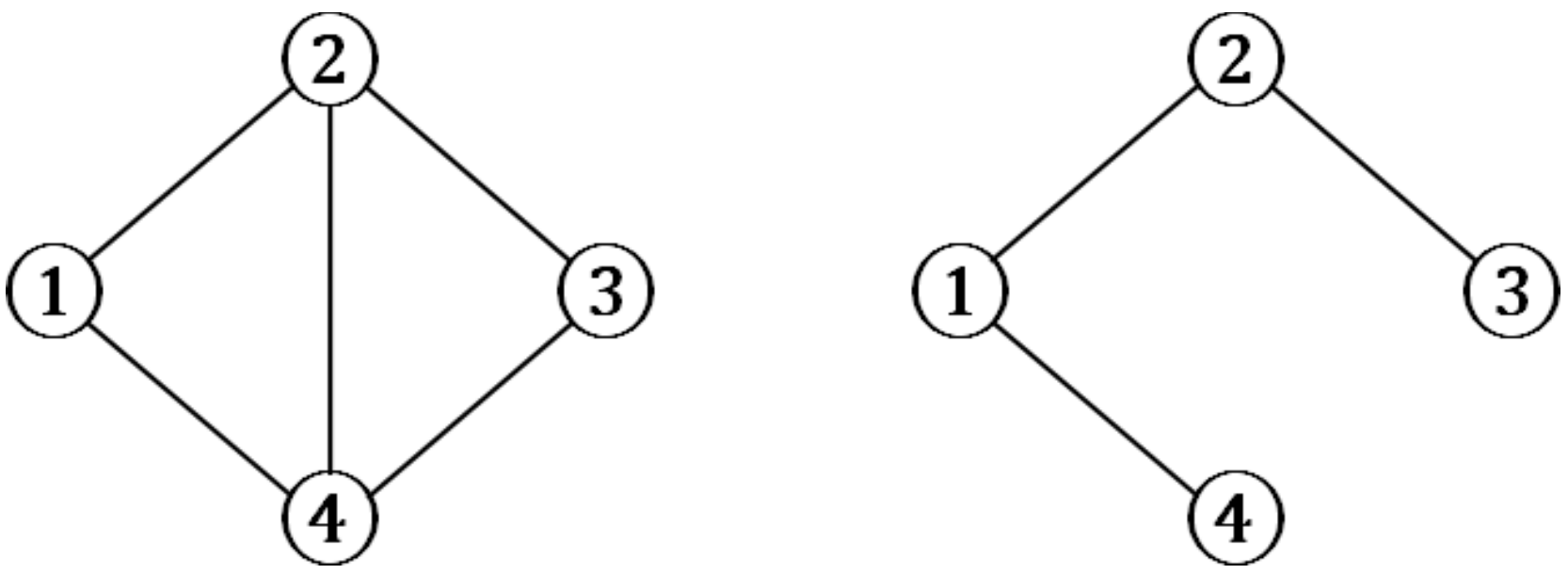

**Figure 8.1.** A graph (left), and the spanning tree of it that corresponds to $x_{12}x_{14}x_{23}$ in WMTT.

Many proofs of MTT work by first showing that the absolute values of the minors are all equal. They then use the Binet–Cauchy theorem (Lancaster & Tismenetsky 1985, section 2.5) to express one of the minors in terms of determinants of smaller matrices. These determinants are $\pm 1$ if the corresponding subgraph is a spanning tree and 0 otherwise.

## 8.2 The normalizing constant for discrete-valued tree graphical models

For factored distributions on the set of trees, Meilă & Jaakkola (2006) show how to calculate the normalizing constant in polynomial time, using WMTT. Without their method this would be impractical, since the obvious way to calculate this quantity requires summing over all possible trees, and the number of possible trees is super-exponential.

The main theorem follows from applying WMTT to the complete graph $K_p$. Suppose you have a factored distribution on trees, defined as in equation (1) in section 5.4. In WMTT, let each of the indeterminate variables $x_{ij}$ equal the corresponding edge-factor $w_e = w_{(i,j)}$ from the factored distribution. Now WMTT states that $M = \sum_T h(T)$, where the sum is over all the spanning trees of $G$, in other words all the trees. But

$$h(T) = \prod_{(i,j)\in E_T} x_{ij} = \prod_{e\in E_T} w_e \propto \mathbb{P}(T).$$

The normalizing constant for the factored distribution is

$$\sum_T \prod_{e\in E_T} w_e = \sum_T h(T) = M,$$

which can be calculated in polynomial time using standard algorithms for calculating determinants.

Meilă & Jaakkola (2006) is about Bayesian structure-learning for discrete-valued tree graphical models. Undirected tree graphical models are equivalent to rooted-tree DAG graphical models, and they use both forms. They show that if certain reasonable-sounding assumptions about the parameters of the discrete distribution are satisfied, then the prior distributions of these parameters must be a product of Dirichlet distributions. It then follows that if the prior distribution on the graph structure is factored, then so is the posterior. (The abstract says the posterior "can be completely determined analytically in polynomial time", but calculating the entire posterior distribution in polynomial time is impossible since the number of trees, and hence the amount of information in the posterior, is super-exponential.)

Being able to calculate the normalizing constant means that the posterior probability of any given tree can be calculated exactly. Meilă & Jaakkola (2006) also show that under a factored distribution, the expectations of real-valued "additive" or "multiplicative" functions of trees can be calculated quickly by using derivatives of the normalizing constant (expressed as a function of the edge-factors). A function $f$ is additive if it is of the form $f\big((V,E)\big) = \sum_{(u,v)\in E} f_{uv}$, where $f_{uv}$ are weights on the edges, and multiplicative if it is of the form $f\big((V,E)\big) = \prod_{(u,v)\in E} f_{uv}$.

I will now show that several useful quantities to do with factored posterior distributions can be calculated using very simple additive functions. Let $f = \mathbb{I}_{(i,j)\in E}$. This is the indicator function for the edge $(i,j)$. This is obviously additive, with

$$f_{uv} = \begin{cases} 1 \text{ if } \{u,v\} = \{i,j\} \\ 0 \text{ otherwise.} \end{cases}$$

The method of Meilă & Jaakkola (2006) can therefore be used to calculate $\mathbb{E}(f(E))$ in the posterior distribution. This is simply the posterior probability that the edge is in the graph, $\mathbb{P}((i,j) \in E)$, which could very easily be a quantity of interest.

Let $p_{ij} = \mathbb{P}((i,j) \in E)$, and suppose that this has been calculated for every possible edge. If the data was simulated from a distribution with a known graph structure $(V, E_{true})$, then the expected number of true-positives, another quantity that might be of interest, is $\sum_{(u,v)\in E_{true}} p_{uv}$. The expected number of false-positives is $\sum_{(u,v)\in E_{all}\setminus E_{true}} p_{uv}$, and other related quantities can be found in similar ways. (See section 10.2 for more on these quantities.)

The degree of $i$ is $\sum_{v\neq i} \mathbb{I}_{(v,i)\in E}$, so the expected degree of $i$ is $\mathbb{E}(\deg(i)) = \sum_{v\neq i} p_{vi}$. Alternatively, the expected degree of $i$ can be calculated directly, using

$$f_{uv} = \begin{cases} 1 & \text{if } u = i \text{ or } v = i \\ 0 & \text{otherwise.} \end{cases}$$

The value of the corresponding additive function $f$ is $\deg(i)$.

In these ways several quantities that might be of interest can be expressed in terms of the $p_{ij}$'s. One quantity that cannot is the expected maximum degree. Finding this requires quantities like $\mathbb{P}(\deg(i) = 1)$. This cannot be calculated using additive or multiplicative functions, because it is the expectation of $\mathbb{I}\{\deg(i) = 1\}$, which is not just the sum or product of fixed weights on the edges.

Other limitations of these methods are that they do not work with general prior distributions and they do not work with forests—essentially because there is no MTT for forests. Another possible drawback is that in Bayesian structure-learning it may be preferable to work with only a subset of the possible models, rather than average over all of them. Madigan & Raftery (1994), for example, argue that models with much lower probability than the best ones should be discarded completely.

## 8.3 The normalizing constant for GGMs

### How the methods work for GGMs

Meilă & Jaakkola (2006) mention that their results still work with GGMs and Bayesian learning of tree-structure but do not give any details. For GGMs, if the uniform graph prior on trees is used, the posterior is

$$\mathbb{P}(G \mid x) \propto (2\pi)^{-np/2} \prod_{v\in V} K(v) \prod_{(u,v)\in E} \frac{K(u,v)}{K(u)K(v)},$$

where $x$ is the $n \times p$ matrix of observed data and $K$ is defined in section 7.4. (The expression for $K$ involves $U = x^T x$ and the HIW hyperparameters, but $K$ is written as a function of one or two nodes for simplicity.)

This is clearly a factored distribution, with the weight on edge $(u, v)$ being $\frac{K(u,v)}{K(u)K(v)}$. It follows that all the facts and methods in the previous section can be used. For Bayesian structure-learning on forests, $\frac{K(u,v)}{K(u)K(v)}$ can be regarded as the Bayes factor for the presence of the edge $(u, v)$, as mentioned in section 7.4 and Lauritzen (2006, 2012).

If the graph prior is factored then the results in the previous section still hold. If the prior is

$$\mathbb{P}(G) \propto \prod_{e \in E} w_e ,$$

then the posterior is

$$\mathbb{P}(\, G \mid x \,) \propto \prod_{(u,v) \in E} w_{(u,v)} \frac{K(u, v)}{K(u)K(v)} ,$$

which is clearly a factored distribution.

In the computer experiments in section 11.5, Meilă & Jaakkola (2006)'s methods are used for finding the expected true-positive rate under the posterior distribution. The limitations of their methods are the same in the case of GGMs as they are for discrete-valued graphical models.

## A computer program for GGMs

I have written a computer program that uses the MTT-based method to find the normalizing constant, the expected degree of each node, and the expected number of true positives, for GGMs. I used this to produce the results in section 11.5. The input to the program is the data and the prior values of the HIW hyperparameters. But the input to the subroutines that actually perform the MTT-based methods is just the symmetric matrix of edge-factors that defines the posterior distribution. So the program could easily be adapted to work with any factored prior.

For $p = 30$, the normalizing constant is beyond the range of the usual double-precision floating-point numbers that are used by computers. But it can be found if you use special classes and packages for arbitrary- or high-precision decimals. In the Java programming language, objects of the BigDecimal class (Oracle 2012) are stored as $x \times 10^y$ where $x \in \{0,1, \dots\}$ and $y \in \{-2^{31}, -2^{31} + 1, \dots, 2^{31} - 1\}$. The size of $x$ is limited only by the size of the Java virtual machine, which in turn is limited by the host computer. Matrix algebra with BigDecimals can be done using any of the classes that implements the FieldMatrix interface in the Commons Math package (Apache Software Foundation 2012). These classes and packages are of a high quality, though arithmetic with them is naturally slower than with the usual floating-point decimals.

On the topic of high-precision arithmetic, Wang & Li (2012) say that their methods may need to calculate quantities such as $e^{-10000}$, but "to our knowledge, [no] current soft-ware for Gaussian graphical models has yet supported this level of precision." Perhaps my program for the MTT-based methods is the first. Lauritzen (2006) says certain algorithms for forests do not work well because most of the values are essentially zero. BigDecimal might be able to overcome these problems.

## 8.4 Generating random trees or forests

As discussed in sections 5.1 and 5.7, one of the main things you might want to do with a graph distribution is generate from it. This section is a review of methods for generating spanning trees or forests of a given graph according to a uniform or factored distribution. For Bayesian learning of tree graphical models, the given graph would be $K_p$. (Generating from uniform distributions has no purpose in graphical model structure-learning but is closely related to generating from factored distributions.)

Propp & Wilson (1998) give a history of the algorithms for generating a spanning tree uniformly at random. The first one was by Guénoche (1983) and a faster algorithm appears in Colbourn et al (1989). The basic idea is that if you repeatedly choose an edge uniformly at random and discard it if it creates a cycle, you will not get the uniform distribution. But if you go through the edges and accept each one according to the proportion of spanning trees that contain it, then you will. This proportion can be calculated using MTT. The same idea works for factored distributions, using WMTT—see Kulkarni (1990).

A different type of algorithm for generating a spanning tree was discovered by Broder (1989) and Aldous (1990). Do a simple random walk on the graph until you have visited every node. For each node apart from the first one, record the edge by which the node was first visited. The set of these edges constitutes a spanning tree chosen uniformly at random.

Propp & Wilson (1998) give two algorithms for generating from a uniform or factored distribution. One uses "coupling from the past", which is a way of generating exactly from the invariant distribution of an ergodic Markov chain that has a finite number of states. The other uses "cycle-popping".

Generating forests from a factored distribution is much more difficult. Dai (2008) presents two sets of algorithms for this problem. (He uses "forests" to mean subgraphs of a given graph that contain all its nodes and have no cycles.) The first set uses coupling from the past and the second set uses rejection methods. The main rejection-type algorithm is 8: add an extra node and edges from it to all the other nodes, run one of the algorithms for generating a random tree, remove the extra node and the edges that include it, and then accept this forest with a certain probability.

## 8.5 Supplementary notes: the history of MTT

Meilă & Jaakkola (2006) state that their main theorem was first proved in Jaakkola et al (2000) but that they later discovered a similar result, which must have been WMTT, in Harary (1967). The idea of using WMTT to find the normalizing constant for random trees appears implicitly in Kulkarni (1990). It was also conceived independently by Koo et al (2007), Smith & Smith (2007), and McDonald & Satta (2007), in the field of computational linguistics.

MTT and WMTT, or theorems that are essentially equivalent to them, were known in the 19th century and rediscovered multiple times in the 20th. Moon (1970, page 42) and Knuth (1997, pages 583 and 586) give detailed accounts of their origins. Kirchhoff (1847) is often credited with MTT or WMTT. His main result is essentially the same as WMTT, but it is about the dual problem of finding all the sets of edges that can be removed to leave a tree. A similar version appears in Maxwell (1892, pages 403–410). These publications are about electrical circuits and resistances, and some mental exertion is needed to interpret them as graph theory. In mathematics, a version of WMTT appears in Cayley (1856) and Sylvester (1857). Books that contain proofs of WMTT include Moon (1970) and Bollobás (1998, page 57), and of course MTT follows from WMTT.

The normalizing constant is called the "normalization constant" in Meilă & Jaakkola (2006). It is also known as the "partition function", for example in Murray & Ghahramani (2004).

# 9 Local moves in forests and trees

## 9.1 Preamble

Algorithmic graph theory (Even 1979, Gibbons 1985) is mostly about solving problems for given graphs. Typical problems are testing whether a graph is planar or colouring the nodes so that no two adjacent nodes have the same colour. Chapter 6 of Bondy & Murty (2008) is called "Tree-search algorithms". This includes breadth-first search, depth-first search, and algorithms to find minimum-weight spanning trees, shortest paths, and so on.

In contrast, this chapter is about algorithms for storing and manipulating graphs, with a view to exploring the posterior graph distribution in graphical model structure-learning. The main algorithms are designed for manipulating graphs by repeatedly adding and deleting edges. The main issue is how to store the graph in order to take advantage of the information from the previous step, avoid wasteful repeated searches through the graph, and enable the information that is stored to be updated in an efficient way.

In computer programs there are two common ways to store undirected graphs. The first is the adjacency matrix. This is usually a symmetric square matrix of 1s and 0s but can also be regarded and stored as a triangular matrix, or a square or triangular matrix of booleans. The other way is a list of edges. This is regarded as more suitable for sparse graphs, since it uses less memory. Forests and trees can be stored in a different way by regarding each component as a rooted tree, with arbitrary root, and storing just the parent of each node, or "null" if it is a root.

## 9.2 Storing forests and trees for local moves

### The purposes of the algorithms

As mentioned in section 3.1, in Bayesian analysis of the graph structure it is impossible to calculate the posterior probability of all the $2^{\binom{p}{2}}$ possible graphs on $p$ nodes. For decomposable graphs there are reversible-jump MCMC algorithms for sampling from the posterior distribution of the graph structure and the covariance matrix (Giudici & Green 1999, Green & Thomas 2013). Jones et al (2005) proposed a stochastic search algorithm for moving through the space of all possible graphs and calculating the exact posterior probabilities of the graphs that are visited. Restricted versions of these algorithms can be applied to forests, and adapted versions of them can be applied to trees. (Details are given in section 10.1.)

Consider a Bayesian analysis in which attention is restricted to forests. To explore the posterior distribution of the graph structure the most obvious, natural, and "local" type of move is to add or delete one edge at a time. For trees, the most obvious type of move is to move an edge. (I use the word "move" with two different meanings. For forests a move means adding or deleting an edge; for trees it means literally moving an edge from one position to another. I treat forests and trees completely separately, so there should be no ambiguity.)

For forests, it is easy to describe which edges can be added and removed while ensuring that the graph is still a forest. Any existing edge can be removed, and an edge can be added if and only if its two nodes are in different connected components. For trees, it is similarly easy to describe which moves are possible. First choose an edge, temporarily remove it, and identify the two connected components that result; the edge can then be put back between any two nodes that are not both in the same connected component. (Two alternative ways of making moves in trees are described in the last paragraph of this subsection.)

These conditions are easy to describe verbally, but they are less easy to program or write in the form of detailed algorithms, and they are time-consuming to carry out. To see whether a particular move is possible it is necessary to identify connected components. Identifying a connected component means doing a breadth-first search, or possibly a depth-first search, through the component (Golumbic 1980, pages 37–42; Cormen et al 2009, pages 594–612). This means finding all the neighbours of a node at each step. Finding several or all of the possible moves from the current graph, which is necessary for the algorithm of Jones et al (2005), would require doing all of this many times. It is efficient and elegant to be able to choose a move straight away, rather than having to choose one, test whether it is a legal move, and if not then reject it and repeat.

Another issue is to do with how to choose moves randomly for MCMC proposal distributions or other algorithms that explore the graph space. To achieve good mixing, it may be desirable to be able to choose a move uniformly at random from among all the possible moves.

Section 9.3 describes how to store a forest in such a way that it is easy to choose a legal move uniformly at random, and how to update the stored information after a move. Section 9.4 describes an analogous system for trees. These systems make it simple to program graph-search algorithms that choose these moves uniformly at random. They are computationally efficient because the update algorithms are "local"—they never need to search through all the nodes or all the edges. (On the other hand, the algorithms that store the graph, and check that it is a forest or tree, are not local and do search through all the nodes. But these usually only need to be done once.)

For exploring the space of decomposable graphs, Thomas & Green (2009a, b) state that it is desirable to be able to find a decomposable neighbouring graph straight away, not by choosing a random neighbour and then checking whether it is decomposable. (A "neighbour" of a graph is a graph formed by making one move from it.) The reason is that for large $p$ the former way should be much faster. This is essentially the same as the main reason behind my approach for forests and trees.

There are at least two alternative ways of making basic moves on trees. The first is in Propp & Wilson (1998, page 196) and is for rooted trees. Choose a node, other than the root, to be the new root; draw an edge from the new root to the old root; and delete the edge that goes in to the new root. The second is from Climaco et al (2008). Add an edge, identify the cycle that results, and then remove an edge from the cycle.

## How the algorithms are shown

In the following subsections, each algorithm is preceded by an explanation of what it does and how it works. The algorithms are written in a style that is intended to be easy to translate into computer code. The right-hand columns contain verbal descriptions of what is being done, where this is not completely obvious, and other comments.

- Algorithms V and VIII, for storing the graph and checking its properties, assume that the graph is supplied in the form of its adjacency matrix, $A$.
- $X \leftarrow Y$ means that $X$ is assigned the value $Y$.
- For loops, the scope is shown by indentation.
- Whereas in directed graphs $pa(v)$ is usually a set, here it is a single node, because all the connected components are rooted trees.

## Notation and partitions

The algorithms are written in pseudo-code or plain English rather than traditional set-theory notation. One reason for this is that they use partitions. A partition of a set $Z$ is a set $\{Z_1, \ldots, Z_k\}$ such that $Z_i \cap Z_j = \emptyset$ for all $i \neq j$ and $\bigcup_{i=1}^{k} Z_i = Z$. The $Z_i$'s are called "parts". Simple operations such as "move $v$ to a new part" are long and difficult to read when written in set-theory notation.

In programming, probably the most natural way to work with a partition is to store it as a pair of associative arrays. In one associative array, each key is an object (an element of $Z$), and the value associated with this key is the "label" of the part that the object is in. The labels can be positive integers. In the other associative array, each key is the label of a part, and the value associated with this key is the set of objects that are in this part. Queries of the form "which part is this object in?" and "which objects does this part contain?" can be answered quickly and easily since they each involve just a single look-up. When an object is moved from one part to another, both the associative arrays have to be updated.

## Facts about rooted trees

Here are several simple results about rooted trees that are used by the algorithms. As stated in section 2.1, in directed graphs I use "path" to mean "undirected path".

***Definition 9.1.*** A rooted tree is a directed tree in which one node is designated the root and the paths from the root to all the other nodes are directed paths. (In other words, all the edges point away from the root.)

Rooted trees can also be defined as directed trees with any of the following three properties. Proofs that the definitions are equivalent are omitted.

- The root is an ancestor of all the other nodes.

- The root has no parents, and all the other nodes have exactly one parent each.
- For each edge, the node nearer to the root is the parent and the node further from the root is the child (where "nearer" and "further" refer to the length of the path from the node to the root).

Rooted trees are especially easy to deal with in algorithms and computer programs. Together with each node are stored references to its children, and together with each node except the root is stored a reference to its parent. Obviously $ne(v) = \{pa(v)\} \cup ch(v)$. It is trivially easy to find the path from $v$ to the root—this is simply $v, pa(v)$, $pa(pa(v))$, ..., until you get to the root. It is easy to find all the descendants of $v$, by "fanning down" from $v$ to its children, then all their children, and so on—this is done in Algorithms VII and IX.

In Algorithms V–VII, for forests, each connected component of the graph is regarded as a rooted tree. In Algorithms VIII–IX, for trees, the whole graph is regarded as a rooted tree. The directions on the edges are just for the purpose of computational convenience. They do not have any meaning in the graphical models.

***Definition 9.2.*** In a directed graph $G$, a reverse-directed path $(u, u_1, \dots, u_k, v)$ is a path such that $(v, u_k, \dots, u_1, u)$ is a directed path in $G$.

Proposition 9.3 defines the "youngest common ancestor" of two nodes in a rooted tree and gives some of its properties.

***Proposition 9.3.*** For any two nodes $u$ and $v$ in a rooted tree, there is a unique node $w$ that has the following properties:

- $u, v \in \{w\} \cup de(w)$, and
- all nodes $x$ such that $u, v \in \{x\} \cup de(x)$ are on the path between $w$ and the root.

I will call $w$ the "youngest common ancestor" of $u$ and $v$. (Note that $w$ might be equal to $u$ or $v$, so it is not necessarily one of their ancestors.) It also has this property:

- $w$ is on the path between $u$ and $v$.

***Proof.*** Let $P$ be the reverse-directed path from $u$ to the root, and let $X = \{x \in V : u, v \in \{x\} \cup de(x)\}$. If $x \in X$ then $x \in \{u\} \cup an(u)$. This means there is a reverse-directed path from $u$ to $x$. Each node has at most one parent, so $x$ must lie on $P$. Therefore all elements of $X$ are in $P$. (It is not strictly true that "$X \subseteq P$", since $P$ is a sequence.)

Let $w$ be the first element of $P$ that is in $X$. Since $u, v \in \{w\} \cup de(w)$, it follows that $\{u, v\} \subset de(pa(w)) \subset de(pa(pa(w))) \subset \cdots \subset de(root)$. This says that all the subsequent elements of $P$ are also in $X$. So $X$ consists of the nodes on the reverse-directed path from $w$ to the root. This shows the existence and uniqueness of the node that has the first two properties.

As for the third property, let the unique path from $u$ to $w$ be $(u, u_1, \dots, u_k, w)$ and the unique path from $v$ to $w$ be $(v, v_1, \dots, v_l, w)$. None of the $u_i$'s can be the same as any of the $v_j$'s, because if $u_i = v_j$ then this node would be $w$. So the unique path from $u$ to $v$ is $(u, u_1, \dots, u_k, w, v_l, \dots, v_1, v)$, and this contains $w$. ☐

An alternative way of proving Proposition 9.3 is by noting that for any three nodes in a tree, the three paths between them have exactly one node in common. If the three nodes are taken to be $u$, $v$, and the root, then the node that the paths have in common is $w$.

***Proposition 9.4.*** Let $v$ be a node in a rooted tree. If you reverse all the edges on the path from the root to $v$, the result is a rooted tree with $v$ as its root.

***Proof.*** A tree is rooted at $r$ if and only if, for all $u \in V$, there is a directed path from $r$ to $u$. Consider any node $u \in V$, and let $w$ be the youngest common ancestor of $u$ and $v$ (where $v$ is the node mentioned in the proposition). All the ancestors of $v$ lie on the path from $r$ to $v$, so $w$ must lie on this path. After the edges are reversed, there is a new directed path from $v$ to $w$, and the old directed path from $w$ to $u$ is still there. So there is a directed path from $v$ to $u$, which means the new graph is a rooted tree with root $v$. All these statements still hold if any two or more of $r, u, v$, and $w$ are equal, or if two different pairs of them are equal. ☐

***Proposition 9.5.*** Suppose $B = (V_B, E_B)$ is a rooted tree with root $b$, $C = (V_C, E_C)$ is a rooted tree with root $c$, $V_B \cap V_C = \emptyset$, and $v \in V_B$. Let $D$ be the graph formed by combining $B$ and $C$ and adding the edge $(v, c)$, so $D = \big(V_B \cup V_C, E_B \cup E_C \cup (v, c)\big)$. Then $D$ is a rooted tree with root $b$.

***Proof.*** A node $u$ in $D$ is either in $V_B$ or in $V_C$. If $u \in V_B$, then there is obviously a unique directed path from $b$ to $u$ in $D$, because $B$ is a rooted tree. If $u \in V_C$, then a directed path from $b$ to $u$ in $D$ can be formed by combining, in order, the directed path from $b$ to $v$ (which exists because $B$ is a rooted tree), the edge $(v, c)$, and the directed path from $c$ to $u$ (which exists because $C$ is a rooted tree). ☐

## 9.3 The system for storing a forest

### The purpose of the system

This section describes how to store a forest in such a way that it is easy to choose an edge-removal or edge-addition move uniformly at random from all the possible moves. For this it is necessary to have available the set of edges that can be added and the set of edges that can be removed, so that one of these can be chosen uniformly at random. Any edge can be removed. The non-trivial issue is which edges can be added. This requires knowing whether two nodes are in the same connected component.

Each component of the forest is regarded as a rooted tree. There are three algorithms. Algorithm V is for storing a forest and checking that it is a forest, Algorithm VI is for adding an edge, and Algorithm VII is for removing an edge.

Reversing an edge is not a possible move, because this would violate the condition that the components are rooted (in all cases except components with two nodes). Moreover, the directions are only for computational convenience, so reversing an edge would not change the graphical model.

## What is stored

- The $p$ nodes. Each node $v$ stores references to its parent, $pa(v)$, and its children, $ch(v)$. (Some nodes do not have a parent, and some nodes have no children.)
- A partition of the nodes into connected components.
- A partition of the edges into the three parts $existing$, $addable$, and $nonaddable$.
- The bit-pattern that constitutes the lower triangle of the adjacency matrix.

The bit-pattern is the most compact way to store a graph. The method for updating it is trivial, so this is omitted from the algorithms. The point of storing the bit-pattern is that the user will probably want to keep a record of some or all of the graphs that are visited. For this it is not necessary to have all the detailed information about parents, children, and partitions, so just the bit-pattern can be used.

To choose a move uniformly at random from among all the possible moves, simply choose an edge uniformly at random from $addable \cup existing$. If the edge is in $addable$, do Algorithm VI, and if it is in $existing$, do Algorithm VII.

### Algorithm V: store a forest *G(V,E)*, and check that it is a forest

**Set the nodes' parents and children, create the partition of the nodes, and check that it is a forest:**

| Step | Comment |
|---|---|
| 1. $undiscovered \leftarrow V$ | |
| 2. $waitlist \leftarrow \emptyset$ | This will be the set of nodes that have been discovered but not dealt with. |
| 3. Do | Each iteration of this loop will deal with a new connected component. |
| 4.     Move an arbitrary node $root$ from $undiscovered$ to $waitlist$. | |
| 5.     $pa(root) \leftarrow null$ | This indicates that $root$ has no parent. |
| 6.     In the node partition, create a new part called $currentPart$, and add $root$ to it. | |
| 7.     While $waitlist \neq \emptyset$ | This loop does breadth-first search of the component. Each iteration visits (deals with) one node, namely $current$. |
| 8.         Remove an arbitrary node $current$ from $waitlist$. | |
| 9.         $ch(current) \leftarrow ne(current) \setminus pa(current)$ | To find $ne(current)$, use the adjacency matrix $A$. If $pa(current)$ is $null$, regard it as $\emptyset$. |
| 10.         If $ch(current) \nsubseteq undiscovered$, it is not a forest; exit. | |
| 11.         For each node $child \in ch(current)$, | |
| 12.             $pa(child) \leftarrow current$. | |
| 13.         Move the nodes in $ch(current)$ from $undiscovered$ to $waitlist$. | |
| 14.         Add the nodes in $ch(current)$ to $currentPart$. | |
| 15. Until $undiscovered = \emptyset$. | |

**Create the edge partition:**

```
Create the edge partition, with the three parts
   existing, addable, and nonaddable.
For each pair of nodes (u, v)
      if A_uv = 1, put (u, v) in existing
      else if u and v are in the same part of the
      node partition, put (u, v) in nonaddable
      else put (u, v) in addable.
```

## Algorithm VI: add an edge *(u,v)*

See Figure 9.1, in which $u$ is ② and $v$ is ①. When the new edge is added, it needs to be given a direction. Suppose it is directed from $v$ to $u$. In the original graph before the edge is added, let $oldComp$ be the component that contains $v$ and $youngComp$ be the component that contains $u$. In Figure 9.1, $oldComp$ is {⑧⑪①③⑤} and $youngComp$ is {⑨②⑥⑦}. When the new edge is added, $oldComp$ and $youngComp$ combine to form a new component.

***Proposition 9.6.*** If the edges on the path from $u$ up to the root of $youngComp$ are reversed, then the new component will be a rooted tree.

***Proof.*** By Proposition 9.4, reversing the edges on the path from $u$ to the root of $youngComp$ will make $youngComp$ be a rooted tree with root $u$. By Proposition 9.5, if you combine $oldComp$ and $youngComp$ and add the directed edge $(v, u)$, the result is a rooted tree. □

So to update the edge-directions, all that is necessary is to reverse the edges from $u$ up to the root of $youngComp$. The only nodes whose parents or children change are $u$, $v$, and the nodes on the path from $u$ to the root of $youngComp$. The parents and children of the nodes on this path are updated in the loop in lines 4–11, and the children of $v$ are updated in line 12.

The new edge changes to $existing$. This edge is changed to $nonaddable$ in one iteration of the nested loops in lines 13–15 and then changed to $existing$ in line 16. All the other possible edges between the two components change from $addable$ to $nonaddable$. This is done in lines 13–15. The update of the node partition is obvious and is done in line 17.

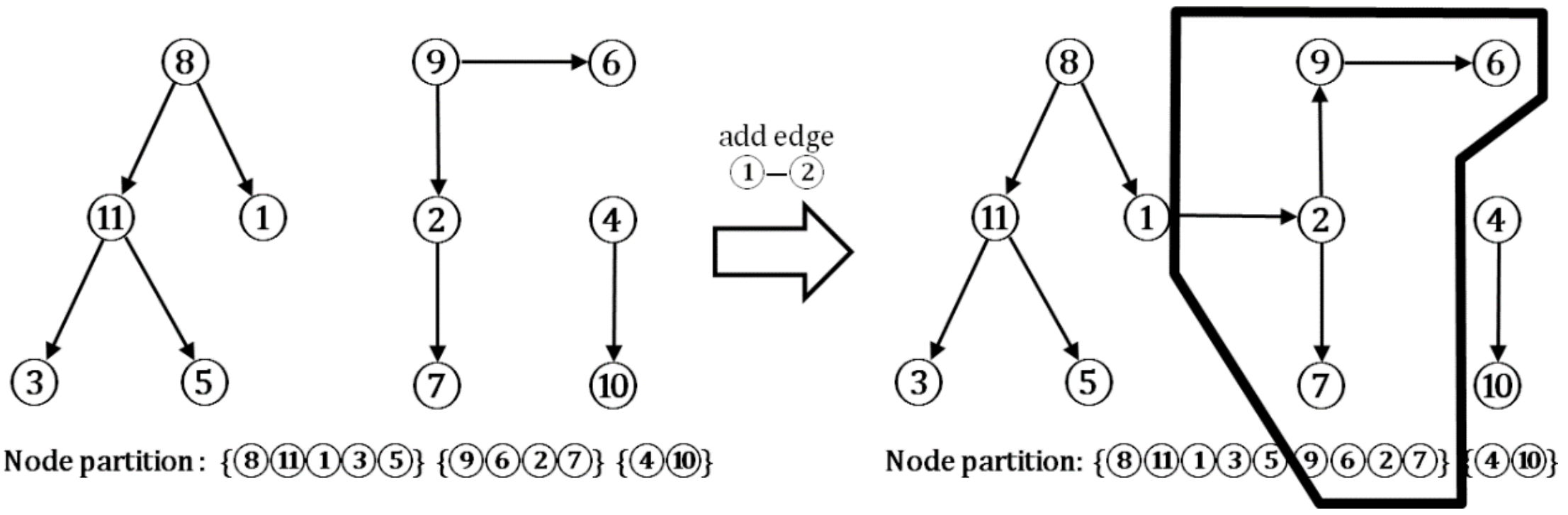

**Figure 9.1.** Algorithm VI, for adding an edge. The new edge can be oriented either way. To update the parents and children, go up from the new child ② to the root ⑨ while reversing the arrows. To update the node partition, move all the nodes from the part that contains the new child, ②, to the part that contains the new parent, ①. The changes are enclosed by the thick line.

## Algorithm VI: add an edge *(u,v)*

| | |
|---|---|
| **Check that adding the edge does not create a cycle:** | |
| 1. Check that $(u, v) \in addable$. | |
| **Update the nodes' parents and children:** | |
| 2. $current \leftarrow u$ | $v$ will be the parent of $u$. |
| 3. $previous \leftarrow v$ | |
| 4. Do | This loop goes "up" from $u$ and reverses all the arrows. Each iteration deals with one node, $current$. |
| 5.     Unless this is the first iteration, remove $previous$ from $ch(current)$. | |
| 6.     $next \leftarrow pa(current)$ | |
| 7.     $pa(current) \leftarrow previous$ | |
| 8.     If $next = null$, break from the loop. | The former root must have been reached. |
| 9.     Add $next$ to $ch(current)$. | |
| 10.     $previous \leftarrow current$ | |
| 11.     $current \leftarrow next$ | |
| 12. Add $u$ to $ch(v)$. | |
| **Update the edge partition:** | |
| 13. For each node $w$ in the same part as $u$ | |
| 14.     For each node $x$ in the same part as $v$ | |
| 15.         Move $(w, x)$ from $addable$ to $nonaddable$. | |
| 16. Move $(u, v)$ from $nonaddable$ to $existing$. | |
| **Update the node partition:** | |
| 17. Move all the nodes in $u$'s part to $v$'s part. | |

## Algorithm VII: remove an edge *(u,v)*

The first step in removing the edge $(u, v)$ is to rename $u$ and $v$ as $parent$ and $child$, in the appropriate order. See Figure 9.2, in which $parent$ is ⑧ and $child$ is ⑪. None of the edge-directions will change. The only nodes whose parents or children change are $parent$ and $child$. This is done in lines 4–5. The node partition is identified by "fanning down" from $child$ to identify its new connected component. This is a breadth-first search and is done in lines 7–11.

The edge that is removed changes from $existing$ to $addable$. All the edges between the two new components change from $nonaddable$ to $addable$. These updates are done in lines 12–15.

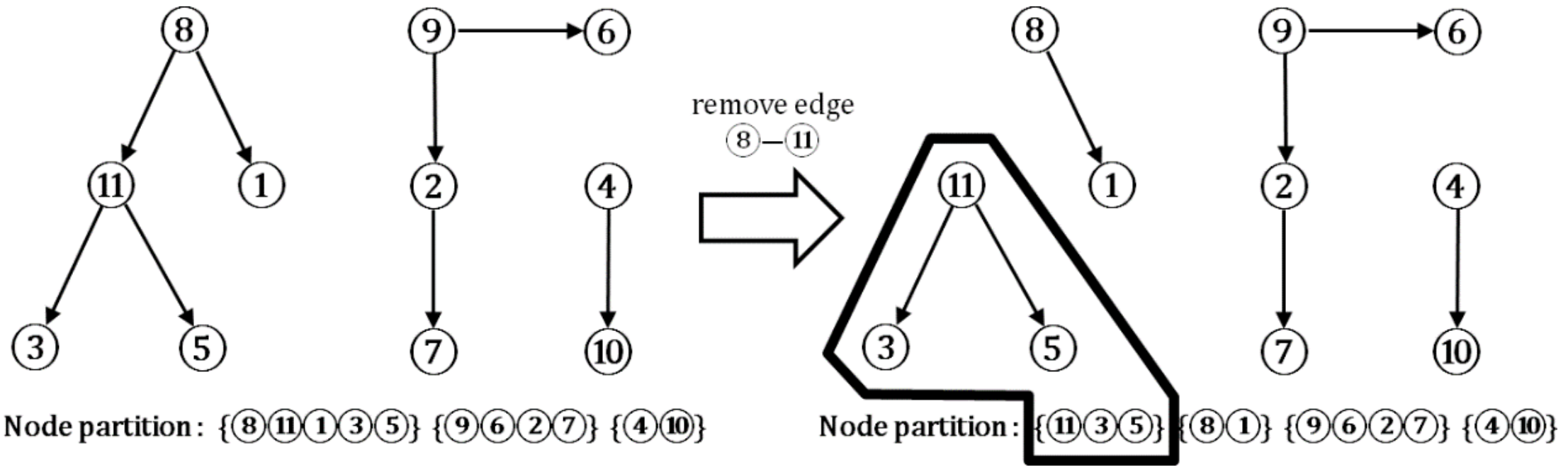

**Figure 9.2.** Algorithm VII, for removing an edge. To update the node partition, fan down from the newly orphaned node, ⑪, to all its descendants, and move all these nodes to a new part. The changes are enclosed by the thick line.

## Algorithm VII: remove an edge *(u,v)*

| | |
|---|---|
| **Check that the edge can be removed:** | |
| 1. Check that $(u, v) \in existing$. | |
| **Update the nodes' parents and children:** | |
| 2. If $pa(u) = v$,<br>$parent \leftarrow v$; $child \leftarrow u$ | Find which is the parent and which is the child. |
| 3. else<br>$parent \leftarrow u$; $child \leftarrow v$ | |
| 4. $pa(child) \leftarrow null$ | |
| 5. Remove $child$ from $ch(parent)$. | |
| **Update the node partition:** | |
| 6. Move $child$ to a new part, $newPart$ | |
| 7. $waitlist \leftarrow \{child\}$ | |
| 8. While $waitlist \neq \emptyset$ | This loop "fans down" from $child$ to find all its descendants and put them in $newPart$. |
| 9. Remove an arbitrary node $current$ from $waitlist$. | |
| 10. Add $ch(current)$ to $waitlist$. | |
| 11. Move all of $ch(current)$ to $newPart$. | |
| **Update the edge partition:** | |
| 12. Move $(u, v)$ from $existing$ to $nonaddable$. | This is temporary. |
| 13. For each node $u$ in $newPart$ | |
| 14. For each node $v$ in the same part as $parent$ | |
| 15. Move $(u, v)$ from $nonaddable$ to $addable$ | |

# 9.4 The system for storing a tree

## The purpose of the system

This section describes how to store and update a tree in such a way that it is easy to select edge-moves uniformly at random from among all the possible edge-moves. Algorithm VIII is for storing a tree and checking that it is a tree, and Algorithm IX is for choosing an edge-move uniformly at random and then updating the information that is stored.

## Choosing an edge-move uniformly at random

Choosing an edge-move consists of choosing an edge, removing it, and then choosing where to reinsert it. If at the initial step you choose the edge uniformly at random from among all the existing edges, then not all edge-moves will be equally likely. Consider the tree in Figure 9.3. If you choose and remove edge $A$, then there are 6 places it can be reinserted (while making sure the graph is still a tree); so choosing $A$ is the first step in 6 possible edge-moves. But if you choose and remove edge $B$, there are $4 \times 3 = 12$ places it can be reinserted; choosing $B$ is the first step in 12 possible edge-moves. To choose the edge-move uniformly at random, you need to be twice as likely to choose $B$ as to choose $A$.

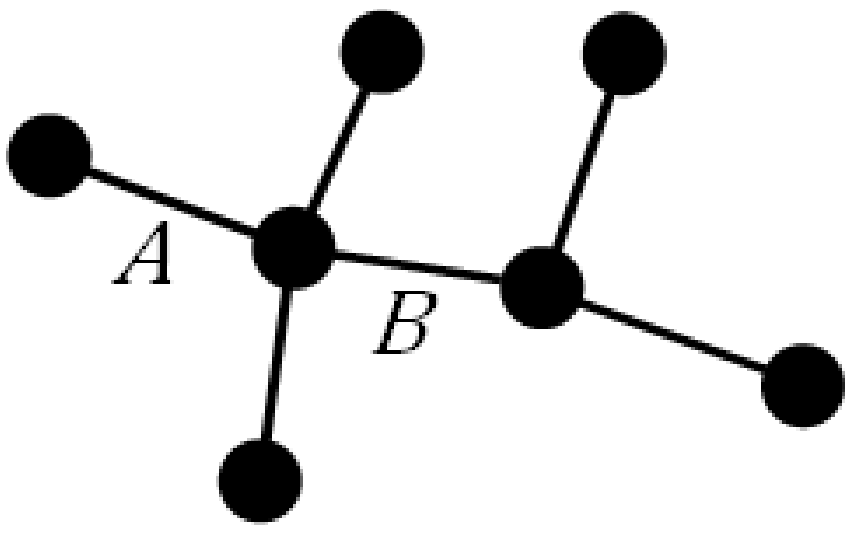

**Figure 9.3.** A tree. To choose an edge-move uniformly at random from among all the possible edge-moves, you need to have $\mathbb{P}(\text{choose edge } B) = 2\mathbb{P}(\text{choose edge } A)$.

In general, to be able to choose an edge-move uniformly at random, it is necessary to know for each edge the sizes of the two connected components that would result from removing that edge. The most convenient way to store this information is by assigning a "weight" to each node except the root. The weight of each node is the number of its descendants plus one and will be denoted by $W(\cdot)$. If the edge between $v$ and $pa(v)$ is removed, then the sizes of the two connected components are $W(v)$ and $p - W(v)$, and the edge can be reinserted in any of $W(v) \times (p - W(v))$ possible places.

So the procedure to choose an edge-move uniformly at random is as follows. Choose a node at random, with the probability of node $v$ being proportional to $W(v) \times (p - W(v))$. Remove the edge between $v$ and $pa(v)$. From each of the two connected components, choose one node uniformly at random. Finally, reinsert the edge between these two nodes.

The most complicated part of this system is the updates of the node-weights after the edge-move. This is the main work of Algorithm IX.

After an edge is removed, the graph consists of two connected components. Hereafter the component that contains the root will be called $oldComp$ and the other component will be called $youngComp$.

## Using uniformly chosen edge-moves

Choosing an edge-move uniformly at random could be useful in a tree version of Giudici & Green (1999)'s MCMC method or Jones et al (2005)'s stochastic search method. In one form of the latter method, all the neighbouring graphs are analyzed, so there is no need to choose edge-moves uniformly at random. (It would still be convenient to store the tree as a rooted tree, but there is no need for the weights.) In the general form, only some of them are analyzed. It is easy to adapt Algorithm IX to produce not just one but any number of edge-moves uniformly at random from among all the possible edge-moves.

The idea of choosing edge-moves uniformly at random is that this may give better mixing among the possible graphs. For example, if you remove an edge that includes a leaf (a node of degree 1), then that node will definitely still be a leaf after the edge is reinserted. Section 11.1 will show that if you choose the edge to move uniformly at random, rather than choose the edge-move uniformly at random, then leaves are more likely to remain leaves. Section 11.2 presents the results of experiments to see when choosing edge-moves uniformly at random is beneficial.

## Two slightly different versions of the system

With the system described above, there is positive probability that the edge will be reinserted in the same place as it was removed from, so the "edge-move" will consist of the graph staying the same. It is easy to adapt the method and the algorithms to avoid this. When you calculate the probabilities, use $W(v) \times \big(p - W(v)\big) - 1$ instead of $W(v) \times (p - W(v))$, and when deciding where to reinsert the edge, exclude the original position of the edge. (In the computer experiments in chapter 11, I use this adapted version.)

It would also be possible to store the weights on the edges rather than the nodes. The weight of each edge would be the number of nodes that are "downstream" of it—that is, on the same side of the edge as the child. To convert from node-weights to edge-weights, from each node $v$ remove $W(v)$ and put it on $(v, pa(v))$ instead. To choose an edge-move uniformly at random, choose edge $e$ with probability proportional to $W(e) \times (p - W(e))$.

With weights on the edges, there is no need to treat the root as a special case. The updates of the weights in $oldComp$ are also simpler to describe, since there is no need to talk about a path to "just before" $commonAncestor$. Overall, using edge-weights is more natural than using node-weights. But the differences are trivial, and with each node you already have to store the parents and children, so in practice it is simpler to use node-weights.

## What is stored

- The $p$ nodes. The tree is regarded as a rooted tree. Each node $v$ stores references to its parent, $pa(v)$, and its children, $ch(v)$. The first node is the root and has no parent; some nodes have no children. The root never changes.
- Each node except for the root also stores a weight, which is the number of its descendants plus one.
- The bit-pattern that constitutes the lower triangle of the adjacency matrix. As with forests, the storing and updating of the bit-pattern are omitted from the algorithms.

## Algorithm VIII: store a tree, and check that it is a tree

The loop in lines 6–13 of Algorithm VIII is exactly the same as the inner loop of Algorithm V (lines 7–14). It fans down from *root* to identify one connected component of the graph. This time there are two ways the graph could fail to be a tree. Firstly, it might have cycles, which is tested in line 9. Secondly, this one component might not include all the nodes, which is tested in line 14.

### Algorithm VIII: store a tree, and check that it is a tree

| | |
|---|---|
| **Set the nodes' parents and children and check that it is a tree:** | |
| 1. $undiscovered \leftarrow V$ | |
| 2. $waitlist \leftarrow \emptyset$ | |
| 3. Move an arbitrary node $root$ from $undiscovered$ to $waitlist$. | |
| 4. $pa(root) \leftarrow null$ | |
| 5. $nodesFound \leftarrow 1$ | |
| 6. While $waitlist \neq \emptyset$ | This loop fans down from $root$ to all its descendants, which hopefully means the entire graph. |
| 7.     Remove an arbitrary node $parent$ from $waitlist$ | |
| 8.     $ch(parent) \leftarrow ne(parent) \setminus pa(parent)$ | To find $ne(parent)$, use $A$. |
| 9.     If $ch(parent) \nsubseteq undiscovered$, it is not a tree; exit. | This tests for cycles. |
| 10.     For each node $child \in ch(parent)$ | |
| 11.         $pa(child) \leftarrow parent$ | |
| 12.     Move $ch(parent)$ from $undiscovered$ to $waitlist$. | |
| 13.     $nodesFound \leftarrow nodesFound + \lvert ch(parent) \rvert$ | |
| 14. If $nodesFound \neq p$, it is not a tree; exit. | This tests whether all the nodes have been found. |
| **Calculate the node-weights:** | |
| 15. $findWeight(root)$ | The subroutine $findWeight$ is immediately below. |
| 16. Discard $W(root)$ | |
| **Recursive subroutine $\boldsymbol{findWeight(v)}$:** | |
| i. $W(v) \leftarrow 1$ | This 1 counts the node itself. |
| ii. For each node $child \in ch(v)$ <br> iii. $\quad W(v) \leftarrow W(v) + findWeight(child)$ <br> iv. Return $W(v)$ | This loop calculates $W(v)$ and ensures that the weights of $v$'s children will be calculated. |

## Notation for Algorithm IX

Algorithm IX chooses and makes an edge-move, and updates the edge-directions and node-weights as necessary. Suppose the edge is moved from $(oldParent, oldChild)$ to $(newParent, newChild)$. Of course it is possible that $oldParent = newParent$ or $oldChild = newChild$.

After $(oldParent, oldChild)$ is removed, the graph has two connected components. Call the component that contains the root $oldComp$ and the other component $youngComp$. (The nodes in $youngComp$ might actually be "older" on average than the nodes in $oldComp$, if the "age" of a node is such that each parent is 1 older than its children, but this does not matter.)

To preserve the rootedness of the tree, the direction of the new edge has to be from $oldComp$ to $youngComp$. So $newParent \in oldComp$ and $newChild \in youngComp$.

## Facts used by Algorithm IX

Figure 9.4 shows a typical edge-move and how the node-weights and edge-directions change. This is intended to give an intuitive understanding of the propositions in this section and how Algorithm IX works.

Figure 9.5 shows the eight different possibilities for the relative positions of $root$, $oldParent$, and $newParent$ in $oldComp$. For example, if $root$ is on the path between $oldParent$ and $newParent$, then $oldComp$ looks like (c) in Figure 9.5; the configuration in Figure 9.4 is a special case of (d) in Figure 9.5.

The propositions and proofs below all hold completely generally, whichever of the possibilities in Figure 9.5 holds for $oldComp$, and even if $newChild = oldChild$. For example, if $newChild = oldChild$ then Proposition 9.7 simply states that no edge-directions need to be changed.

Recall that because the graph is a tree, the path between any two nodes is unique.

***Proposition 9.7.*** If the edges on the path between $newChild$ and $oldChild$ are reversed, the graph that results will be a rooted tree whose root is the root of $oldComp$.

***Proof.*** First note that $oldChild$ must be the root of $youngComp$. This is because $pa(oldChild)$ used to be $oldParent$, but the edge $(oldParent, oldChild)$ has been removed; so in $youngComp$, $oldChild$ has no parent, which means it must be the root. The proof then follows from Proposition 9.4 and Proposition 9.5 in the same way that Proposition 9.6 does. Reversing the edges makes $youngComp$ into a tree rooted at $newChild$ , and relinking $oldComp$ and $youngComp$ with the new edge $(newParent, newChild)$ then makes a rooted tree whose root is the root of $oldComp$. ☐

***Proposition 9.8.*** In $oldComp$, the only nodes whose weights can possibly change are the ones on the path between $newParent$ and $oldParent$.

***Proof.*** For conciseness I will sometimes regard a path as a set of nodes rather than a sequence. The proof will consist of gradually narrowing down the set of nodes whose weights can possibly change. Let $v \in oldComp$. The weight of $v$ only changes if $de(v)$ changes, where $de(v) = \{u \in V\text{: there exists a directed path from } v \text{ to } u\}$. Let $de_{old}(v)$ be the descendants of $v$ in the old graph, before the edge is moved, and $de_{new}(v)$ be its descendants in the new graph, after the edge is moved. Note that $an(v)$ does not change when the edge is moved, so there is no need for any subscript on it.

It is sufficient to consider nodes in $F = \{oldParent\} \cup an(oldParent) \cup \{newParent\} \cup an(newParent)$, since $de_{old}(v) \neq de_{new}(v)$ is only possible if $v \in F$. To see this, first note that $de_{old}(v) \neq de_{new}(v)$ means there exists some $u$ such that either $u \in de_{old}(v)$ and $u \notin de_{new}(v)$ or $u \in de_{new}(v)$ and $u \notin de_{old}(v)$. If the former holds, then the directed path from $v$ to $u$ in the old graph must include the edge $(oldParent, oldChild)$, which implies that $v \in \{oldParent\} \cup an(oldParent)$. If the latter holds, then the directed path from $v$ to $u$ in the new graph must include $(newParent, newChild)$, which implies that $v \in \{newParent\} \cup an(newParent)$. Combining these two possibilities shows that $de_{old}(v)$ can only differ from $de_{new}(v)$ if $v \in F$.

However, the proposition makes no claim about $oldParent$ or $newParent$ themselves. So it is sufficient to consider nodes in $G = an(oldParent) \cup an(newParent)$. Let $commonAncestor$ be the youngest common ancestor of $oldParent$ and $newParent$. By Proposition 9.3, $an(oldParent)$ consists of $an(commonAncestor)$ and the directed path from $w$ to $oldParent$, and $an(newParent)$ consists of $an(commonAncestor)$ and the directed path from $w$ to $newParent$. So $G$ is the union of $an(commonAncestor)$, the path from $commonAncestor$ to $oldParent$, and the path from $commonAncestor$ to $newParent$. The union of these two paths is the path from $oldParent$ to $newParent$ (this holds even if $commonAncestor = oldParent$ or $newParent$). So $G$ is the union of $an(commonAncestor)$ and the path from $oldParent$ to $newParent$.

To prove the proposition it therefore suffices to check that if $v \in an(commonAncestor)$ then $de_{old}(v) = de_{new}(v)$. Suppose $v \in an(commonAncestor)$. None of the ancestor–descendant relationships in $oldComp$ change when the edge is moved, so $de_{old}(v) \cap oldComp = de_{new}(v) \cap oldComp$.

If $u \in de_{old}(v) \cap youngComp$, then in the old graph there must be a directed path $(v, \ldots, commonAncestor, \ldots, oldParent, oldChild, \ldots, u)$. Here it is possible for any of the ellipses to signify no nodes (for example, if $v = pa(commonAncestor)$ then the first ellipsis disappears); it is even possible that $commonAncestor = oldParent$ or $oldChild = u$, in which case the path "collapses" in the obvious way; but it is not possible that $v = commonAncestor$. By Proposition 9.3 there is a directed path from $commonAncestor$ to $newParent$, and by Proposition 9.7 there is a directed path in the new graph from $newChild$ to $u$. So in the new graph there is a directed path $(v, \ldots commonAncestor, \ldots, newParent, newChild, \ldots, u)$, in which similar "collapsings" are possible. The existence of this path shows that $u \in de_{new}(v) \cap youngComp$. A similar argument shows the converse, that if $u \in de_{new}(v) \cap youngComp$ then $u \in de_{old}(v) \cap youngComp$. Therefore $de_{old}(v) \cap youngComp = de_{new}(v) \cap young$-$Comp$. Putting this together with $de_{old}(v) \cap oldComp = de_{new}(v) \cap oldComp$ shows that $de_{old(v)} = de_{new}(v)$. ☐

***Proposition 9.9.*** In $youngComp$, the only nodes whose weights can possibly change are the ones on the path between $newChild$ and $oldChild$.

***Proof.*** The nodes that are not on the path between $newChild$ and $oldChild$ are all on the ends of arrows that emanate from nodes on that path, or in tree structures on the ends of these arrows. The descendants of these nodes consist entirely of other nodes in these tree structures, and these sets of descendants do not change when the edge is moved. ☐

For an illustration of Propositions 9.8 and 9.9, see Figure 9.4, in which $youngComp$ is the right part of the two graphs and the nodes that are not on the paths mentioned in the propositions are white.

***Proposition 9.10.*** The weight of $newChild$ changes to $|youngComp|$. For the other nodes on the path from $newChild$ to $oldChild$, the weight changes to $|youngComp| - x$, where $x$ is the original weight of the previous node on this path.

***Proof.*** After the edge-move there is an edge from $newParent$ to $newChild$. So all the other nodes in $youngComp$ must be descendants of $newChild$, and the weight of $newChild$ is therefore $|youngComp|$. Next consider a node $v$ on the path from $newChild$ to $oldChild$ (for example $v =$ ❼ in Figure 9.4). After the edge-directions are updated, the edge going into $v$ comes from the previous node on this path (in this case, ⓭). So the descendants of $v$ consist of all the nodes in $youngComp$ except for $v$ itself and the nodes on the other side of this edge. The number of nodes on the other side of the edge is the original weight of the previous node on the path; call this $x$. So the new weight of $v$ is $|youngComp| - x$. ☐

To describe the updates for the nodes on the path between $newParent$ and $oldParent$, it is necessary to split this path into two parts. As before, let $commonAncestor$ be the youngest common ancestor of $oldParent$ and $newParent$. (Figure 9.5 shows the eight possibilities for the relative positions of these nodes; the arguments hold in all cases.) Consider separately the path from $newParent$ to $commonAncestor$ and the path from $oldParent$ to $commonAncestor$.

***Definition 9.11.*** If the path between $u$ and $v$ is $(u, u_1, \ldots, u_k, v)$, then the path from $u$ to "just before" $v$ is $(u, u_1, \ldots, u_k)$. If $k = 1$ then this is $(u, u_1)$, if the path between $u$ and $v$ is just $(u, v)$ then it is $(u)$, and if $u = v$ then it is $\emptyset$.

***Proposition 9.12.*** For the nodes on the path from $newParent$ to just before $commonAncestor$, the weight increases by $|youngComp|$.

***Proof.*** Consider a node $v$ on this path (for example, $v =$ ⓮ in Figure 9.4). The nodes in $youngComp$ are not descendants of $v$ before the move, but they are after. So the weight of $v$ increases by $|youngComp|$. ☐

***Proposition 9.13.*** For the nodes on the path from $oldParent$ to just before $commonAncestor$, the weight decreases by $|youngComp|$.

***Proof.*** Consider a node $v$ on this path (for example, $v =$ ❻ in Figure 9.4). The nodes in $youngComp$ are descendants of $v$ before the move, but not after. So the weight of $v$ decreases by $|youngComp|$. ☐

***Proposition 9.14.*** The weight of $commonAncestor$ does not change.

***Proof.*** This node has the same descendants before and after the move. ☐

## Algorithm IX: choose and make an edge-move

Line 1 decides which edge to move, lines 3–4 update $pa(oldChild)$ and $ch(oldParent)$, and lines 5–12 choose where to move the edge to. Lines 7–9 is a breadth-first search that identifies all the descendants of $oldChild$ and puts them in $youngComp$.

Lines 16–26 traverse the path up from $newChild$ to $oldChild$, updating the node-weights as described in Proposition 9.10 and reversing the arrows. The paths from $oldParent$ and $newParent$ to $commonAncestor$ cannot immediately be identified. Lines 27–32 identify the path from $oldParent$ up to the root. Lines 33–36 then go up from $newParent$ to this path, updating the node-weights according to Proposition 9.12. Line 37 identifies $commonAncestor$. Lines 38–41 then go up from $oldParent$ to just before $commonAncestor$, updating the node-weights according to Proposition 9.13.

Finally, lines 42–43 update $ch(oldParent)$ and $ch(newParent)$.

### Algorithm IX: choose and make an edge-move

| | |
|---|---|
| **Choose which edge to move, and remove it:** | |
| 1. Choose a node $oldChild$ at random; the probability of choosing $v$ is proportional to $W(v) \times (p - W(v))$. | The edge to be removed will be $(oldChild, oldParent)$. |
| 2. $oldParent \leftarrow pa(oldChild)$ | |
| 3. $pa(oldChild) \leftarrow null$ | |
| 4. Remove $oldChild$ from $ch(oldParent)$. | |
| **Choose where to reinsert the edge:** | |
| 5. $youngComp \leftarrow \{oldChild\}$ | $youngComp$ will be the component |
| 6. $waitlist \leftarrow \{oldChild\}$ | that currently contains $oldChild$. |
| 7. While $waitlist \neq \emptyset$ | This loop "fans down" from |
| 8. Remove an arbitrary node $current$ from $waitlist$. | $oldChild$ and puts all its descendants in $youngComp$. |
| 9. Put $ch(current)$ in $waitlist$ and $youngComp$. | |
| 10. $oldComp \leftarrow V \setminus youngComp$ | $oldComp$ is the component that |
| 11. Choose a node $newChild$ uniformly at random from $youngComp$. | contains $oldParent$ and the root. |
| 12. Choose a node $newParent$ uniformly at random from $oldComp$. | The edge will be reinserted at $(newChild, newParent)$. |
| **Traverse the path up from $newChild$ to $oldChild$, updating the node-weights and reversing the arrows:** | |
| 13. $current \leftarrow newChild$ | |
| 14. $previous \leftarrow newParent$ | |
| 15. $x \leftarrow 0$ | In the loop, $x$ will be the former |
| 16. Do | weight of the previous node. |
| 17. $temp \leftarrow W(current)$ | $temp$ is temporary and can be |
| 18. $W(current) \leftarrow \lvert youngComp \rvert - x$ | discarded after line 19. |
| $x \leftarrow temp$<br>$next \leftarrow pa(current)$<br>Remove $previous$ from $ch(current)$<br>$pa(current) \leftarrow previous$<br>If $current = oldChild$, break from the loop.<br>Add $next$ to $ch(current)$.<br>$previous \leftarrow current$<br>$current \leftarrow next$ | |
| **Identify the path from $\boldsymbol{oldParent}$ to the root:** | |
| $path \leftarrow \emptyset$<br>$current \leftarrow oldParent$<br>Do<br>Add $current$ to $path$.<br>$current \leftarrow pa(current)$<br>Until $current = null$ | $path$ does not need to be ordered; it can just be an ordinary set. |
| **Go up from $\boldsymbol{newParent}$ till just before you meet $\boldsymbol{path}$, updating the node-weights along the way:** | |
| $current \leftarrow newParent$<br>While $current \notin path$<br>$W(current) \leftarrow W(current) + \lvert youngComp \rvert$<br>$current \leftarrow pa(current)$<br>$commonAncestor \leftarrow current$ | |
| **Go up from $\boldsymbol{oldParent}$ to just before $\boldsymbol{commonAncestor}$ and update the node-weights:** | |
| $current \leftarrow oldParent$<br>While $current \neq commonAncestor$<br>$W(current) \leftarrow W(current) - \lvert youngComp \rvert$<br>$current \leftarrow pa(current)$ | |
| **Update the children of $\boldsymbol{oldParent}$ and $\boldsymbol{newParent}$:** | |
| Remove $oldChild$ from $ch(oldParent)$.<br>Add $newChild$ to $ch(newParent)$. | |

When choosing which edge to move, $W(v)$ is not used directly; what is used instead is $g(W(v)) = W(v) \times (p - W(v))$. So it might seem better to store $g(W(v))$ instead of $W(v)$. However, this is not possible, because $g$ is not invertible. Specifically, $W(v)$ will sometimes get updated to $W(v) + y$, and from $g(W(v))$ it is not possible to calculate $g(W(v) + y)$.

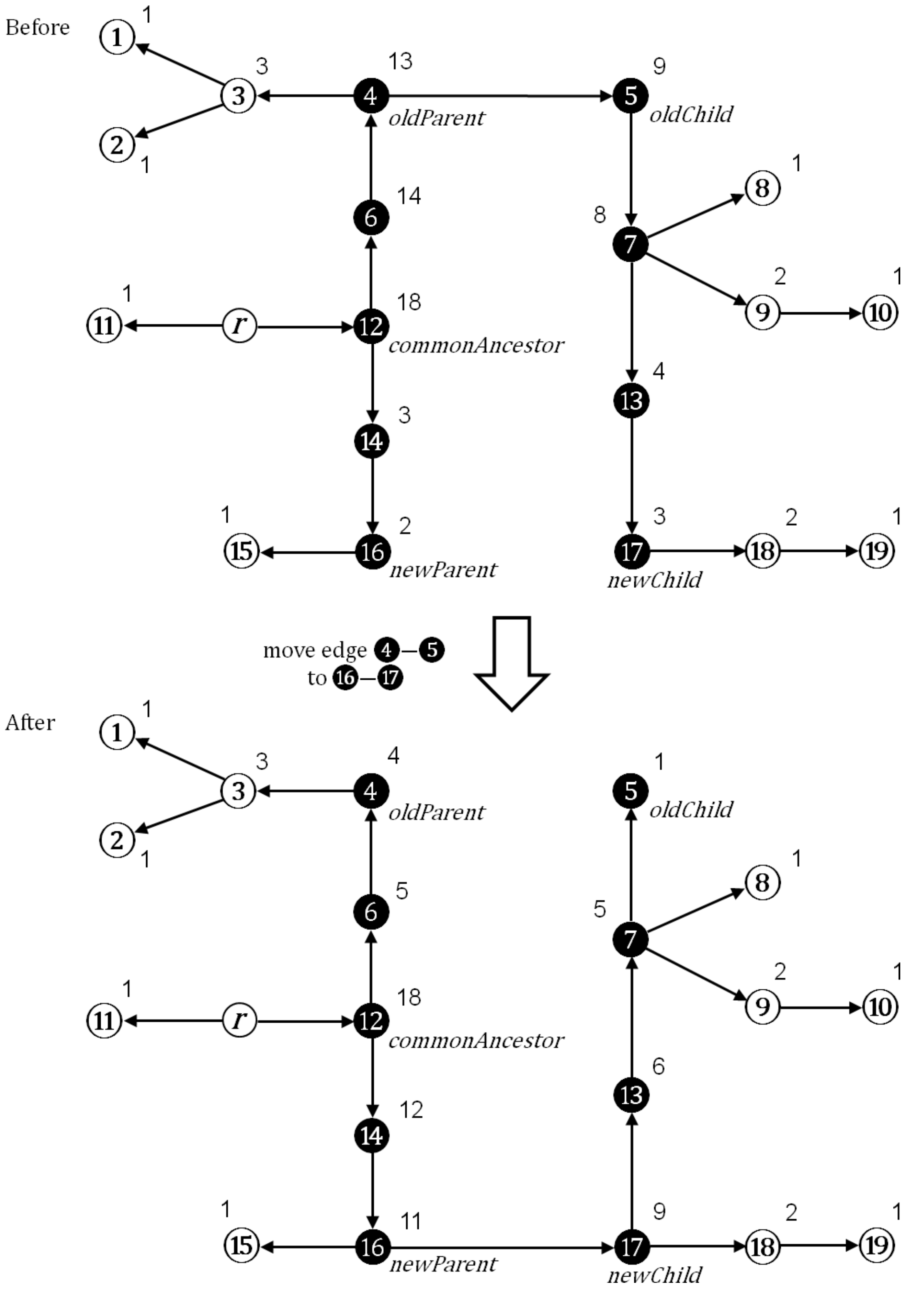

**Figure 9.4.** A tree before and after an edge is moved according to Algorithm IX. The weights are shown next to the nodes; $r$ is the root. The white nodes are unaffected.

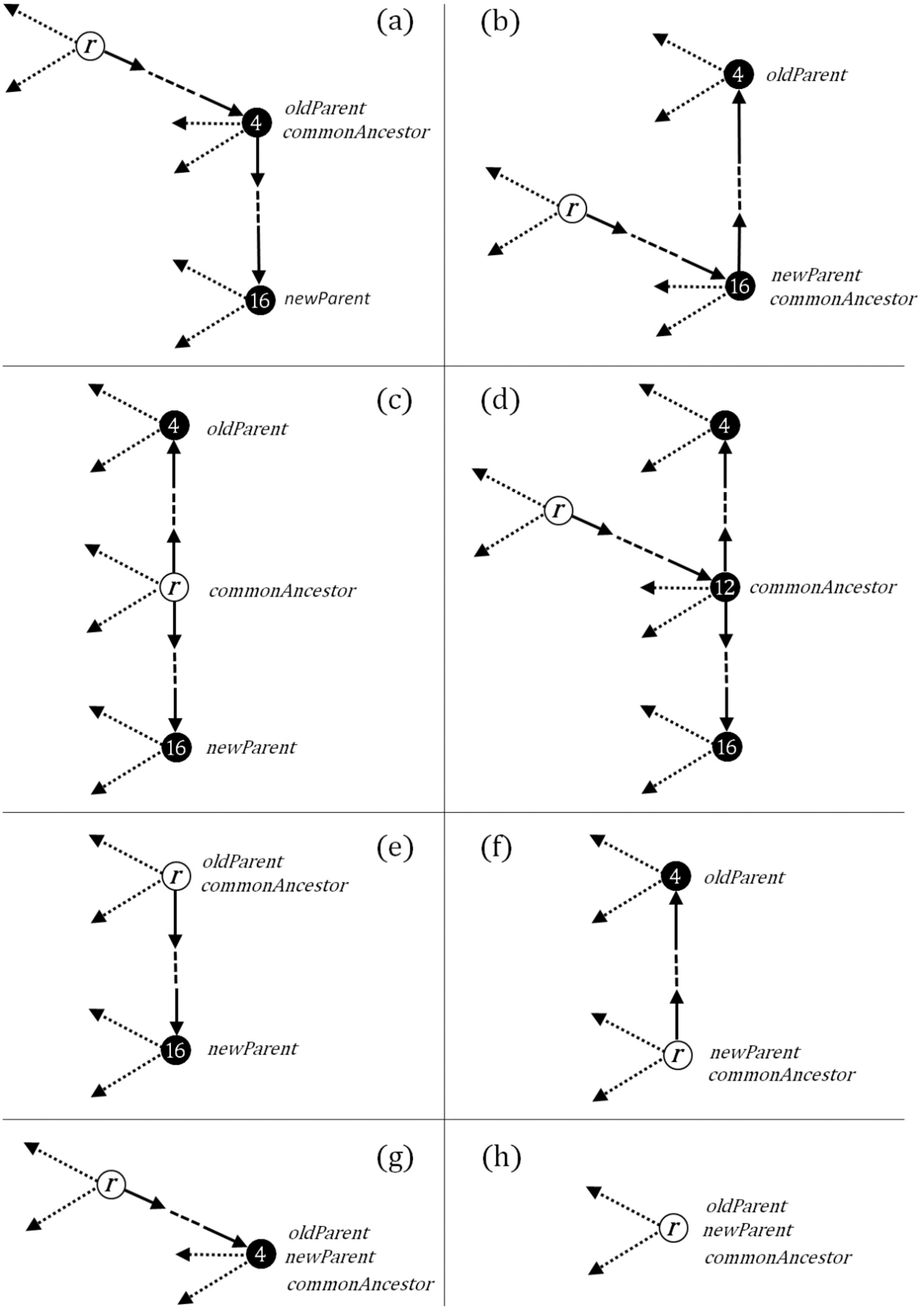

**Figure 9.5 (previous page).** The eight possibilities for the relative positions of $root$ (shown as $r$), $oldParent$, and $newParent$ in $oldComp$. Where a node has dotted arrows going out of it, this means there may be any number of edges going out of it, and on the ends of these edges there may be any tree structures. The dashed lines indicate directed paths that may be of any length. (All nodes on these paths should also be regarded as having dotted arrows going out of them.)

## 9.5 Supplementary notes: Prüfer sequences

Prüfer sequences (Prüfer 1918, Wu & Chao 2004), also known as Prüfer codes, are an alternative way to store trees. They are sequences of length $p - 2$ whose elements are the labels of the nodes (or, equivalently, integers in $\{1, \dots, p\}$). There is a one-to-one correspondence between all the possible trees on $p$ nodes and all the possible Prüfer sequences, and there are algorithms for working out the Prüfer sequence from the tree and vice versa. The one-to-one correspondence trivially implies Cayley's formula for the number of trees on $p$ labelled nodes, $p^{p-2}$ (Cayley 1889). Changing one letter in the Prüfer sequence does not correspond to anything so simple as moving one edge in the graph, so it does not seem sensible to use Prüfer sequences for the present purpose.

# 10 Algorithms for exploring the posterior distribution

## 10.1 Adaptations of two algorithms

### Preamble

In Bayesian structure-learning for GGMs restricted to forests or trees, if there are 15 or more nodes then there are still too many graphs for it to be possible to analyze all of them. Instead the posterior distribution has to be approximated in some way (as mentioned in “Exploring the posterior distribution” in section 3.1). This section describes two ways of doing this.

### Reversible-jump MCMC for structure-learning

One way to approximate the posterior distribution is reversible-jump MCMC, based on the method for decomposable graphs described in Giudici & Green (1999). This and the next two subsections describe how this method can be adapted for forests or trees. Much of this is closely based on section 3.2 of Giudici & Green (1999), and most of the notation is the same. If $M$ is a matrix, then $M_A$ means the submatrix of $M$ that consists of the rows and columns indexed by the elements of $A$.

The standard Metropolis–Hastings algorithm creates a Markov chain whose distribution converges to a given invariant distribution; values from this Markov chain are used as an approximate sample from the distribution. Reversible-jump MCMC is similar, but the dimension of the state-space can change from one step to the next, so a more complicated formula has to be used for the acceptance probability. Reversible-jump MCMC is mostly used for approximating posterior distributions that include models of several different dimensions.

The formula for the acceptance probability in reversible-jump MCMC is equation (7) in Green (1995). It is somewhat complicated, so I will give the formula for the special case of proposing a move to a higher-dimensional variable. Let the variable be $y$ and the desired invariant distribution be $\pi(y)$, and suppose that the proposed move is from $y$ to $y'$, which has higher dimension. Sample $u$ from a distribution with density $q$ and let $y' = y'(y,u)$ be an invertible deterministic function. The move is accepted with probability

$$\alpha(y,y') = \min\left\{1, \underbrace{\frac{\pi(y')}{\pi(y)}}_{\substack{T\,=\,\text{ratio of}\\ \text{target densities}}} \times \underbrace{\frac{r(y')}{r(y)q(u)}}_{\substack{P\,=\,\text{proposal}\\ \text{ratio}}} \times \underbrace{\left|\frac{\partial y'}{\partial(y,u)}\right|}_{\text{Jacobian}}\right\}.$$

Here $r(y)$ is the probability of choosing this type of move, starting from $y$. This formula ensures that the Markov chain satisfies detailed balance and its distribution converges to $\pi$.

Giudici & Green (1999) explain how to use reversible-jump MCMC to create an approximate sample from the posterior distribution in Bayesian structure-learning of GGMs, for decomposable graphs, using the HIW prior distribution for $\Sigma$. In the next two subsections I describe adaptations of this algorithm for forests and trees, respectively, and show how the formulas for the various acceptance probabilities can be derived from general formulas. These adapted algorithms produce an approximate sample from the posterior distribution of $y = (G, \Gamma)$, where $G$ is the graph and $\Gamma$ is the incomplete covariance matrix. $\Gamma$ only contains the elements that correspond to edges in $G$; the other elements are blank. (See Giudici & Green 1999 for why it is convenient to use $\Gamma$ rather than $\Sigma$ or $\Sigma^{-1}$.)

The dimension of $\Gamma$ is the same as the number of edges in $G$. In the case of trees, the dimension of $\Gamma$ always stays the same but the positions of its elements change when $G$ changes, so reversible-jump MCMC is still appropriate. The main object of interest is the posterior distribution of $G$, which is simply the marginal distribution of $G$.

The MCMC for forests repeatedly performs the following two types of move:

(a) add or delete an edge from $G$ (this also requires changes to $\Gamma$),

(b) change all the elements of $\Gamma$.

The MCMC for trees is the same except that, instead of adding or deleting an edge from $G$, it moves an edge from one position to another.

Giudici & Green (1999) use a slightly more elaborate algorithm, with a hierarchical prior for the HIW parameters $\delta$ and $D$. Their variable is $y = (G, \Gamma, \delta, D)$, and they have two further move-types, for updating $\delta$ and $D$.

## MCMC on forests

MCMC on forests, which I will call McmcF, is a simplified version of MCMC on decomposable graphs. For move-type (a), the proposal is to update $y = (G, \Gamma)$ to $y' = (G', \Gamma')$, where $G = (V, E)$ and $G' = (V, E')$. The edge to add or remove, $(v_i, v_j)$, is chosen uniformly at random from the edges that can be added or removed.

*Adding an edge*

First consider the case where the edge is to be added, so that $E' = E \cup (v_i, v_j)$. The formula for the acceptance probability can be derived from the formula in the previous subsection. Firstly consider the ratio of the target densities, which I will call $T_d^+$. (Here $d$ means "decomposable graphs" and $+$ means "adding an edge"; I will use similar notations for other quantities and other types of move.) This is

$$T_d^+ = \frac{\pi(y')}{\pi(y)} = \frac{h(\Sigma_S)h\left(\Sigma'_{S\cup\{i,j\}}\right)}{h\left(\Sigma_{S\cup\{i\}}\right)h\left(\Sigma_{S\cup\{j\}}\right)},$$

where $S$ is a separator on the path between cliques that contain $i$ and $j$, and $h(\Sigma_A) = IW(\Sigma_A; \delta, D_A) \times N(x_A, \Sigma_A)$. Here $IW$ is the inverse-Wishart density and $N$ is the multivariate Gaussian likelihood.

In a forest, $(v_i, v_j)$ can only be added if $v_i$ and $v_j$ are in different components. It follows that $S = \emptyset$ and the ratio simplifies to

$$T_f^+ = \frac{h(\Sigma'_{ij})}{h(\Sigma_i)h(\Sigma_j)}.$$

Calculating $h(\Sigma_i)$ and $h(\Sigma_j)$ involves the one-dimensional inverse-Wishart distribution $IW(\sigma; \delta, D)$, which is the same as the inverse-gamma distribution with parameters $\delta/2$ and $D/2$.

The next part of the formula is $P$, the proposal ratio:

$$P_f^+ = \frac{r^-(y')}{r^+(y)q(u)}.$$

Here $r^+(y)$ is the probability of choosing this particular move, which is

$$r^+(y) = \frac{1}{|existing_G| + |addable_G|},$$

where $existing_G$ is the number of existing edges in $G$ and $addable_G$ is the number of addable edges in $G$; $r^-(y')$ is the probability of choosing the reverse move, from $G'$ to $G$. To calculate $r^-(y')$, let $I$ be the component in $G$ that contains $v_i$ and let $J$ be the component that contains $v_j$. Compared to $G$, $G'$ has one more existing edge and $|I||J|$ fewer addable edges, so

$$r^-(y') = \frac{1}{|existing_G| + 1 + |addable_G| - |I||J|}.$$

As with decomposable graphs, $\Gamma$ is updated to $\Gamma'$ by adding a new element in positions $(i, j)$ and $(j, i)$. The new element is $\gamma'_{ij} = u$, and this is drawn from a zero-mean Gaussian distribution with variance $\sigma_G^2$, so

$$q(u) = \frac{1}{\sigma_G\sqrt{2\pi}} \exp\left(-\frac{u^2}{2\sigma_G^2}\right).$$

The variance $\sigma_G^2$ is chosen by the user. The last part of the formula for the acceptance probability is the Jacobian. As with decomposable graphs, this is 1, because the new parameter $u$ is used with no transformation (Giudici & Green 1999, Green 2003). The acceptance probability is therefore just

$$\min\{1, T_f^+ P_f^+\}.$$

If the graph prior distribution is not uniform, then $T_f^+$ needs to be multiplied by $p(G')/p(G)$, where $p(G)$ is the prior probability of $G$. The same is true in the subsequent cases (for $T_f^-$ and $T_t$).

*Removing an edge*

If the edge is to be removed, then $E' = E \setminus (v_i, v_j)$. For decomposable graphs, the ratio of the target distributions is

$$T_d^- = \frac{\pi(y)}{\pi(y')} = \frac{h\left(\Sigma_{S\cup\{i\}}\right)h\left(\Sigma_{S\cup\{j\}}\right)}{h(\Sigma_S)h\left(\Sigma'_{S\cup\{i,j\}}\right)},$$

which is just $1/T_d^+$. Similarly, for forests the ratio is $T_f^- = 1/T_f^+$.

The proposal ratio is

$$P_f^- = \frac{r^+(y')q(u)}{r^-(y)}.$$

(There is a $q$ in the numerator and not in the denominator because the dimension is being decreased; this follows from equation (7) in Green 1995.) Here $r^-(y)$ is the probability of choosing this move. This is the same as $r^+(y)$, which is given above. To calculate $r^+(y')$, let $I$ be the component in $G'$ that contains $v_i$ and let $J$ be the component that contains $v_j$. Compared to $G$, $G'$ has one less existing edge and $|I||J|$ more addable edges, so the probability of the reverse move is

$$r^+(y') = \frac{1}{|existing_G| - 1 + |addable_G| + |I||J|}.$$

$\Gamma$ is updated by removing its $(i,j)$ and $(j,i)$ elements; $u = \gamma_{ij}$ and $q$ is as above. The Jacobian is 1 and the acceptance probability is

$$\min\left\{1, T_f^- P_f^-\right\}.$$

*Updating the incomplete covariance matrix*

Move-type (b), the update of $\Gamma$, is exactly as in Giudici & Green (1999). Each element is perturbed by adding a zero-mean Gaussian random variable with variance $\sigma_{ij}^2$. In symbols,

$$\gamma'_{ij} \sim N\left(\gamma_{ij}, \sigma_{ij}^2\right).$$

Here $\sigma_{ij}$ is a single value, chosen by the user, though it could be a different value for each pair $(i,j)$. This is not a dimension-changing move, so the appropriate acceptance probability can be found using the formula for the standard Metropolis–Hastings algorithm. It consists of two factors. The first is the ratio of the target distributions, which is

$$T_f^\Gamma = \frac{HIW(\Sigma' \mid \delta, D, G)N(x \mid \Sigma', G)}{HIW(\Sigma \mid \delta, D, G)N(x \mid \Sigma, G)}.$$

$HIW$ is the HIW density and $N$ is the multivariate Gaussian likelihood. The second factor is the ratio of the proposal distributions. This is 1 since the proposal distribution is symmetric. The acceptance probability is therefore $\min\{1, T_f^\Gamma\}$.

## MCMC on trees

I will call this McmcT.

*Updating the graph*

Updating a tree by moving an edge requires removing one element from $\Gamma$ and inserting a different element. The dimension of the parameter space stays the same, but the meaning of the parameters changes. The formula for the acceptance probability can be derived from the formula for general reversible-jump MCMC.

First consider move-type (a). Suppose the proposed update to the graph is to move the edge $(v_i, v_j)$ to $(v_k, v_l)$, so that $E' = E \cup (v_k, v_l) \setminus (v_i, v_j)$. The ratio of the target densities is

$$T_t = \frac{h(\Sigma_{kl})}{h(\Sigma_k)h(\Sigma_l)} \times \frac{h(\Sigma_i)h(\Sigma_j)}{h(\Sigma_{ij})},$$

where $h(\Sigma_A)$ is as for forests.

The proposal ratio is

$$P_t = \frac{r(y')}{r(y)} \times \frac{q(\gamma_{ij})}{q(\gamma'_{kl})}.$$

The factors in this will be explained in turn. Firstly, $r(y)$ is the probability of the current move, which is $1/m(G)$, where $m(G)$ is the number of possible moves from $G$; $r(y')$ is the probability of the reverse move, which is $1/m(G')$. Assume the edge-move is chosen uniformly at random from among all the possible edge-moves, as described in section 9.4, and $v_1$ is the root. If the current graph is $G$, then the number of possible moves is

$$m(G) = \sum_{z=2}^{p} \left[ W(v_z)\big(p - W(v_z)\big) \right],$$

where $W(v) = |de(v)| + 1$. The values of $m(G)$ and $m(G')$ can be calculated when they are needed. Alternatively, $m(G')$ can mostly be calculated from $m(G)$—most of the values in the sum for $m(G')$ are the same as the values in the sum for $m(G)$, since the only nodes whose weights change are the ones on two particular paths (see section 9.4). As for forests, the incomplete covariance matrix $\Gamma$ is updated by removing $\gamma_{ij}$ and inserting $\gamma'_{kl}$, which is drawn from $N(0, \sigma_G^2)$, whose density is $q$.

The Jacobian is 1. Putting all these together, the acceptance probability is $\min\{1, T_t P_t\}$.

The simplest alternative to choosing edge-moves uniformly at random is to choose an edge uniformly at random, then remove it, then reinsert it uniformly at random. Suppose the components that result from removing the edge are $I$ and $J$. The probability of choosing that edge to remove is $1/(p-1)$, and the probability of putting it back in any particular position is $1/(|I||J| - 1)$, so the probability of any particular move is

$$\frac{1}{p-1} \times \frac{1}{|I||J| - 1}.$$

The probability of the reverse move is the same, so these two elements cancel out, and the proposal ratio is just $q(\gamma_{ij})/q(\gamma'_{kl})$.

*Updating the incomplete covariance matrix*

The update of $\Gamma$ is the same as in the case of forests.

## Stochastic shotgun search on forests and trees

An alternative to MCMC is the shotgun stochastic search algorithm that appears in section 6 of Jones et al (2005). This algorithm moves around in the space of possible graphs, calculating the unnormalized posterior probability of the graphs that it visits and some of their neighbours, and usually moving towards graphs with higher probability. It does not involve a Markov chain and it does not give an approximation to the posterior distribution of $\Sigma$. Below is a version of this algorithm that has been adapted for forests or trees. I call the version for forests SSSF and the version for trees SSST.

1. Start with a forest/tree $G$, and calculate and store its unnormalized posterior probability.
2. Choose $\omega$ distinct moves from $G$. (For forests a move consists of adding or removing an edge, and for trees it consists of moving an edge. Use the algorithms in section 9.3 for forests and section 9.4 for trees.)
3. Calculate and store the unnormalized posterior probabilities of the $\omega$ neighbouring forests/trees that result from doing these moves (except in the case of graphs for which this has previously been done).
4. Select one of the $\omega$ neighbouring graphs by choosing each with probability proportional to its unnormalized posterior probability, and set $G$ to be this graph.
5. Go back to step 2 and repeat many times. (Either stop after a fixed amount of time or after a fixed number of iterations.)

The unnormalized posterior distribution is taken to be the values that were calculated for the graphs whose probabilities were calculated, and zero for all other graphs. The algorithm is intended as a simple alternative to MCMC with the possible advantage that it always moves to a new graph at every iteration, so it cannot get stuck at a single graph. It simply explores the space of possible graphs, finding their unnormalized posterior probabilities, and tends to move towards graphs that have higher probabilities. It sometimes moves to graphs of lower probability, so it is not just deterministic greedy hill-climbing. Any particular route through all the possible graphs has positive probability, so if run for enough time it will eventually visit all the graphs. In this trivial sense it asymptotically gives the true posterior distribution.

As well as the restriction to forests or trees, the above algorithm is different from the original one in Jones et al (2005) in three other ways. Firstly, in the original algorithm, at step 3 only the top $x_2$ neighbouring graphs are retained. Secondly, at step 4 the neighbouring graph $G_i$ is chosen with probability proportional to $p_i^{\alpha}$, where $p_i$ is its unnormalized posterior probability and $\alpha$ is a positive annealing parameter. (As $\alpha \to \infty$ the original algorithm becomes deterministic greedy hill-climbing.) Thirdly, at step 5 only a list of the top $x_3$ graphs is stored. Of these three differences, the third is the most likely to be useful, since storing all the graphs takes a lot of memory.

The experiments in sections 7 and 8 of Jones et al (2005) use $\alpha = 1$ and $x_2 = \omega$, which make their algorithm similar to the one given above. They also set $\omega$ to be the number of neighbouring graphs, so the algorithm calculates the unnormalized posterior probabilities of all the neighbouring graphs, not just some of them. (The set of all the neighbouring graphs consists of all the graphs that can be made by making a single move from the present graph.) If all the neighbouring graphs are analyzed, then in step 2 there is no need to choose moves at random, uniformly or otherwise. In the case of trees, this would mean that the node-weights are not needed.

With trees and large $p$, the number of neighbouring graphs is huge, as shown in Table 10.1, so if all of them are analyzed it would take a long time to do even one iteration of SSST. For this reason I use the version where only some of the neighbouring graphs are checked at each iteration.

Jones et al (2005) say their algorithm is designed for distributed implementation (which means using multiple computers at once), and that "distributed computation is essential to the development of search and constructive methods beyond moderate dimensions." Scott & Carvalho (2008) imply that using distributed computing is the main purpose of Jones et al (2005)'s algorithm. Certainly, step 3 can be parallelized in an obvious way. But my programs to implement my versions of their algorithm are serial, not parallel, and they give reasonable results in a short amount of time (see the experiments in chapter 11).

| **Graph** | $p = 100$ | | $p = 1000$ | |
|---|---|---|---|---|
| | star | chain | star | chain |
| **Number of neighbours / possible edge-moves** | 9 801 | 998 001 | 166 650 | 166 666 500 |

**Table 10.1.** The number of neighbouring graphs (equivalently, the number of possible edge-moves) within the space of trees, for four selected graphs. "Star" and "chain" are defined in section 11.1 and the values were calculated using Propositions 11.4 and 11.5.

## How to store decomposable graphs

In section 11.7, SSSF and SSST are compared with the stochastic shotgun search algorithm on decomposable graphs. My programs for these experiments store and manipulate decomposable graphs in basically the same way as Giudici & Green (1999) and Jones et al (2005)—see also Jones et al (2004), which is a longer version. Full details are in section 3.1 and the appendix of Giudici & Green (1999) and section 2.1 of Green & Thomas (2013). Here I will just give aspects that are specific to my programs.

My programs store decomposable graphs as junction trees and manipulate them by adding or removing any edge that can be added or removed, as in Giudici & Green (1999). If the proposed edge is not in the graph and the closest two cliques that contain the two nodes are not neighbours in the junction tree, then the junction tree is manipulated to make the two cliques be neighbours, as described in the second-last paragraph of Giudici & Green (1999).

Using junction trees, rather than junction forests, has the advantage that adding an edge between two separate components does not need to be treated separately, since it is a special case of the move shown by the downwards arrow in Figure 3(d) of Green & Thomas (2013). It also means that separators are sometimes empty.

## 10.2 Analyzing posterior graph distributions and assessing algorithms

### How frequentist algorithms are evaluated

Frequentist algorithms for graphical model structure-learning produce a single graph. (See section 3.2.) If the true graph is known, the natural way to measure how well one of these algorithms does is to compare the graph produced by the algorithm with the true graph. There are two scenarios in which you would know the true graph. One is that you used simulated data that was generated from a distribution that corresponds to this graph. The other is that the data corresponds to objects that have been analyzed using non-statistical methods, and a supposedly true graph-structure has been deduced from this analysis. The latter scenario is sometimes the case with networks of gene or protein interaction—see for example Albieri (2010).

Probably the simplest ways to measure the success of a frequentist algorithm are the numbers of true-positives, false-positives, false-negatives, and true-negatives. Table 10.2 shows the meanings of these phrases.

| | | **True graph** | |
|---|---|---|---|
| | | Edge | Non-edge |
| **Graph produced by the algorithm** | Edge | true-positive | false-positive |
| | Non-edge | false-negative | true-negative |

**Table 10.2.** The meanings of "true-positive" and related phrases, for a single graph produced by a frequentist algorithm. For example, a true-positive is an edge that is in both the true graph and the graph produced by the algorithm.

True-positives and the other three quantities are not specific to graphs or graphical model structure-learning. They can be used with any type of binary classification, for example frequentist statistical hypothesis tests or medical tests to diagnose whether a person has a disease—a test is reliable if it seldom gives false-positives or false-negatives.

Also used are several ratios (Albieri 2010, page 50). In the following, $TP$ stands for the number of true-positives, and the other abbreviations are similar:

$$\text{precision} = \frac{TP}{TP+FP}$$

$$\text{true-positive rate} = \text{recall} = \text{sensitivity} = \frac{TP}{TP+FN}$$

$$\text{true-negative rate} = \text{specificity} = \frac{TN}{TN+FP}$$

$$\text{false-positive rate} = \frac{FP}{TN+FP}$$

$$\text{false-negative rate} = \frac{FN}{TP+FN}$$

$$\text{accuracy} = \frac{TP+TN}{TP+FN+FP+TN}$$

$$\text{error rate} = \frac{FP+FN}{TP+FN+FP+TN}$$

For example, the true-positive rate is the proportion of edges in the true graph that were correctly identified by the algorithm. Probably the most-used rates are the first three. For these, higher values are better.

Many frequentist algorithms have a tuning parameter. Varying this and repeating the algorithm gives different values of the rates. An algorithm can be assessed by plotting the precision (on the vertical axis) against the recall, for different values of the tuning parameter. Alternatively it can be assessed by plotting the recall against the false-positive rate—this is called a receiver operating characteristic (ROC) curve. For examples see Albieri (2010) or Guo et al (2011).

Some algorithms for estimating multivariate Gaussian distributions produce an estimate of the covariance matrix. This can be assessed using measures such as entropy loss and Frobenius loss (Guo et al 2011). Some research on estimating these distributions talks only about the covariance matrix and does not mention graphs at all (though Dempster 1972, for example, does have elements of the precision matrix set to exactly zero).

All these methods can be used either to compare different algorithms or to compare different parameters in the same algorithm. For ways of evaluating algorithms for directed acyclic graphical models, see Gasse et al (2012).

Frequentist and Bayesian algorithms can also be assessed by using the estimated graphs or covariance matrices they produce to make predictions. These predictions can be compared with reality or with other data that were not used by the algorithm. But making predictions is a different goal from learning the structure.

## How Bayesian methods are evaluated

Assessing a Bayesian method is more complicated than assessing a frequentist method, because the former produces a graph distribution rather than a single graph. In this section I will refer to the posterior probability that an edge is present in the graph as the probability of that edge. This is sometimes called an edge-inclusion probability.

One way to evaluate a Bayesian method is to use the MAP graph, and see for example how many true-positives it has. But a single graph is rather simple given the large amount of information in the posterior distribution. There is another drawback that is easiest to explain with an example. It might be the case that there are ten graphs that have high probability, and the 2nd to 9th most likely graphs have many edges in common with each other but few edges in common with the most likely (MAP) graph. In this case, if you are going to use a single graph, then it would probably be better to use the 2nd most likely graph, or the graph consisting of all edges whose probabilities are above a certain threshold, rather than the MAP graph.

Any of the algorithms described in section 10.1 produces an approximation to the graph posterior distribution, in the form of a set of graphs and estimates of their posterior probabilities. The MTT-based methods described in chapter 8 do not produce the entire posterior distribution itself, or an approximation to it, but they can produce several exact quantities that can be used to compare algorithms.

As with frequentist algorithms, there are two scenarios in which the true graph is known. If the true graph is known, the simplest quantities with which to evaluate algorithms are the expected number of true-positives and the expected values of the other quantities in Table 10.2.

Below are listed the main objects or types of information that have been used in previous research to summarize estimated graph posterior distributions for undirected graphical models, or to evaluate the algorithms that produced these distributions.

- **The probabilities of all of the edges, in the form of a triangular matrix.** The elements of the matrix are $p_{ij} = \mathbb{P}((i,j) \in E \mid x)$. See Wang & Li (2012)'s example with six nodes or Dobra et al (2011)'s example with ten nodes. This seems to be the most commonly used information.
- **The probabilities of all the edges, in the form of a diagonally symmetric square grid where the shade of grey in each little square indicates that edge's probability.** See Wong et al (2003), Scott & Carvalho (2008), or Armstrong et al (2009).
- **The probabilities of certain specific edges.** See Jones et al (2005) or Carvalho & Scott (2009).
- **A graph consisting of all the edges whose probabilities are above a certain threshold.** See Wang & Li (2012)'s example with 100 nodes, where the threshold is ½ and they call this object the "posterior mean graph", or Armstrong et al (2009)'s example with 11 nodes, where the threshold is 70%.
- **The top-ranking (most likely) graph or, less commonly, graphs.** See Giudici & Green (1999) or Jones et al (2005).
- **The probabilities of the top-ranking graph or, less commonly, graphs.** These may be normalized or unnormalized. See Giudici & Green (1999), Jones et al (2005), or Scott & Carvalho (2008). Scott & Carvalho (2008) judge their search algorithms by the posterior probabilities of the models they find—the higher, the better. For each algorithm they show the top 1000 posterior probabilities on a histogram.
- ***ETPR* (expected true-positive rate) and related quantities.** Moghaddam et al (2009) give a plot of $TPR$ against $FPR$ in which each graph appears as a single point, and the expected values of these rates (under Bayesian model-averaging) is plotted as a single point in a different colour.

These numbers or graphs are usually then commented on and discussed, and conclusions are reached about which algorithm, prior, or parameter is best.

Most of the objects listed above only convey separate information about each edge; they do not show which combinations of edges are likely to be present. Consequently they fail to reveal certain notable features of the graph distribution. As an extreme and unlikely example, if the top few graphs were stars centred at different nodes (see Definition 11.1), then the matrix of edge-probabilities would only show that all the edges in those stars were likely. It would not reveal that the graph was almost certainly a star. Only the MAP graph and the other top-ranking graphs convey any information about the graph as a whole, not just separate edges.

Section 2.4 mentioned the importance of hubs or stars. Albieri (2010) found that when the true graph contains a star, frequentist algorithms mistakenly find that these nodes form a clique. In evaluating Bayesian methods it would be desirable to be able to notice any consistent structural "bias" such as this. The simplest way to judge whether stars are correctly identified would be to look at the posterior expected degrees of the hub and the nodes it is connected to. (If the entire true graph is a star, then the expected degree of the hub is exactly the same as $ETPR$.) In chapter 11, I evaluate various Bayesian algorithms and priors using simulated datasets, but my main algorithms only consider forests or trees, so there is no possibility of misidentifying a star as a clique.

Friedman et al (2000) use indicator functions for features of the graph. They work with directed graphical models. The first type of features they consider is "Markov relations", about whether one node is in the Markov blanket of another (which holds if the two nodes are connected by an edge or share a child). The other is "order relations", about whether one node is an ancestor of another.

## Single numbers for evaluating Bayesian methods

In abbreviations hereafter, "E" means "expected". It is useful to have a small number of numerical quantities to evaluate how well a Bayesian method does, because these are ordered and easier to interpret than large matrices or graphs. Possible quantities are $ETPR$, the three quantities related to it, and the expected degrees of the nodes. The algorithms in section 10.1 are random, so the estimates of $ETPR$ and the other quantities will probably vary between different runs.

If the Bayesian analysis is restricted to trees, then there are restrictions on $ETP$ and the three related quantities. In particular, $ETP + EFP = p - 1$. If the true graph is also a tree, $ETP + EFN = p - 1$, so only one of the four quantities is worth looking at. If the analysis is restricted to forests, then it may be useful to look at two values, for example $ETP$ and $ETN$, but if all the high-ranking graphs are trees then $ETP + ETN$ will be close to $p - 1$.

In the case of trees, the MTT-based methods from section 8.2 and 8.3 can be used to find $ETPR$. I will use them for these purposes in section 11.5.

## Formulas for evaluating Bayesian algorithms

As examples, I will give formulas for four quantities to do with the posterior graph distribution: (a) the probability of any particular edge, (b) the degree of any particular node, (c) $ETP$, and (d) $ETPR$.

Suppose the true graph is $G = (V, E)$ and the graphs produced by the algorithm are $\{G_i = (V, E_i)\}$. In the case of the MTT-based method, this set contains all the possible trees. Let

$$\mathbb{I}_{G_i}^{(u,v)} = \begin{cases} 1 \text{ if } (u,v) \in E_i \\ 0 \text{ otherwise.} \end{cases}$$

(a) The posterior probability of edge $(u, v)$ is

$$\sum_i p(\, G_i \mid x\,)\mathbb{I}_{G_i}^{(u,v)} .$$

This formula uses the normalized posterior probabilities of the graphs, so you have to add all the unnormalized probabilities to find the normalizing constant (except in the case of the MTT-based method, which calculates the normalizing constant without using all the separate unnormalized probabilities).

(b) The expected degree of node $v$ is

$$\sum_i \left( p(\, G_i \mid x\,) \sum_{u \neq v} \mathbb{I}_{G_i}^{(u,v)} \right) = \sum_i \sum_{u \neq v} p(\, G_i \mid x\,)\mathbb{I}_{G_i}^{(u,v)}.$$

(c) The number of true-positives in $G_i$ is

$$TP = \sum_{u,v \in V} \mathbb{I}_G^{(u,v)} \mathbb{I}_{G_i}^{(u,v)} ,$$

so

$$ETP = \sum_i \sum_{u,v \in V} p(\, G_i \mid x\,)\mathbb{I}_G^{(u,v)} \mathbb{I}_{G_i}^{(u,v)}$$

$$= \sum_{(u,v) \in E} \sum_i p(\, G_i \mid x\,)\mathbb{I}_{G_i}^{(u,v)}.$$

The second expression shows that $ETP$ is the sum of the posterior probabilities of all the edges in the true graph.

(d) For $G_i, TPR = TP/|E_G|$. So $ETPR$ is

$$ETPR = \sum_i \sum_{u,v \in V} p(\, G_i \mid x\,) \frac{\mathbb{I}_G^{(u,v)} \mathbb{I}_{G_i}^{(u,v)}}{|E_G|} = \frac{ETP}{|E_G|},$$

which lies between 0 and 1.

For algorithms that produce an approximation to the entire posterior distribution, an alternative would be to calculate these values by only using the top $N$ most likely graphs, for some $N$.

## Visual representations of graph distributions

What is the best way to visually represent or summarize a graph posterior distribution? A single graph is too simple. A triangular or symmetric matrix of edge-probabilities contains more information, but for eight or more nodes it is probably impossible to notice any overall patterns. A square grid, with colours or shades representing the values in this matrix, works well if the graph has a strong structure, but if the nodes are in no particular order and the high-probability graphs have complicated structure then it may be hard to take in.

Another possibility is a graph in which the thickness of each edge is proportional to its probability, and only edges whose probabilities are greater than a certain threshold are shown. This would be easy to take in, because there is no need to peer at numbers or count which column or row an entry is in. But it still only conveys separate information about each edge.

The best way of showing an entire graph distribution is an animation that consists of graphs generated from it. I realized this when Peter Green said it during a talk in November 2012. Java programs that can show animations of the MCMC in Green & Thomas (2013) are normally available from Alun Thomas's JPSGCS website, at *http://balance.med.utah.edu/wiki/index.php/JPSGCS*, though as of February 2013 this is not working.

## Supplementary notes: further details on evaluation of Bayesian methods

It may be useful to give more detail about some of the papers that evaluate Bayesian methods. Dobra et al (2011)'s example with ten nodes is about matrix-variate GGMs. As well as the estimated edge probabilities they also give the standard errors of these estimated probabilities. One of their methods did very well, giving probability 1 to all the edges in the true graph and less than 0.1 to all the edges that were not. Matrix-variate distributions have two graphs; one of their true graphs had $p = 5$ and the other was a loop with $p = 10$. Wang & Li (2012) found in one of their examples that all the edges in the true graph got probability 1 and all the other edges got probability below 0.08. The true graph was a loop with $p = 100$.

Carvalho & Scott (2009) and Wang & Li (2012) evaluate their posterior distributions by using them for prediction in mutual funds (schemes that pool money from many investors and invest it in stocks or other financial assets—see U.S. Securities and Exchange Commission 2010). Moghaddam et al (2009) evaluate posterior distributions by using them for prediction, but I cannot understand whether they use all the graphs or just one of them.

Giudici & Green (1999) also give the expected number of edges under the posterior distribution. Some algorithms produce estimates of the posterior distribution for the covariance matrix. For how these can be assessed see Giudici & Green (1999) or Wong et al (2003).

Armstrong et al (2009) compared their MCMC for GGM structure-learning to one in Brooks et al (2003). To do this they gave a Manhattan plot that showed the number of edges in the graph at each iteration and the cumulative number of graphs visited at each iteration. They also used effective sample sizes. All the methods in this section (10.2) can of course just as well be used for directed acyclic graphical models. See for example Altomare et al (2011).

# 11 Experiments

## 11.1 Facts about star and chain graphs

The subsequent sections of this chapter describe experiments on simulated data. Most of the datasets were generated from distributions that correspond to star and chain graphs. The reasons for using these shapes of graph were that they are extremal in certain senses, described by the three propositions in this section, to do with exploring the space of trees by making local moves. To define these two types of graph, which are also trees, let $V = \{v_1, \dots, v_p\}$.

***Definition 11.1.*** $(V, E)$ is a star if $E = \{\{v_1, v_i\}, \dots, \{v_{i-1}, v_i\}, \{v_{i+1}, v_i\}, \dots, \{v_p, v_i\}\}$ for some $i$. (See also section 2.3.)

***Definition 11.2.*** $(V, E)$ is a chain if the nodes can be relabelled in such a way that $E = \{\{v_1, v_2\}, \{v_2, v_3\}, \dots, \{v_{p-1}, v_p\}\}$.

The weight of node $v$ is $W(v) = 1 + |de(v)|$. The number of edge-moves that start with removing $\{v, pa(v)\}$ is $g\big(W(v)\big) = W(v)\big(p - W(v)\big)$, and the total number of edge-moves is $\sum_{v \neq root} g(W(v))$. In considering the number of possible edge-moves from stars and chains, the root can be chosen arbitrarily from among all the nodes since this does not affect the number of edge-moves.

***Proposition 11.3.*** Stars are the only trees where all the nodes are chosen with equal probability in line 1 of Algorithm IX (section 9.4).

***Proof.*** Suppose there is a tree that contains a path of length 4 and that this tree's edges would be chosen with equal probability in line 1 of Algorithm IX. Regard the tree as a rooted tree with root $r$ at one end of this path. Call the subsequent nodes on the path $v_1$, $v_2$, and $v_3$—see Figure 11.1, in which other nodes are not shown. Now $de(v_1) \supset de(v_2) \supset de(v_3)$, so $W(v_1) > W(v_2) > W(v_3)$. These three nodes being chosen with equal probability means that $g(W(v_1)) = g(W(v_2)) = g(W(v_3))$. But $g$ is a quadratic function, so it is impossible for three distinct values of $x$ to have the same value of $g(x)$. Therefore no such tree can exist. The only trees that contain no paths of length 4 are stars, which completes the proof. ☐

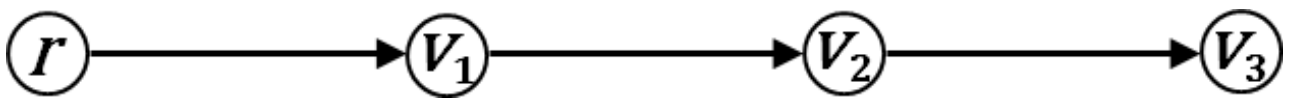

**Figure 11.1.** A path of length 4.

***Proposition 11.4.*** A star with $p$ nodes has $(p-1)^2$ possible edge-moves, and this is the fewest of any tree with $p$ nodes.

***Proof.*** The value of $g(w)$ is minimized at $w = 1$ or $p-1$, and its minimum value is $p-1$. Consider a star with $p$ nodes and suppose that its root is the hub (the node at the centre of the star). All the other nodes have weight 1, so the total number of possible edge-moves is $(p-1)^2$, which is the minimum. ☐

***Proposition 11.5.*** A chain with $p$ nodes has $(p^3 - p)/6$ possible edge-moves, and this is the most of any tree with $p$ nodes.

***Proof.*** Regard the chain graph as a horizontal line, take the leftmost node to be the root, and number the nodes from left to right. The weight of the $(i+1)$th node is $p-i$, so the total number of possible edge-moves is

$$\sum_{i=1}^{p-1} g(p-i) = \sum_{i=1}^{p-1} (p-i)i = \frac{p^3 - p}{6}.$$

To show that this is the maximum, consider a tree that contains a node of degree 3 or more. Let this node be the root, label its children $v_1, v_2, v_3, \ldots$, and let $w_1, w_2, w_3, \ldots$ be these children's weights. Without loss of generality assume that $1 \leq w_1 \leq w_2 \leq w_3 \leq \cdots \leq p-1$. Now $w_1 + w_2 + w_3 + \cdots = p-1$, because the left-hand side counts all the nodes in the graph except the root exactly once. It follows that $w_1 \leq (p-1)/3$ and $w_1 + w_2 \leq 2(p-1)/3$. Because $g(x)$ is a quadratic function with peak at $x = p/2$, it must be the case that $g(w_1 + w_2) > g(w_1)$.

The number of edge-moves that start with the removal of $(root, v_1)$ is $g(w_1)$. Create a new tree by deleting the edge $(root, v_2)$ and replacing it with $(v_1, v_2)$. In the new tree $W(v_1) = w_1 + w_2$, so the number of edge-moves that start with the removal of $(root, v_1)$ is $g(w_1 + w_2) > g(w_1)$. For all other nodes $v$, $W(v)$ and hence $g(W(v))$ are the same as in the original tree. So the new tree has more edge-moves than the old one.

It follows that any tree with the largest possible number of edge-moves must have no node of degree 3 or more. The only such trees are chains. ☐

The three propositions still hold if the "non-move" is excluded, so that $g\big(W(v)\big) = W(v)\big(p - W(v)\big) - 1$. The only differences are that in Proposition 11.4 the number of edge-moves is $(p-1)(p-2)$ and in Proposition 11.5 it is $(p^3 - 7p)/6 + 1$.

In trees with other shapes, individual nodes can be extremal. Let $e = (v, pa(v))$. If $v$ or $pa(v)$ has degree 1, then $g(W(v))$ has its lowest possible value. If $e$ splits the tree as nearly as possible in half, so that $W(v) \in \{(p-1)/2, p/2, (p+1)/2\}$, then $g(W(v))$ has its highest possible value.

## 11.2 Experiments with systems for storing trees

### Different systems for storing trees

Section 9.4 describes a way of storing trees so that edge-moves can be chosen uniformly at random. I will call this System A. To assess System A it is desirable to compare it to other systems for storing trees and choosing edge-moves. Here I describe three other systems. In all four, trees are stored as rooted trees, because this makes it easy to check whether moves are legal. All four systems produce a set of $\omega$ distinct edge-moves. The systems will subsequently be compared using SSST, with the non-move excluded.

**System A (weights).** Trees are stored and edge-moves are chosen as described in section 9.4.

**System B (unused weights).** Edge-weights are stored but not used. To choose $\omega$ edge-moves, first create a list $L$ that contains $p-2$ copies of each edge. Choose $\omega$ edges from $L$ uniformly at random and put these in a list called $M$. These are the edges that are to be removed (and obviously they are not necessarily distinct). For each distinct edge $e$ in $M$, let $m_e$ be the number of times $e$ appears in $M$, identify the two components that result when you remove $e$, and choose $m_e$ distinct places to reinsert it.

**System C (no weights).** No edge-weights are stored. Edge-moves are chosen as in System B.

**System D (rejection).** No edge-weights are stored. To choose $\omega$ edge-moves, repeat the following as many times as necessary: choose an edge uniformly at random, remove it and identify the two components that result, choose where to reinsert it uniformly at random, and accept this edge-move if and only if it has not already been chosen.

System D has one drawback. If $\omega$ is large relative to the total number of possible edge-moves, then it is likely that many edge-moves will be rejected and choosing $\omega$ different edge-moves will take a long time.

Systems B and C are designed to avoid this drawback. The number of edge-moves that start with removing $(v, pa(v))$ is $g(W(v)) = W(v)(p - W(v)) - 1 \geq p - 2$, so for each distinct edge $e$ in $L$ there are at least $p-2$ possible places where it can be reinserted. $L$ contains $p-2$ copies of each edge, so however many copies of $e$ are chosen to be put in $M$, there will certainly be enough possible places for it to be reinserted. There is never any need to reject and repeat. Systems B and C will not work if $\omega > |L| = (p-1)(p-2)$, but this does not matter unless you want to find more edge-moves than that.

The only difference between Systems A and B is that System B does not use the weights (it does store and update them). So if using the weights, and choosing edge-moves uniformly at random, gives some advantage, then this should be evident by comparing the results of experiments that use these two systems.

The only difference between Systems B and C is that System B wastes time storing and updating the node-weights. So System C should always do at least as well as System B. If storing and updating the weights takes little time, then there should be little difference between Systems B and C.

## Datasets

To compare the various algorithms and ways of storing trees, a large number of simulated datasets were generated. The values of $p$ that were used were 30 and 100, and the values of $n$ (the number of data) were 50 and 500. For each value of $p$, two covariance matrices were created, one corresponding to a star and the other corres-ponding to a chain. The diagonal elements of the covariances were all 1 and the non-zero partial correlations were all $0.99/\sqrt{p-1}$; these two conditions, together with the graph, specify the covariances completely. (The reason for using this formula is the inequality in section 2.3 about the partial correlations in stars.)

The four covariance matrices and two values of $n$ give eight combinations of covariance matrix and $n$, whose descriptions can be seen on the horizontal axes in Figure 11.2. If just a single dataset were generated for each of these, then these datasets might be atypical and the results might fail to show the effects of the different systems for storing trees (and the different shapes of graph and values of $p$ and $n$). For this reason, 500 datasets were generated for each combination of covariance matrix and $n$, and the algorithm was run on all of these. The datasets were all generated from zero-mean multivariate Gaussian distributions.

Star and chain graphs were used because they are extremal in the senses described in section 11.1. Suppose the true graph is a star. If at a certain point in SSST the current graph is the true graph, then, by Proposition 11.3, System A is equally likely to choose any of the edges to move. Systems B–D always do this. It follows that all four systems will choose edge-moves uniformly at random. If the current graph is not the true graph but something similar to it, as will probably be the case most of the time, then Systems B–D will choose edge-moves almost uniformly at random. In contrast, when the true graph is a chain, Systems B–D will often choose edge-moves with a distribution that is far from uniform—for example, all chains contain nodes that have the lowest and highest possible values of $g(W(v))$.

## Experiments

The four systems were compared by using them with SSST, as described in section 10.1, and running the algorithm under the same computational conditions and for the same amount of CPU time, with the same parameters. For the hyper inverse Wishart prior on $\Sigma$, the scalar hyperparameter $\delta$ was 3 and the matrix hyperparameter was $I_p(\delta+2)$ (see Jones et al 2005, the erratum listed in the references, and Donnet & Marin 2012). The prior distribution on the graph structure was uniform on trees with $p$ vertices. For these experiments $\omega$ was chosen to be $p^2/20$ so that it scaled appropriately with the number of possible edge-moves; this means $\omega = 45$ in the cases where $p = 30$ and $\omega = 500$ where $p = 100$. For each value of $p$, all runs were started at a fixed graph that was different from either of the graphs that the data were generated from.

The issue of how to evaluate Bayesian structure-learning was discussed in section 10.2. Each run of the algorithm was done for 20 seconds, and the following quantities were recorded:

- the number of distinct graphs visited
- $ETPR$
- the true-positive rate in the top graph
- the score of the top graph (its unnormalized log posterior probability)
- the sums of the scores of the top 10 graphs.

## Results

The results are shown in Figure 11.2. Each bar-chart corresponds to one of the quantities in the bullet-list above. Each group of four bars corresponds to one combination of covariance matrix and $n$, and within each group each bar corresponds to one system for storing trees. For all five bar-charts, larger values are better.

Each bar corresponds to 500 runs of the algorithm on the different datasets generated from the same distribution with the same $n$. The heights of the bars are the median values and the "whiskers" show the 25% and 75% quartiles.

To assess the algorithm in the cases where the true graph was a star, it is also of interest to know the posterior expected degree of the node that was supposed to be the hub at the centre of the star. But this is just $ETPR$ multiplied by $p - 1$, so the relative heights of the bars would be the same as in the second bar-chart.

Most of the bar-charts show no difference between the four systems for storing trees, or only tiny differences. The number of graphs visited varies somewhat. System A does better than the other systems on the datasets with $p = 30, n = 500$, and the true graph a star. But with four of the sets of 500 datasets it does worse than Systems B or C. System D does very badly for two of the sets, which is notable as it is probably the most obvious and easy to program.

It might be expected that the weights would make more difference for chains than for hubs, because of the extremal properties shown in section 11.1. The numbers of graphs visited by System A are indeed more different for chains than for stars. But they are lower. However, the interquartile ranges sometimes overlap.

Overall System C does slightly better than System B, as expected, though there is a large overlap between the ranges. The last two bar-charts are of less interest, firstly because they show no differences between the four systems, and secondly because in these bar-charts it is not legitimate to compare quantities that correspond to different sets of datasets (because scores for different sets of datasets have nothing to do with each other).

The differences shown in the bar-charts between the four systems are minor, but the differences between the eight datasets are major. Unsurprisingly, the expected true-positive rates and the true-positive rates in the top graphs are much higher when $n$ is large, and lower when $p$ is large. Of the four combinations of $n$ and $p$, the only one with $n < p$ is $p = 100, n = 50$. This had the worst results in terms of true-positive rates, which was to be expected. It seems that SSST gives better results with stars than with chains. Perhaps stars are easier to approximate by wrong trees than chains.

Summarizing results from 500 datasets in a single bar has the disadvantage that you cannot compare the four systems for any specific dataset. But in many cases the "whiskers" are close to the tops of the bars, showing that there is not too much variation within each set of 500 datasets.

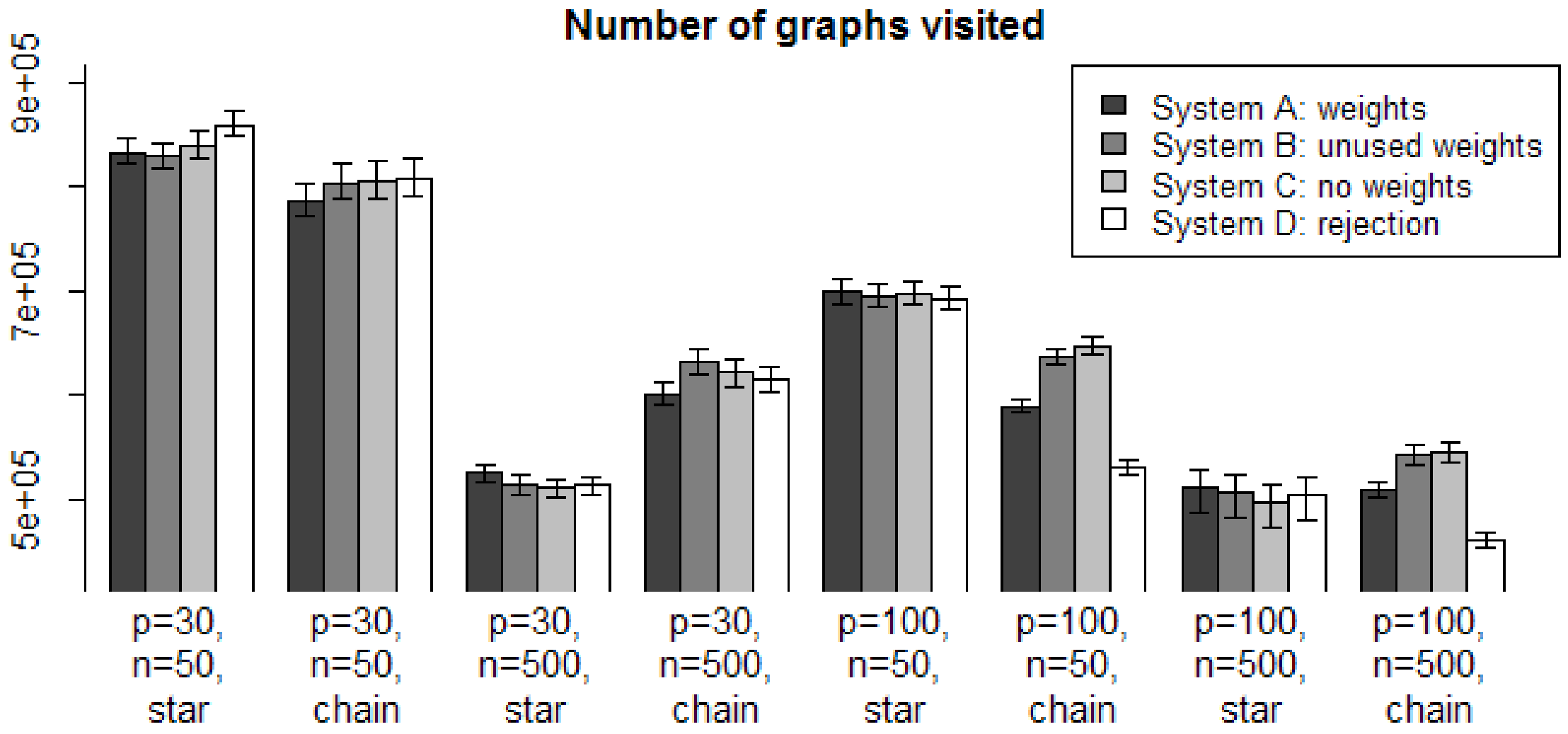

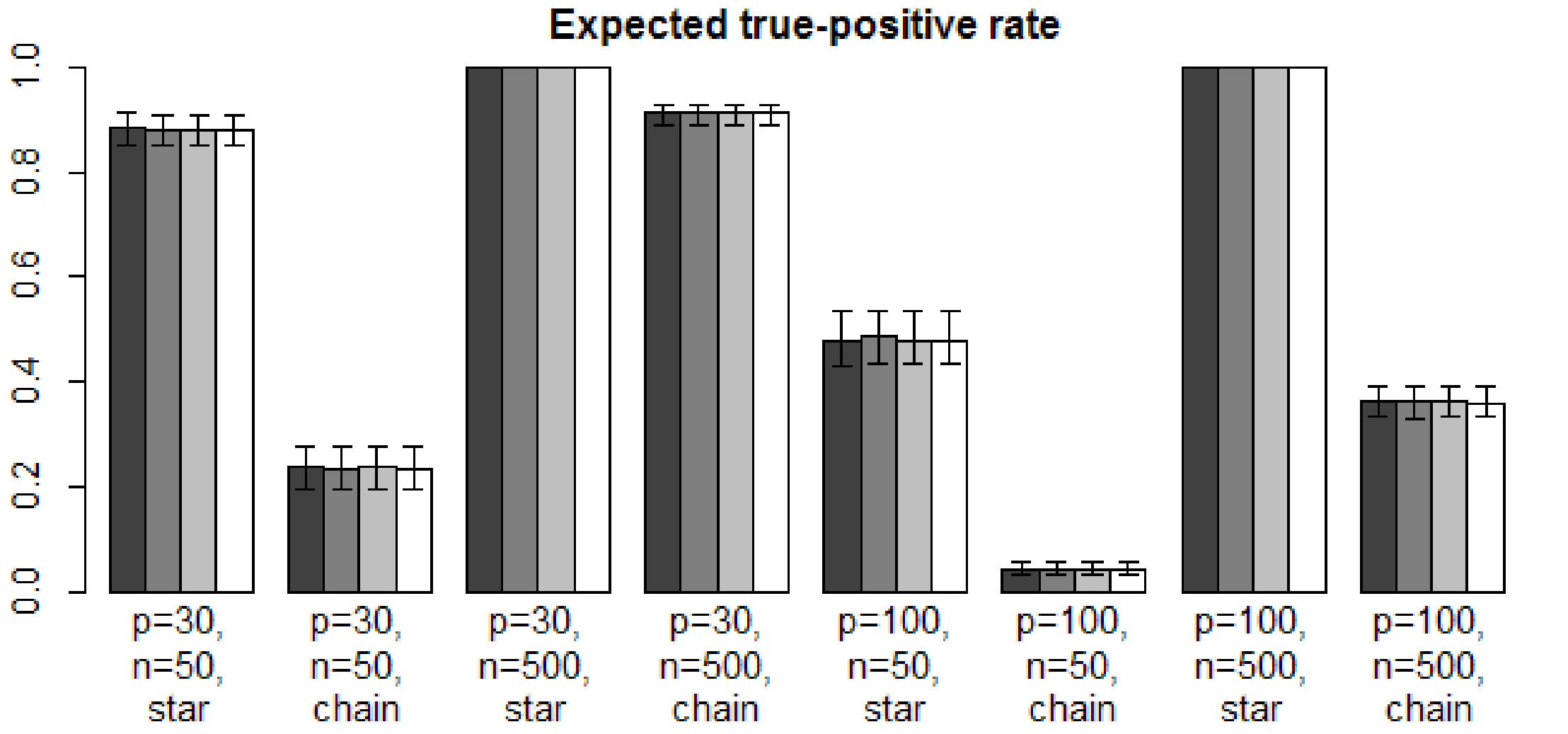

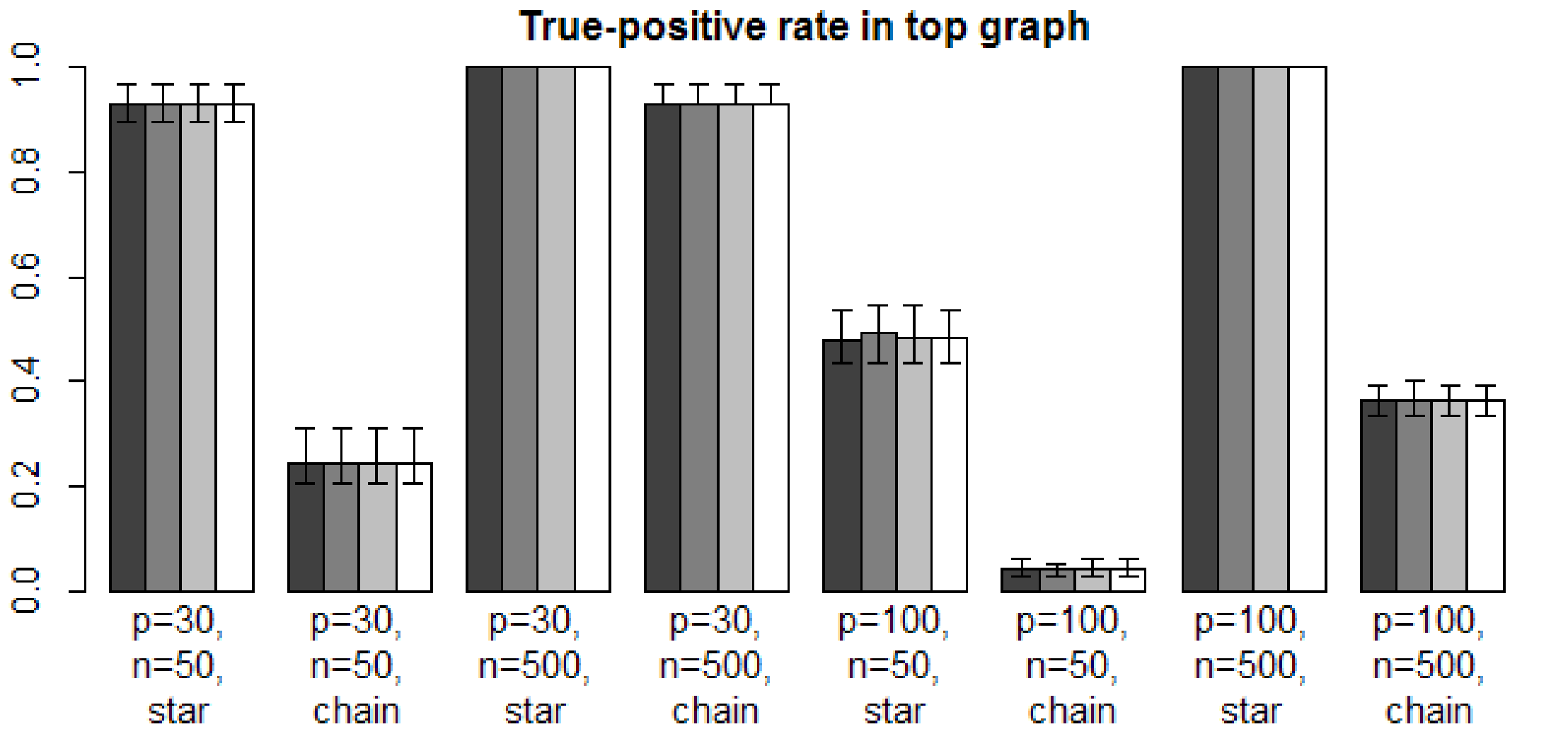

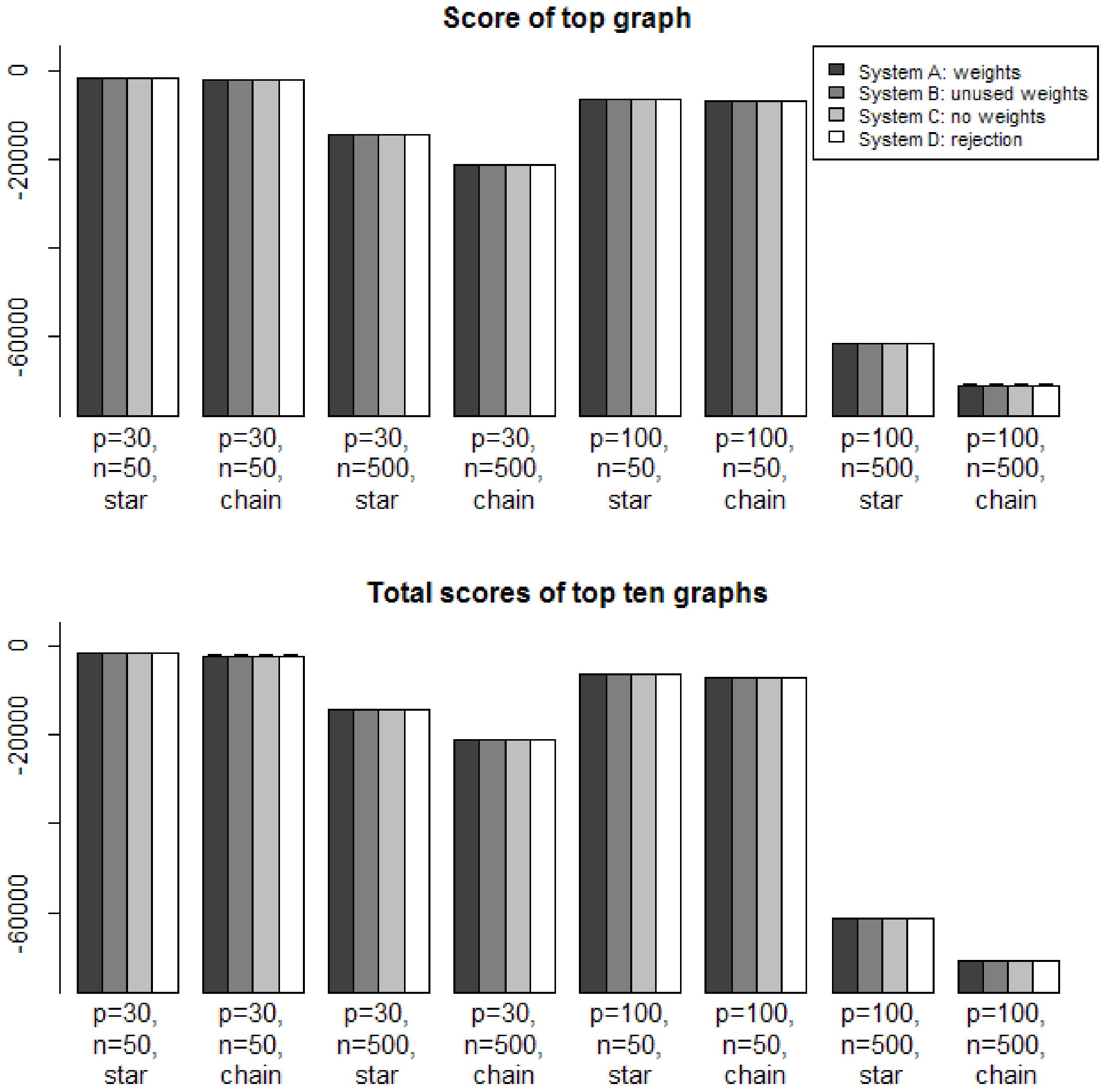

**Figure 11.2 (previous page and this page).** Comparison of four different systems for storing trees in SSST. Each bar-chart shows one measure of how well the four systems did on eight sets of 500 datasets. The heights of the bars are the median values and the "whiskers" show the 25% and 75% quartiles.

## Variation with single datasets

SSST is a random algorithm, so even with a single dataset the results might vary from one run to the next. Figure 11.3 shows how the values vary between different runs on eight particular datasets. Almost all the "whiskers" are very close to the tops of the bars, showing that there is little variation between runs.

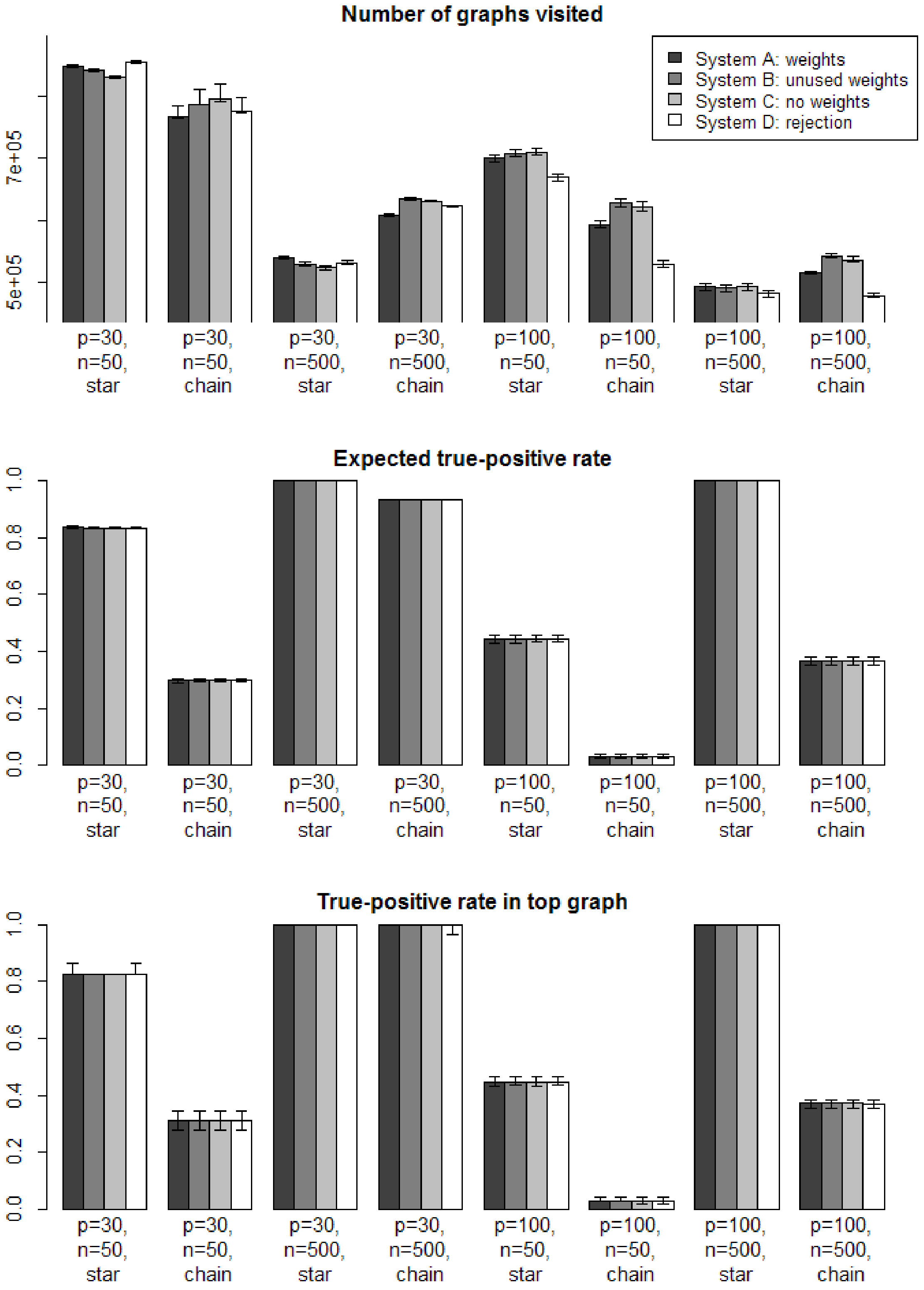

**Figure 11.3.** The variation between different runs of SSST on eight datasets. These bar-charts show the same things as Figure 11.2 except that each bar corresponds to 500 runs of the algorithm on the same dataset.

### False-positives in chains

The measures in Figure 11.2 are mostly lower for chains than for stars. One question that arises is whether there is some pattern to the high-probability graphs that are visited when the true graph is a chain. For example, do they tend to have false-positive edges between nodes that are two apart in the true graph?

Figure 11.4 shows the expected proportions of false-positives that were of this type, for the chain graphs, as produced by SSST. These proportions are all low, showing that there was not much tendency to find these edges. This is somewhat surprising, since it means that graphs that link further-apart nodes have higher probabilities or are more likely to be visited. On the other hand, the interquartile ranges are large.

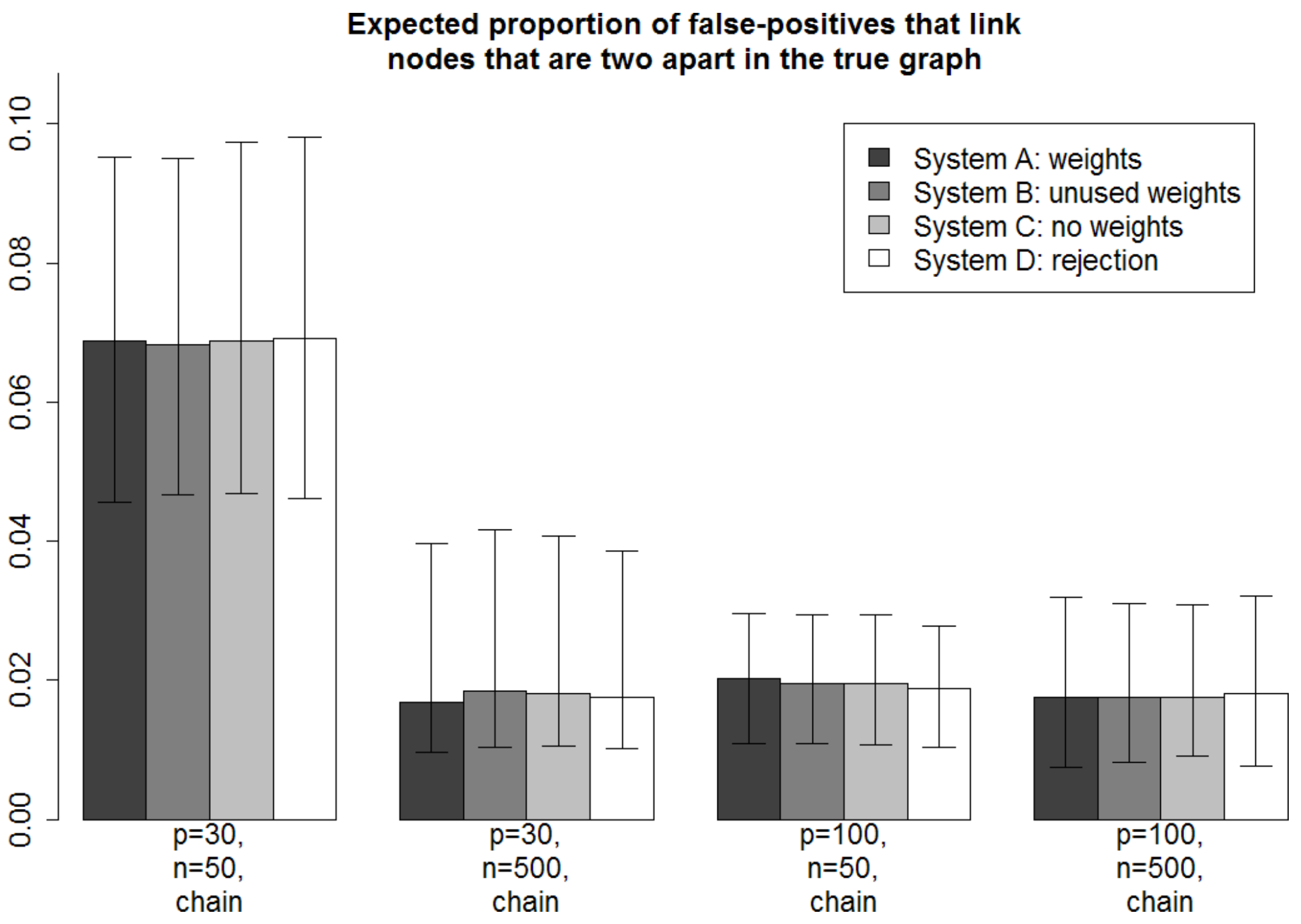

**Figure 11.4.** Expected proportion of false-positives that link nodes that are two apart in the true graph, for the chain graphs.

## 11.3 Experiments with non-forests

The experiments in this section address the question of whether restricting attention to trees gives reasonable results in the case that the true graph is not a tree or forest, but is locally tree-like (see section 6.2). I used SSST, with System A, on datasets generated from graphs that were generated from the second Erdős–Rényi model, where the number of edges is fixed. In this section I will call this model $G(p, M)$, where $p$ is the number of nodes and $M$ is the number of edges. Theorem 6.2 implies that these graphs should have few short cycles and thus be locally tree-like.

For these experiments, generating a dataset consists of randomly generating a graph, then creating a covariance matrix, and finally generating from the multivariate Gaussian distribution. To ensure that the graphs had some cycles, I chose $M$ to be greater than $p - 1$.

Results are shown in Figure 11.5, in the light grey bars. The first and second bars in each group correspond to stars and chains and are taken from Figure 11.2. These are shown for comparison. In the second bar-chart, the white bars show the true values of $ETPR$ for the datasets that correspond to Erdős–Rényi graphs. These were calculated using the MTT-based methods from chapter 8.

Because the algorithm is restricted to trees, all the graphs in the posterior distribution have $p - 1$ edges. But the true graphs have $M$ edges, and $M > p - 1$. So it is impossible for any graph in the posterior distribution to achieve a true-positive rate greater than $(p - 1)/M$. These maximum achievable true-positive rates are shown by thick lines in the second and third bar-charts.

In several cases the results for the Erdős–Rényi graphs are better than the results for the chains. In all cases they are at least similar. Overall, the values are reasonably high. This provides some evidence that the restriction to trees is acceptable for these locally tree-like Erdős–Rényi graphs.

SSST gives very similar values of $ETPR$ to MTT for the datasets with $p = 30, n = 500$ and $p = 100, n = 50$. This means that SSST estimates $ETPR$ very accurately in these cases. With the other datasets it somewhat overestimates $ETPR$.

## 11.4 Experiments with MCMC on forests and trees

### About the experiments

The MCMC algorithms used in this section are McmcF, where only forests are considered, and McmcT, where only trees are considered. These are described in section 10.1. The datasets were generated in the same way as the ones in section 11.1, and the values of $p$ and $n$ are stated below. McmcF and McmcT have two parameters that can be set, $\sigma_G$ and $\sigma_{ij}$. The former is used in the updates to the graph structure and the latter is used in the updates to the covariance matrix.

The findings of this section, in summary, are that McmcF usually fails to mix, and that McmcT mixes but takes much longer than SSST to give useful results. First I will describe the experiments and then I will discuss the results.

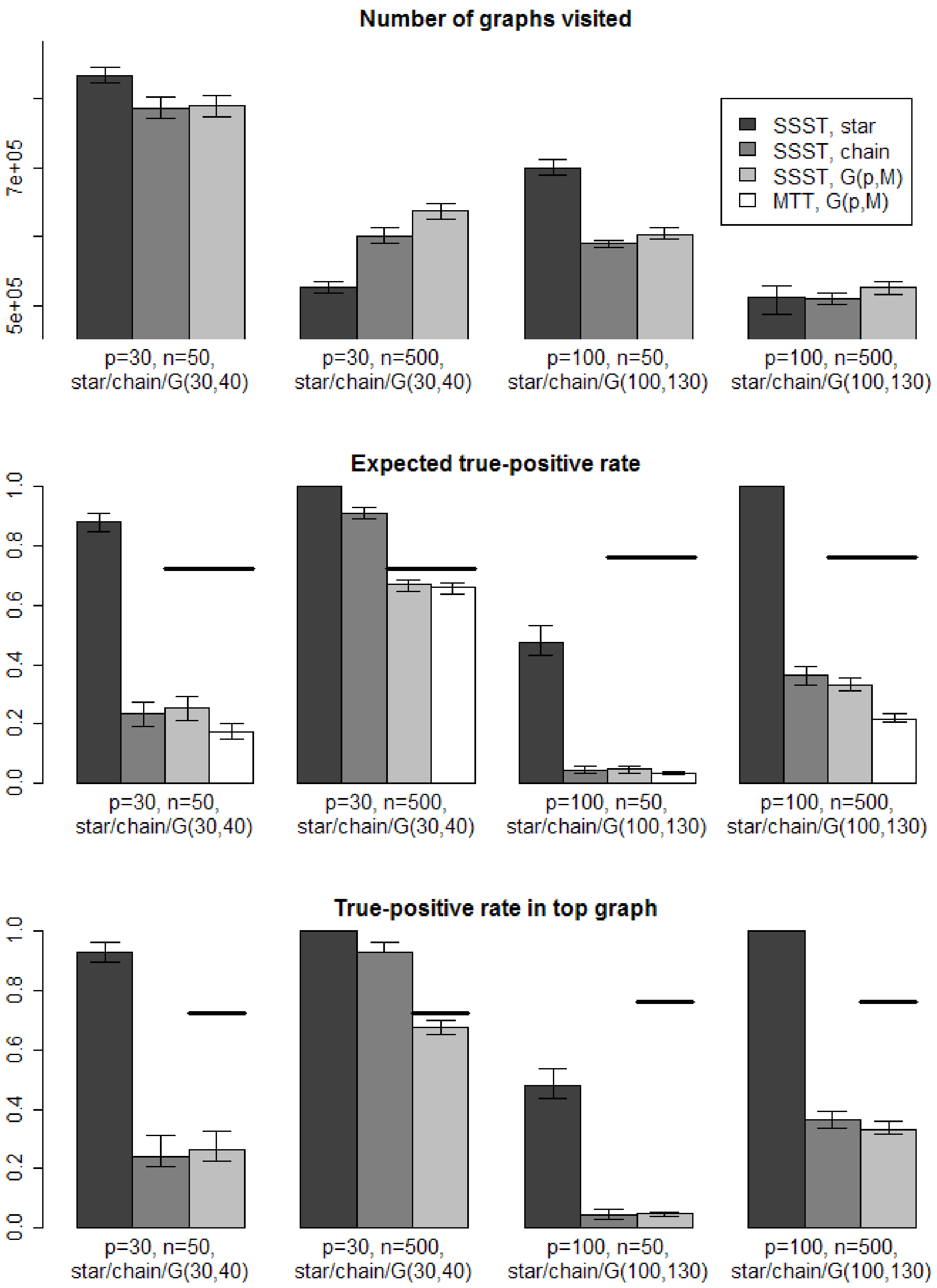

**Figure 11.5.** The light grey bars (the third bar in each group) show measures of how successful SSST is with non-forest graphs generated from $G(p, M)$, the second Erdős–Rényi model. Each bar corresponds to 100 datasets, each generated from a covariance that matches a different graph. The darker bars show values from section 11.2, for comparison, and the white bars show the true values of $ETPR$. The thick lines show maximum achievable values for the experiments with the Erdős–Rényi graphs.

## Experiments with McmcF

*First dataset*

The first set of experiments was done on a dataset with $p = 5$ and $n = 30$, generated from a distribution for which the true graph was a star. McmcF was run with a range of values of $\sigma_G$ (specifically, 0.01, 0.02, 0.05, 0.1, 0.2, 0.5, 1, 2, and 5), for 20 million iterations in each case. In the true distribution on forests, the probabilities of the top two graphs are 0.39 and 0.08. With $\sigma_G = 0.05$, McmcF got the top graph right but estimated its probability to be 0.96. With the other values of $\sigma_G$, McmcF got the top graph wrong and estimated the probabilities of these wrong top graphs to be 0.43 or more. Clearly McmcF failed totally. (In this and the other MCMC experiments there was no problem with the updates of $\Gamma$. In this case $\sigma_{ij}$ was 0.01 and the acceptance-rate for updates of $\Gamma$ was between 77% and 80%.)

In all cases the most-visited graph was a tree. To see whether McmcF might be getting stuck in local optimums, and what kinds of graphs these local optimums might be, I calculated the scores of all the possible forests, in other words the exact true posterior distribution. The top 125 graphs were trees—all the trees had higher scores than all the unconnected forests. The top graph was 1.1 million times more likely than the top unconnected forest.

*Second dataset*

The second set of experiments was done on a dataset with $p = 5$ and $n = 10$. Again the true graph was a star. Because $n$ is smaller this might be expected to give a less peaked posterior distribution and mix better (Friedman & Koller 2003). McmcF was run with the same values of $\sigma_G$ as for the first dataset, for 20 million iterations in each case. For seven of these experiments it got the top graph wrong. With $\sigma_G = 5$ it got the top graph right, and with $\sigma_G = 0.5$ it got the top three graphs right and their probabilities right to within 8%. This was much better than with the first dataset, though it was surprising that the two values of $\sigma_G$ that worked best were so far apart.

Again the most-visited graph was a tree in all the experiments. In the exact true posterior distribution, the top 125 graphs were again all trees. But this time the top graph was only 13.8 times more likely than the top unconnected forest.

*Third dataset*

The third dataset had $p = 5, n = 10$, and the true graph as in Figure 11.6. The purpose of this was to see whether McmcF might mix better when the true graph is not a tree. McmcF was run for $\sigma_G = 0.1$, 0.2, and 0.5, for 20 million iterations in each case. All three times, it correctly identified the two most likely graphs. Figure 11.7 shows the top few graphs in the true posterior distribution and for McmcF with $\sigma_G = 0.1$ (of the three values, this gave the highest acceptance rate for graph updates, 16%). McmcF seems to have got stuck in certain trees for longer than it should have.

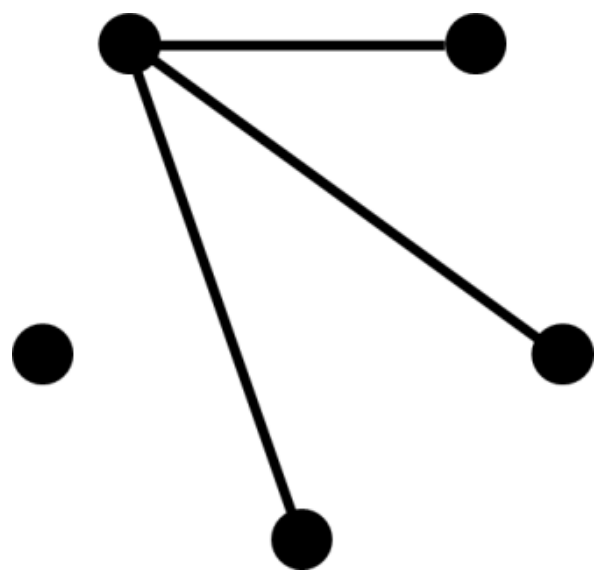

**Figure 11.6.** The true graph, for the third dataset for McmcF.

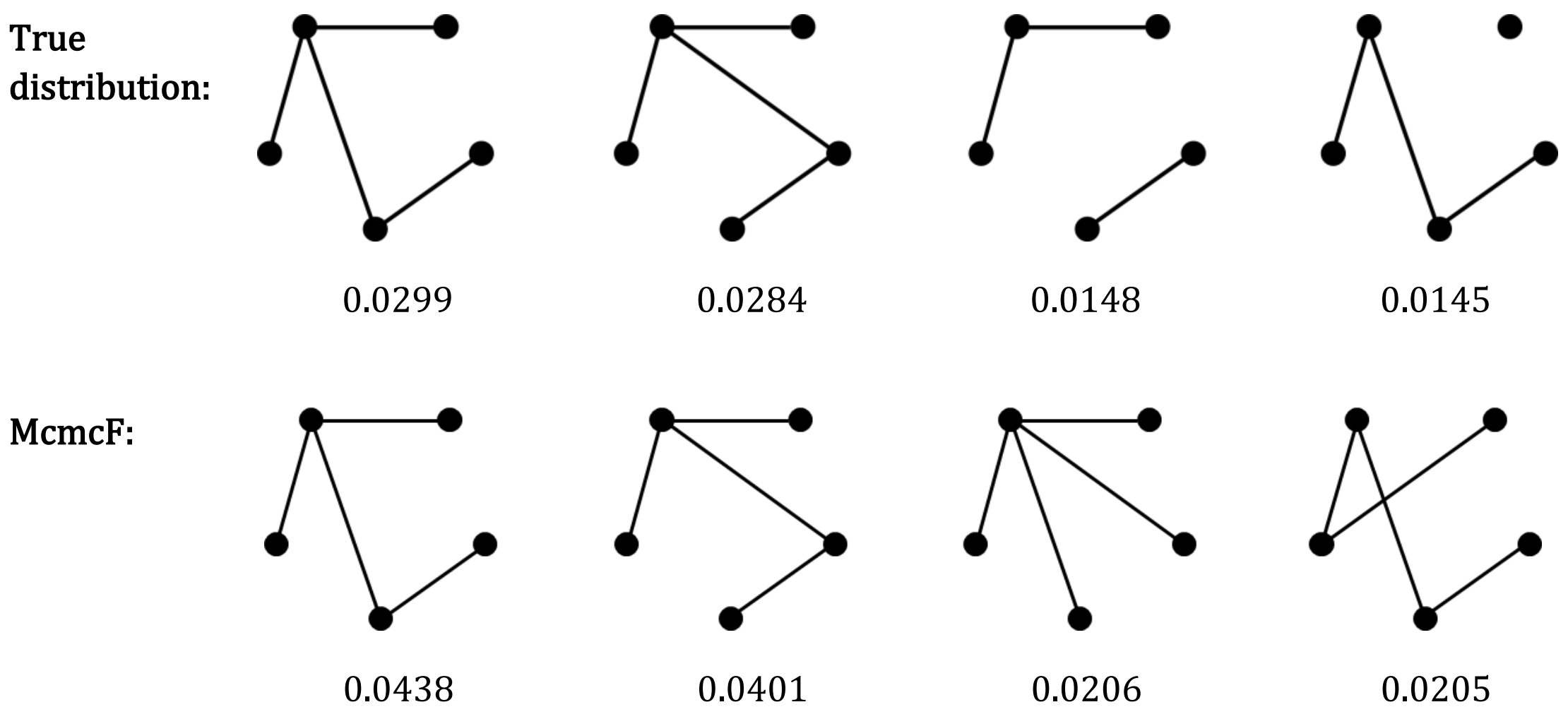

**Figure 11.7.** For the third dataset used with McmcF, the four most likely forests and their probabilities according to the true posterior distribution (restricted to forests) and according to McmcF with $\sigma_G = 0.1$.

*Other datasets*

McmcF works less well with $p > 5$. For example, it often spends more than half the time at a single graph or has an acceptance rate for graph updates of less than 0.01%. It works well for $p = 4$, and gives the correct posterior distribution, but this is no use since exhaustive search of all the possible graphs takes only a few seconds when $p = 4$.

## Experiments with McmcT

I ran McmcT on datasets like the ones used in section 11.2, with $p = 30$ or $p = 100$, using a range of values of $\sigma_G$. It failed to mix with the datasets where $n = 500$, but it mixed well with the ones where $n = 50$. Setting $\sigma_{ij} = 0.01$ usually seems to give an acceptance rate of 40% or 50%, which in MCMC is generally regarded as good. Setting $\sigma_G$ requires more trial and error. If $\sigma_G$ is too small or too big, the updates almost never get accepted, but if it is chosen appropriately then McmcT mixes, at least with the datasets where $n = 50$.

The question arises of whether McmcT gives a reasonable approximation to the true posterior distribution in a reasonable amount of time, or whether it merely mixes well among graphs that do not have high scores. I ran McmcT on two datasets with $p = 30$ and $n = 50$, and recorded several quantities at certain intervals, to see how well it was mixing and how long it was taking. I used $\sigma_G = 0.5$. Figure 11.8 shows the number of distinct graphs visited, $ETPR$, and the true score of the most-visited graph, for these two experiments. (For McmcT, the true score of the most-visited graph can be used as a measure of how well the algorithm does. This is analogous to the highest score found in SSST.)

For both datasets, the number of distinct graphs visited steadily increased. McmcT visited far more graphs with the chain dataset than with the star dataset, again showing that these two types of graph are greatly different. According to the other two quantities, McmcT did somewhat less well than SSST for both graph-shapes. (For the SSST results see Figure 11.2.) Taking the chain as an example, with McmcT $ETPR$ settled around 0.144, but with SSST its median was 0.235, and with McmcT the true score of the most-visited graph settled around $-2125$, but with SSST the highest score found was $-2107$ (which is better).

SSST was only run for 20 seconds, but to do 10 million iterations took McmcT 7 hours and 13 minutes for the star and 9 hours and 5 minutes for the chain. Figure 11.8 shows that at least 1 million iterations are needed to get a reasonable result. Overall, SSST seems to be better in practice than McmcT, firstly because with McmcT it is necessary to experiment to find suitable values of $\sigma_{ij}$ and $\sigma_G$, and secondly because, once the main algorithms are underway, SSST gives reasonable results much faster. For these reasons, I use SSST in the subsequent sections of this chapter.

## The failure of McmcF

Asymptotically, McmcF produces the true posterior distribution. Given a long enough time, it would produce a good approximation. But for $p > 5$ or large $n$, if it is run for a reasonable length of time or a reasonable number of iterations, it does not give a good approximation of the posterior distribution and sometimes completely fails to mix.

The reason is probably the high peaks that often seem to appear in the posterior distributions. When the true graph is a tree, all the high-scoring graphs tend to be trees, and these trees have much higher scores than the highest-scoring unconnected forests. But for McmcF to get from a tree to another tree it has to first visit an unconnected forest. (Obviously, this problem does not arise with McmcT.)

For example, with the first dataset discussed above, the unconnected forests all had far lower scores, in the area of a million times lower, than the high-ranking trees. This shows that the posterior distribution is very peaked and multimodal. There was also a peak at the top-ranking graph, which is also the true graph. Its score is 5 times the score of the next graph. This peak probably results from $n$ being large compared to $p$.

The obvious way to adapt McmcF would be to have different types of moves in the graph spaces. For example, you could add or remove two or more edges at a time. The system for storing forests described in section 9.3 would have to be adapted, but some of the ideas would still be useful. Another possible adaptation would be to sometimes move edges, instead of just adding and removing them—though the algorithm might end up only visiting trees, in which case you might as well use McmcT.

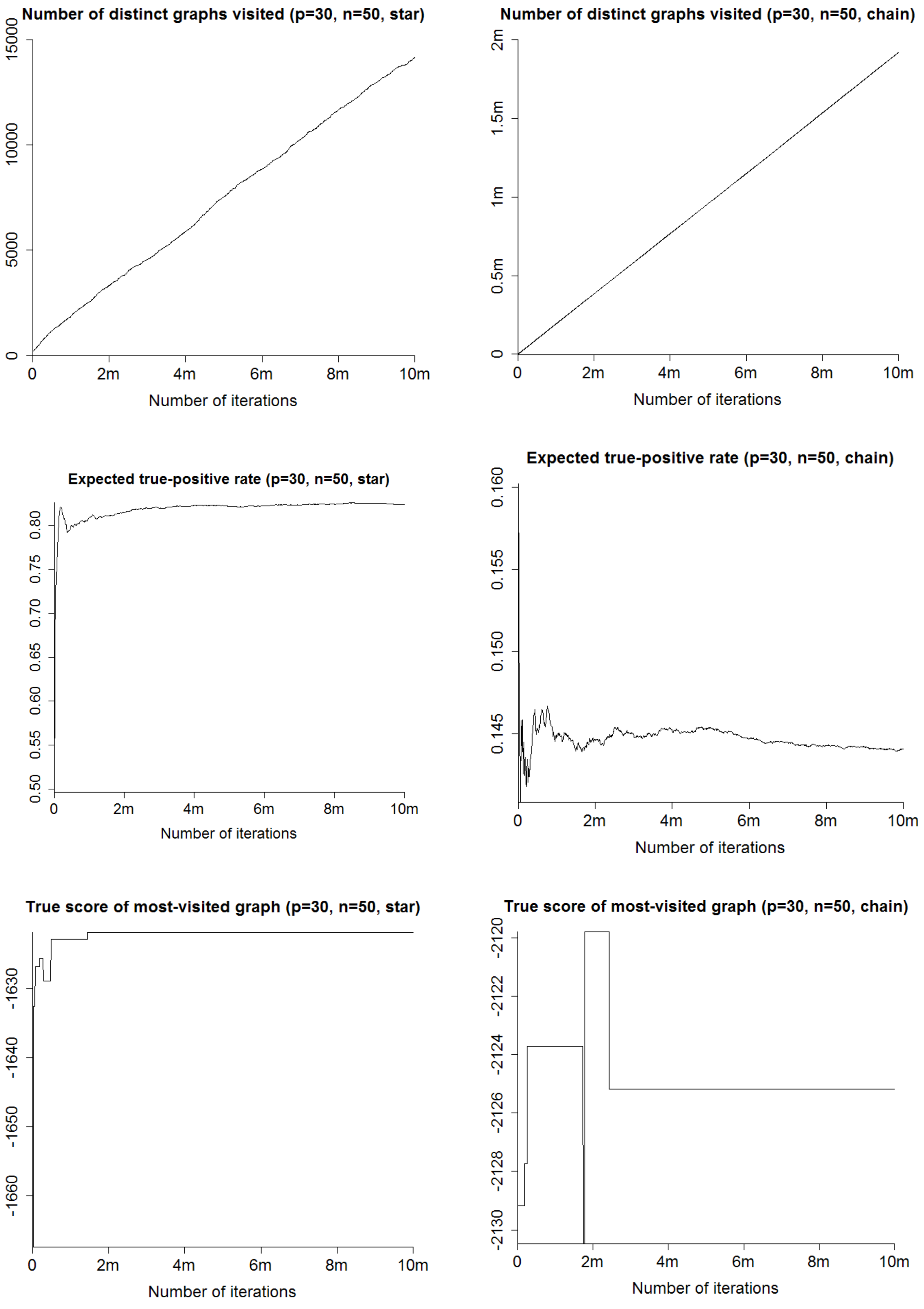

**Figure 11.8.** How three aspects of the estimated posterior distribution change over 10 million iterations of McmcT on two datasets.

Different types of graph-moves is not the only possibility. Karagiannis & Andrieu (2012) describe a method that addresses the wider problem of reversible-jump MCMC algorithms getting stuck. To make a proposal, their algorithm chooses a dimension-changing move and then moves about in the new model-space to find parameters that will give higher acceptance probability. The method has desirable asymptotic properties, and it can be applied to the MCMC method of Giudici & Green (1999), though this application is not addressed in the paper.

### Similar observations in other research

Friedman & Koller (2003) discuss in some depth the issue of MCMC for graphical model structure-learning failing to mix well. Their paper is mainly about directed graphical models but also covers undirected ones. They say that MCMC on the graph structure is slow to mix because the posterior distribution is often peaked, meaning that neighbouring graphs have very different scores. Even small changes such as removing an edge cause large changes in the posterior probability. If $n$ is large then the posterior will be sharply peaked at a single model. This corresponds to what I found with McmcF and the first dataset.

Friedman & Koller (2003) state that "in small domains with a substantial amount of data, it has been shown that the highest scoring model is orders of magnitude more likely than any other." But the source they cite, Heckerman et al (1997), only shows this in one specific example.

Altomare et al (2011) say it is now recognized that MCMC methods are not efficient for these problems, because of the huge number of possible graphs and the multimodal posterior distributions. Scott & Carvalho (2008) make similar comments. Brooks et al (2003) discusses the general issue that in reversible-jump MCMC it is difficult to come up with proposals that will get accepted a reasonable proportion of the time. They suggest adapting the method of Giudici & Green (1999) by retaining the previous values of the elements of the covariance matrix and using them in choosing the proposed new values.

## 11.5 Experiments with methods for trees

Of the quantities in section 11.2, the MTT-based method from section 8.3 can only produce $ETPR$. Figure 11.9 shows $ETPR$ for SSST, run for three different lengths of time, and the MTT-based method. These experiments used the same settings as in section 11.2, and 100 datasets for each group of four bars. The MTT-based method is deterministic and gives exact values, so this is really an assessment of how well SSST approximates the true posterior distribution. SSST does reasonably well, though it overestimates $ETPR$ in the case of the chain graphs. Perhaps SSST only visits high-probability graphs, but the low-probability graphs have fewer true-positives and still make a noticeable difference to $ETPR$.

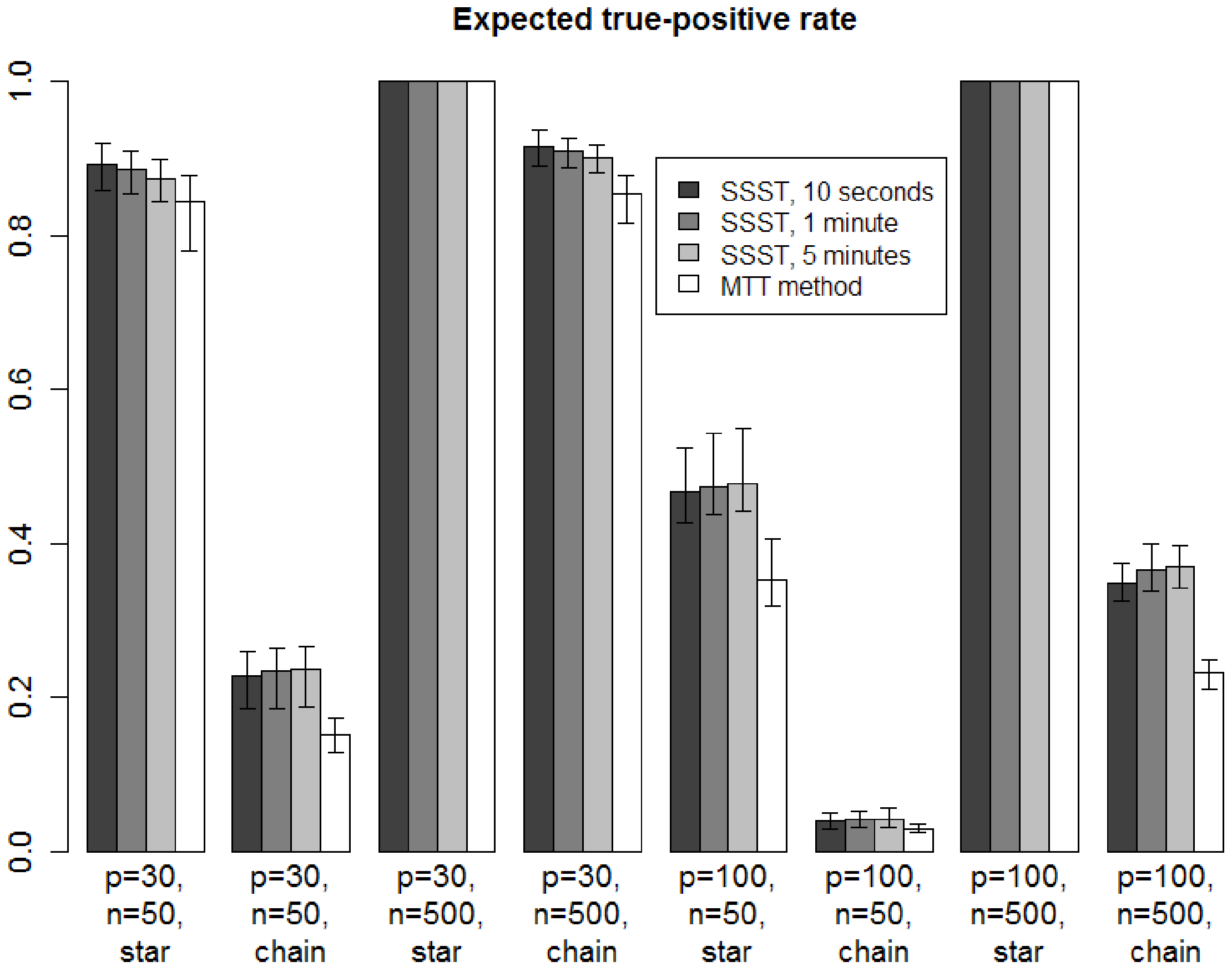

**Figure 11.9.** Comparison of SSST with the MTT-based method. Each group of four bars corresponds to one combination of covariance matrix and $n$, and 100 datasets.

## 11.6 Experiments with graph prior distributions

To see whether hub-encouraging graph priors can give better results than the uniform graph prior, in the case when the true graph is a star, I ran SSST with the following four graph priors.

- the hub-encouraging prior from section 5.7 with $\psi = 1$ and $\chi = 0.9p$
- the hub-encouraging prior from section 5.7 with $\psi = 0.01$ and $\chi = 0.9p$
- the prior defined by $p(G) \propto \exp(\max \deg(v))$
- the uniform graph prior.

Small values of $\psi$ were used because larger values gave almost no improvement over the uniform prior. The third prior was intended to be strongly hub-encouraging. It gives much higher probability to graphs that have a single hub, and much higher probability to graphs where that hub has higher degree. The four priors were used on the datasets from section 11.2 for which the true graph was a star, and the results are shown in Figure 11.10.

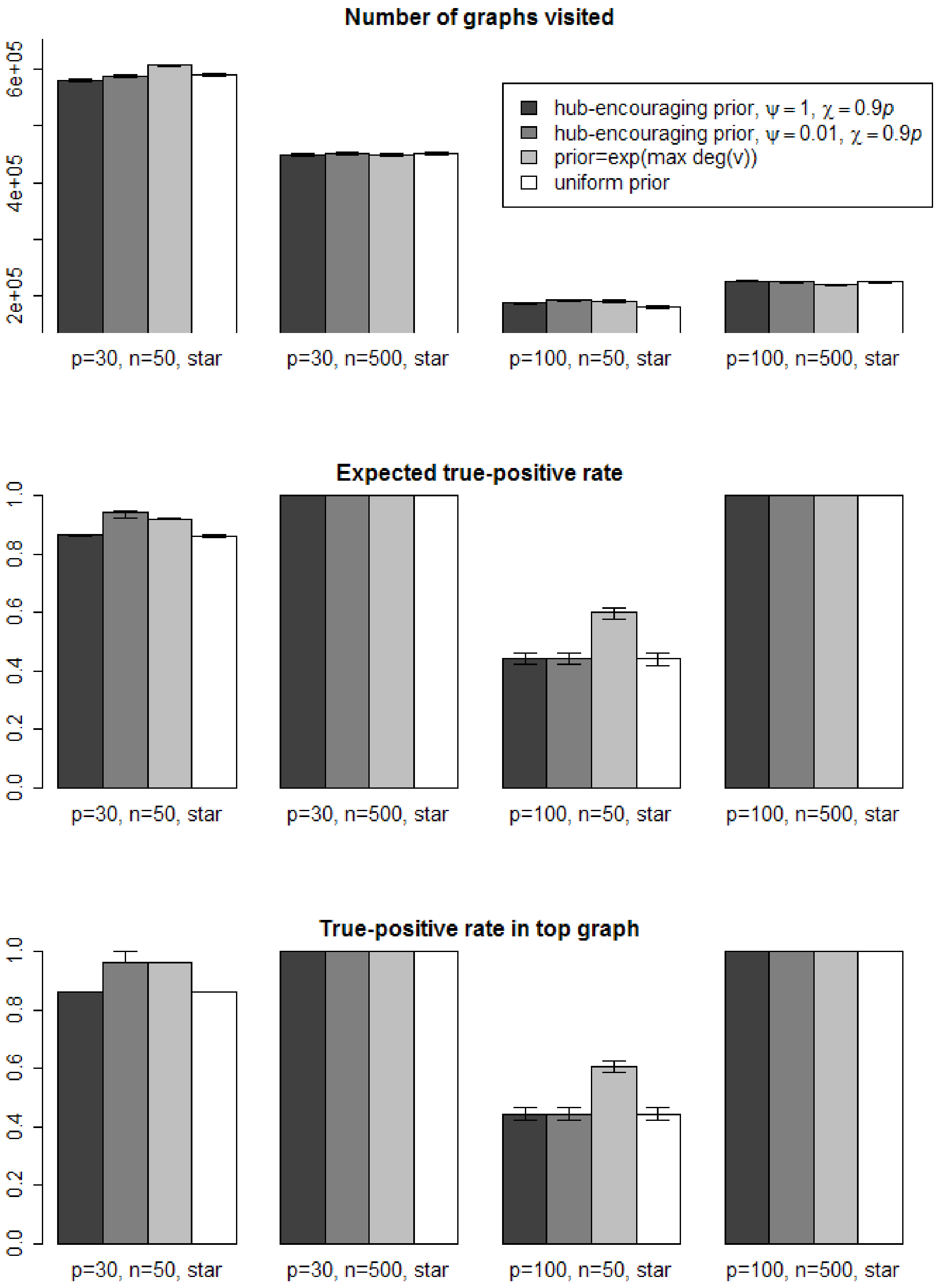

**Figure 11.10.** Experiments to compare hub-encouraging priors with the uniform prior, on four sets of datasets. Each bar-chart shows one measure of how well the algorithm did. Each group of four bars shows three hub-encouraging priors and the uniform prior. (As before, each bar shows results with 500 datasets generated from the distribution with the same covariance matrix.)

The first hub-encouraging prior, with $\psi = 1$ and $\chi = 0.9p$, gave little or no improvement over the uniform prior, and the second, with $\psi = 0.01$, gave some improvement. Overall the third prior was the best.

Evidently, graph priors need to give much greater probability to some graphs than to others if they are to have any effect on the posterior. This is presumably because among the marginal likelihoods of all the graphs, some values are many orders of magnitude greater than others. This suggests that one should look at the range of marginal likelihoods and then decide a suitable range for the graph prior, but that would go against the fundamental principles of Bayesian inference.

As in section 11.2, the datasets with $p = 100$ and $n = 50$ gave the lowest values. But 0.5 or 0.6 are still not bad for the quantities in the second and third bar-charts.

## 11.7 Experiments with forests, trees, and decomposable graphs

The final set of experiments are a further investigation of whether restricting to trees or forests is sensible. I compared SSST and SSSF with one of the original versions of Jones et al (2005)'s stochastic shotgun search algorithm. Jones et al (2005) described versions of their algorithm for both decomposable and general graphs, but found that searching general graphs "becomes very challenging" as $p$ increases past 15. For this reason I used the version that is restricted to decomposable graphs. I will call this SSSD. (Details of how my programs stored decomposable graphs are given at the end of section 10.1.)

The same datasets were used as in section 11.2. The algorithms were all run for 60 seconds on each dataset. For SSSF the values of $\omega$ used in previous sections were too big, because there are often not that many possible moves, so for all the experiments in this section I used $\omega = p/2$.

The results are shown in Figure 11.11. According to the second and third bar-charts, SSST and SSSF did better than SSSD when $n = 500$ and the true graph was a chain, SSSD did best when $n = 50$, and the three algorithms did roughly as well as each other in the other cases. Overall these bar-charts provide some further reassurance that the restriction to trees or forests is reasonable.

As shown by the first bar-chart in Figure 11.11, SSST and SSSF visited far more graphs in the same amount of time than SSSD. But the three types of graph have very different implementations. It might be said that SSSD was not given enough time to visit a reasonable number of graphs. So I repeated the experiment but ran each algorithm for 500 iterations rather than 60 seconds.

The results are shown in Figure 11.12. On average, SSSD took 14.9 times longer than SSST. But SSST still did better than SSSD according to some of the groups of bars and not much worse according to the others. (Obviously it is still not completely fair to compare the numbers of graphs visited by the three algorithms.)

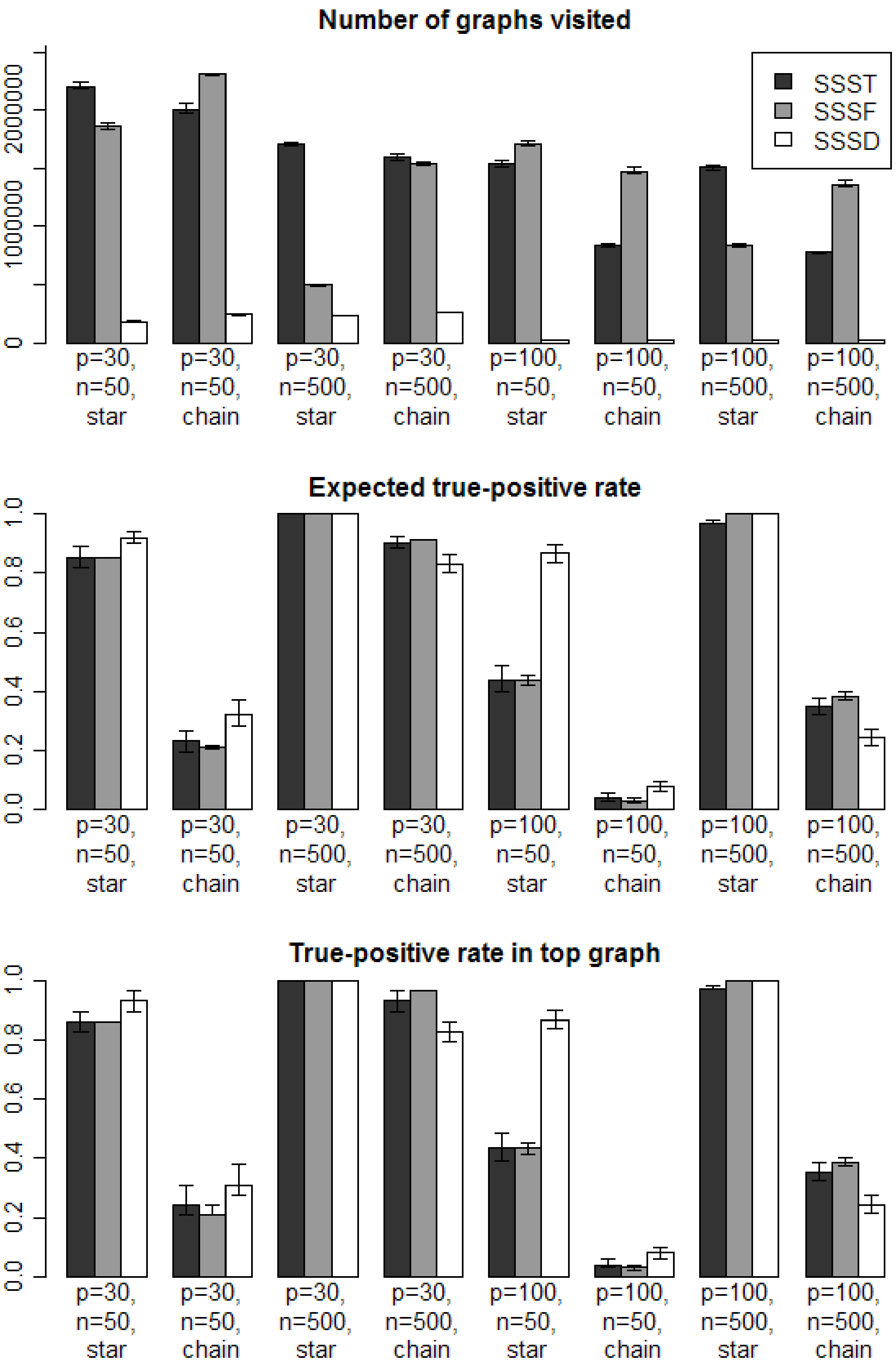

**Figure 11.11.** Comparison of SSST, SSSF, and SSSD, using the same datasets as in section 11.2. Each algorithm was run for 60 seconds.

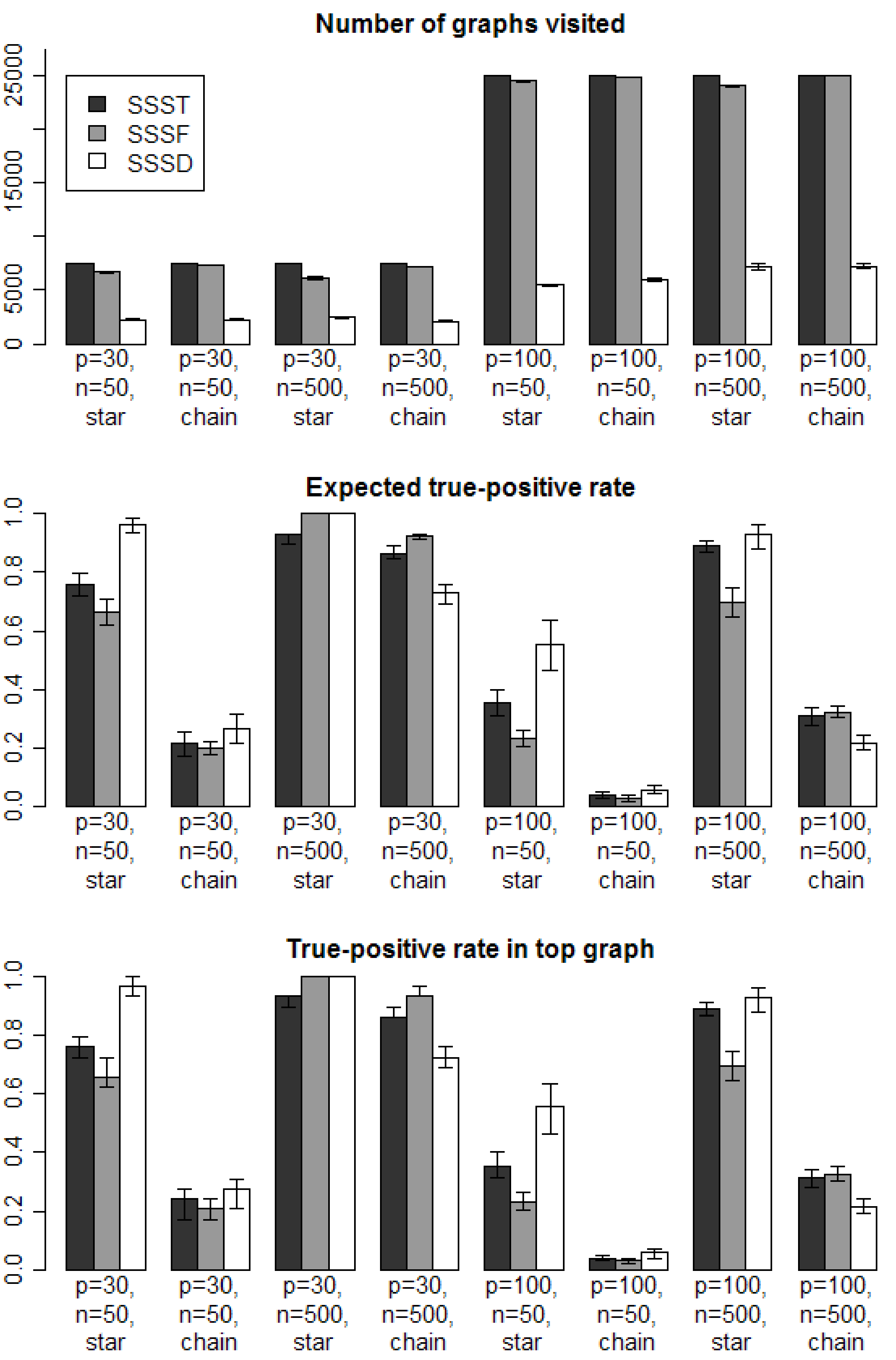

**Figure 11.12.** Comparison of SSST, SSSF, and SSSD. Each algorithm was run for 500 iterations, to give SSSD a chance to visit a reasonable number of graphs.

# 12 Conclusions

## 12.1 Restricting to forests and trees

The reasons in favour of restricting attention to forests or trees, in Bayesian structure-learning of graphical models, can be summarized as follows. Chapter 6 showed that there has been plenty of theoretical and applied research using forests and trees and gave theoretical reasons in favour of them. Chapters 7 and 8 gave fast algorithms that can be used on them. Chapter 11 provided empirical evidence based on several experiments. Firstly, SSST gave good results in terms of $ETPR$ and the other measures, especially for star graphs and even when $n < p$. Secondly, SSST did reasonably well with sparse and locally tree-like graphs that were not forests. Thirdly, SSST did almost as well as SSSD according to the true-positive rates and much better according to the numbers of graphs visited, though these experiments had the drawbacks that the true graphs were trees and the algorithms are not easy to compare because their implementations are so different. SSSF also did better than SSSD on some groups of datasets.

Bayesian structure-learning has no difficulty with $n < p$ and often gives good results in terms of true-positive rates, though naturally it is unlikely to give high $ETPR$ if $n \ll p$. Restricting to forests or trees automatically overcomes the problem of stars being misidentified as cliques. It would be interesting to compare graphs produced by the algorithms in Albieri (2010) with graphs produced by the Chow–Liu algorithm and the adaptations of it in chapter 7.

Which method is best depends on the purpose of the analysis. But overall it seems entirely plausible that there are practical circumstances in which it would be preferable to do Bayesian structure-learning on forests and trees, rather than on decomposable graphs or all graphs.

## 12.2 Graph distributions and theoretical results

The two ways of looking at graph distributions from chapter 5 should be helpful in clarifying ideas about graph distributions. Factored distributions are at least theoretically useful because they can be used in algorithms based on the Chow–Liu method and MTT.

None of the distributions previously used as priors in Bayesian structure-learning is satisfactory for encouraging hubs. The criteria and priors proposed in section 5.7 are more suitable for this purpose.

The theoretical results in section 6.2 showed how the claim that sparse graphs are locally tree-like can be made rigorous and used to justify restricting attention to forests or trees. A possible topic for future research is whether similar theoretical or empirical results can be found for other random graph models such as scale-free graphs.

Chapter 4 gave a proof of correctness for the algorithm for recursive thinning that is used in the R package gRbase, which is one of the main constituents of the large-scale project called "gRaphical models in R". It also gave a simpler algorithm that is sometimes faster.

## 12.3 Algorithms for structure-learning with forests or trees

Chapter 7 discussed how methods based on the Chow–Liu algorithm can be used with GGMs to find the maximum-likelihood tree, the optimal forest using likelihood penalized by AIC or BIC, and the MAP forest in Bayesian structure-learning. Chapter 8 showed how the method based on MTT can be used to find certain types of information about the posterior distribution over all trees. All these methods are very fast compared to any attempt to approximate the whole posterior distribution, and they all work with factored prior distributions. The drawbacks are that they can only answer certain types of questions.

Chapter 9 presented efficient systems for storing forests and trees so that single-edge moves could easily be chosen uniformly at random and the stored information could easily be updated. It might be worth investigating other ways of choosing edge-moves in trees, such as adding an edge, identifying the cycle that results, and then removing an edge, to see if they perform better.

## 12.4 Computer experiments

Many of the experiments in chapter 11 used only four different graphs and two different values of $n$ (the number of data), so there is obviously plenty of scope for more experiments. For example, the algorithms could be used on data generated from different shapes of graphs—perhaps Erdős–Rényi graphs of the first type, scale-free graphs, or specific real-world graphs. It would also be interesting to try them on gene expression data or financial data. Most of the experiments used only trees, so further research might use forests instead.

The SSS algorithms are designed to be run on parallel or distributed processors. For datasets with much higher $p$ it would probably be necessary to use multiple processors. It is certainly advisable to use a fast programming language—I found that Java is 100 times faster than R.

In experiments using SSST, the system for storing trees had some effect on how many graphs were visited but almost no effect on $ETPR$ or the other measures. $ETPR$ and the true-positive rates in the top graphs were generally high, especially for the star graphs. Especially good was the result that $ETPR$ was roughly 0.5 for the datasets with $p = 100$, $n = 50$, and the true graph a star. SSST did reasonably well on datasets generated from sparse and locally tree-like graphs that were not forests. This was the first piece of empirical evidence that it may be sensible to restrict attention to trees.

McmcF did not mix well. McmcT mixed well but took far longer than SSST to give reasonable results, and required trial-and-error to find suitable values of the parameters for the proposal distributions.

In the experiments with the MTT-based methods, the approximations of the true posterior distributions produced by SSST gave higher values of $ETPR$ than the true posterior distribution. This was especially the case for chain graphs. In a sense this is evidence that SSST works well, though it does not address the question of whether restricting to trees is sensible.

Next were experiments with graph prior distributions that were designed to encourage hubs. The priors proposed in chapter 5 had small effects in some cases, but the more extreme prior with $p(G) \propto \exp(\max \deg(v))$ was better at identifying the hub when $n < p$. If graph priors are intended to encourage hubs then they need to give much higher probability to some graphs than others. Obviously there is a large amount of scope for further experiments with graph priors that encourage hubs, scale-free degree sequences, or other features that are believed to be common in real-world networks. Hub-encouraging priors could also be used with other algorithms for GGM structure-learning, such as the MCMC method of Green & Thomas (2013), which works with junction trees.

Lastly, section 11.7 compared SSST, for trees, SSSF, for forests, and SSSD, for decomposable graphs. Given the same amount of time, SSST and SSSF visited far more graphs than SSSD. In terms of $ETPR$ and the true-positive rate in the top graph, SSST and SSSF did better on two sets of datasets, SSSD did best on five, and all three did very close to equally well on one. These results gave further evidence that restricting attention to trees or forests may be sensible. The class of decomposable graphs is bigger but this was outweighed by the computational simplicity of trees or forests. Further research might compare SSST and SSSF with SSSD on data generated from graphs that are not forests but are locally tree-like.

# Appendix I: Graph enumerations

This appendix presents the results of some enumerations of decomposable graphs. These numbers are not important or meaningful. However, they have never been found before, as far as I can tell.

Table A1 shows the number of decomposable graphs with $n$ nodes, for $n$ up to 13. The numbers for $n$ up to 12 are from Sloane (2011), which is sequence A058862 on a website called Online Encyclopedia of Integer Sequences. The number for $n = 13$ does not seem to have appeared anywhere before. I worked it out using a formula on the same webpage and sequence A007134 from the same website. The method for working out these numbers is described in Wormald (1985).

Table A2 shows the number of decomposable graphs with $n = 9$ nodes, for each possible number of edges. The analogous numbers for $n$ up to 8 are given in Table 7.1 of Armstrong (2005). I found the numbers for $n = 9$ by writing a program that does a maximum cardinality search (Tarjan & Yannakakis 1984) on every possible graph, to test whether it is decomposable. This program took one week to run on an average desktop computer. But a parallelized version running on a high-powered computer, with twelve 3GHz processors, did it in 6.5 hours. To do the same thing for $n = 10$ would take much longer, because there are 1024 times more graphs.

| $n$ | Number of decomposable graphs with $n$ nodes | Percent of graphs that are decomposable |
|---|---|---|
| 1 | 1 | 100 |
| 2 | 2 | 100 |
| 3 | 8 | 100 |
| 4 | 61 | 95 |
| 5 | 822 | 80 |
| 6 | 18 154 | 55 |
| 7 | 617 675 | 29 |
| 8 | 30 888 596 | 12 |
| 9 | 2 192 816 760 | 3.2 |
| 10 | 215 488 096 587 | 0.61 |
| 11 | 28 791 414 081 916 | 0.080 |
| 12 | 5 165 908 492 061 926 | 0.0070 |
| 13 | 1 234 777 416 771 739 141 | 0.00041 |

**Table A1.** The number of decomposable graphs with $n$ nodes, for $n$ up to 13.

| $e$ | $n(e)$ | $e$ | $n(e)$ | $e$ | $n(e)$ | $e$ | $n(e)$ |
|---|---|---|---|---|---|---|---|
| 0 | 1 | 10 | 59194170 | 19 | 170178120 | 28 | 1154547 |
| 1 | 36 | 11 | 94169376 | 20 | 130062807 | 29 | 430236 |
| 2 | 630 | 12 | 137060700 | 21 | 92533764 | 30 | 137718 |
| 3 | 7140 | 13 | 181199340 | 22 | 62171838 | 31 | 37800 |
| 4 | 58527 | 14 | 216312390 | 23 | 39638592 | 32 | 10080 |
| 5 | 364140 | 15 | 234891000 | 24 | 23221338 | 33 | 2100 |
| 6 | 1741530 | 16 | 237142836 | 25 | 12310704 | 34 | 252 |
| 7 | 6317460 | 17 | 227923920 | 26 | 5983866 | 35 | 36 |
| 8 | 16933905 | 18 | 204956724 | 27 | 2699508 | 36 | 1 |
| 9 | 33969628 | | | | | | |

**Table A2.** The number of decomposable graphs with 9 nodes, for each possible number of edges. The number of decomposable graphs with 9 nodes and $e$ edges is $n(e)$.

# Appendix II: Glossary of terms related to graphs

See also section 2.1. The definitions refer to a general graph $G = (V, E)$. Vague terms are marked "(Vague.)" Some of the vague terms have been given precise definitions in certain contexts, as described in the main text.

**absent** An edge $e$ is absent if $e \notin E$. (Most authors use "missing", which I think is worse, since firstly it suggests there is something wrong, and secondly it is not the natural opposite of "present", which is clearly the best word for what it means.)

**chordal graph** See section 2.1.

**clique** A maximal complete subgraph.

**component / connected component** A maximal set of nodes such that there is a path between any pair of them.

**connected** A graph is connected if for any two nodes $u, v \in V$ there is a path from $u$ to $v$.

**cycle** A cycle is a path $(u_1, u_2, \ldots, u_k)$ where $k \geq 3$ and $(u_k, u_1) \in E$. (In graph theory, cycles are also called "loops" or "circuits"—see for example Even 1979. In Van Lint & Wilson 2001, a combinatorics book, they are called "polygons".)

**decomposable graph** See section 2.1.

**degree** The degree of a node is the number of edges that are incident to it.

**dense** (Vague.) This is the opposite or negation of "sparse", *q.v.* In Bollobás & Riordan (2011), dense graphs have $\Theta(n^2)$ edges.

**directed edge** An edge $(u, v)$ such that $(v, u) \notin E$. A directed edge $(u, v)$ is drawn as an arrow from $u$ to $v$.

**directed graph** A graph in which all the edges are directed. (Pearl 1988, page 232, uses "multiply connected network" to mean a directed graph that is not necessarily a forest.)

**directed path** In a directed graph, a sequence of nodes $u_1, u_2, \ldots, u_k$ such that $(u_1, u_2), (u_2, u_3), \ldots, (u_{k-1}, u_k) \in E$ and $u_i \neq u_j$ for $i \neq j$.

**distance** The distance between two nodes is the number of edges on the shortest path between them.

**forest** A graph that has no cycles. It can also be defined as a graph whose connected components are all trees, *q.v.* (Pearl 1988 calls directed forests "polytrees" and "singly connected networks".)

**girth** The girth of a graph is the length of the shortest cycle that it contains, or $\infty$ if it has no cycles. So a graph is chordal (*q.v.*) if and only if its girth is either 3 or $\infty$.

**hub** (Vague.) A node whose degree is large.

**incident** An edge $(u, v)$ is incident to a node $w$ if $w = u$ or $w = v$.

**leaf** A node, especially in a tree or forest, whose degree is 1.

**length of a path** The number of edges on the path.

**locally tree-like** (Vague.) This has been interpreted in several ways, for example to mean that there are few short cycles or that there are none. See section 6.2.

**multiple edges** More than one edge between the same pair of nodes.

**neighbour** A neighbour of $v$ is a node $u$ such that $(u, v) \in E$ or $(v, u) \in E$.

**path** See also section 2.1.

(a) In an undirected graph, a path is a sequence of nodes $u_1, u_2, \dots, u_k$ such that $(u_1, u_2), (u_2, u_3), \dots, (u_{k-1}, u_k) \in E$ and $u_i \neq u_j$ for $i \neq j$.

(b) In a directed graph, a path is a sequence of nodes $u_1, u_2, \dots, u_k$ such that either $(u_1, u_2) \in E$ or $(u_2, u_1) \in E$, either $(u_2, u_3) \in E$ or $(u_3, u_2) \in E$, ..., either $(u_{k-1}, u_k) \in E$ or $(u_k, u_{k-1}) \in E$, and $u_i \neq u_j$ for $i \neq j$. This is my definition, and it is non-standard. (Under standard definitions, a path in a directed graph has to be directed, and what I call a path would probably be called an "undirected path".)

(c) For there to be a path between $A \subseteq V$ and $B \subseteq V$ means that there is a path between some $u \in A$ and some $v \in B$.

**present** An edge $e$ is present if $e \in E$.

**rooted forest** A directed forest in which each component is a rooted tree. (Heckerman et al 1995, page 226, calls these "branchings".)

**rooted tree** See Definition 9.1 in section 9.2. A directed tree in which one node is designated the root, and the paths from the root to all the other nodes are directed paths. The text just after Definition 9.1 gives three other equivalent definitions. (Heckerman et al 1995, page 226, calls these "tree-like networks". Pearl 1988, pages 143 and 150, uses "causal tree" to mean a rooted tree graphical model.)

**self-loop** An edge from a node to itself (Mateti & Deo 1976, page 90).

**separate (verb)** Suppose $A, B, C \subseteq V$. If all paths from $A$ to $C$ pass through $B$, then $B$ separates $A$ from $C$. See Lauritzen (1996, page 6).

**simple** A graph is simple if it has no self-loops or multiple edges (Mateti & Deo 1976).

**size** The size of $G$ is $|E|$, the number of edges. (This term is used in graph theory and in Armstrong et al 2009.)

**span (verb)** A graph $H$ spans a graph $G = (V, E)$, or a set of nodes $V$, if it is connected and the nodes of $H$ are $V$. See also *spanning*.

**spanning** A spanning tree of a connected graph $(V, E)$ is a tree $(V, E')$ such that $E' \subseteq E$. This word is also used loosely in the phrase "spanning forest" (Edwards et al 2010, Lauritzen 2006). A spanning forest of a graph $(V, E)$ is a forest $(V, E')$ such that $E' \subseteq E$.

**sparse** (Vague.) A sparse graph is one that has few edges. For concrete definitions see section 6.2.

**tree** A connected graph that has no cycles. It can also be defined as a connected forest. In the machine-learning community, "tree" is sometimes used to mean forest (Bach & Jordan 2003, Bradley & Guestrin 2010), in which case trees are referred to as "spanning trees".

**triangulated graph** See section 2.1.

**undirected edge** An unordered pair $(u, v) \in E$; alternatively, an ordered pair $(u, v)$ such that $(u, v) \in E$ and $(v, u) \in E$.

**undirected graph** A graph in which all the edges are undirected.

**undirected path** See *path* (b).

# Appendix III: Asymptotic notations

These are used in chapter 6.

- $f(n) = O\big(g(n)\big)$ means there are some numbers $k$ and $N$ such that $|f(n)| \leq k|g(n)|$ for all $n \geq N$.
- $f(n) = \Theta(g(n))$ means there are some numbers $a, b, N > 0$ such that $ag(n) \leq f(n) \leq bg(n)$ for all $n \geq N$. This implies that $f(n) = O\big(g(n)\big)$.
- $f(n) = o(g(n))$ means that $\lim_{n\to\infty} \frac{f(n)}{g(n)} = 0$. This implies that $f(n) = O\big(g(n)\big)$.
- $f(n) \sim g(n)$ means that $\lim_{n\to\infty} \frac{f(n)}{g(n)} = 1$. This implies that $f(n) = \Theta\big(g(n)\big)$.
- $f(n) = \Omega(g(n))$ means there are some numbers $k$ and $N$ such that $|f(n)| \geq k|g(n)|$ for all $n \geq N$.

For more notations of this type, and their origins, see chapter 3 of Cormen et al (2009).

# References

I have not read the publications that are in foreign languages, though I have read English translations where these are listed.

==========================================================================================================

The following corrections have been made in this compact version of the thesis.

- In section 9.4, in Algorithm IX, in the comment to the right of line 17, "line 20" has been changed to "line 19".
- Also in Algorithm IX, this new line has been inserted: "21. Remove previous from ch(current)". In the two paragraphs just before this algorithm, the references to line-numbers in the algorithm have been corrected accordingly.
- In section 11.1, in the proof of Proposition 11.3, the second $g(W(v_2))$ has been changed to $g(W(v_3))$.
- In section 11.7, "to give SSSD a chance to visit a reasonable number of graphs" has been moved from the caption of Figure 11.11 to the caption of Figure 11.12.

Several copyediting errors have also been corrected.